\documentclass[lettersize,journal]{IEEEtran}
\usepackage{amsmath,amsfonts}
\usepackage{algorithmic}
\usepackage{algorithm}
\usepackage{array}
\usepackage[caption=false,font=normalsize,labelfont=sf,textfont=sf]{subfig}
\usepackage{textcomp}
\usepackage{stfloats}
\usepackage{url}
\usepackage{verbatim}
\usepackage{graphicx}
\usepackage{cite}
\usepackage{tikz}
\usepackage{bbding}
\usepackage{fontawesome}
\usepackage{amssymb}
\usepackage{lipsum}
\usepackage{multirow}
\usepackage{xcolor}

\usepackage{xcolor}
\usepackage{amsmath}
\usepackage{amssymb}
\usepackage{array}
\usepackage{multirow}
\usepackage{color}
\usepackage{colortbl}
\usepackage{framed}
\usepackage{bm}
\usepackage{bbm}
\usepackage{xspace}
\usepackage{enumitem}
\usepackage{algorithm}
\usepackage{listings}
\usepackage{url}
\usepackage{booktabs}
\usepackage{pifont} 
\usepackage[accsupp]{axessibility}
\usepackage{lipsum}
\usepackage{bm}
\definecolor{LightCyan}{rgb}{0.88,0.95,1}
\definecolor{blond}{rgb}{0.98, 0.94, 0.75}
\definecolor{green}{rgb}{0.0, 0.62, 0.42}
\definecolor{LightGray}{gray}{0.93}
\definecolor{ourcolor}{rgb}{0.925, 0.902, 0.969}
\definecolor{ourcolorl}{rgb}{0.95, 0.945, 0.975}

\definecolor{GrayLight}{gray}{0.35}
\newcommand{\textgray}[1]{\textcolor{GrayLight}{#1}}

\newcommand{\cmark}{\ding{51}}%
\newcommand{\xmark}{\ding{55}}%

\newcommand{\ours}{ReT\xspace}
\newcommand{\oursnew}{ReT-2\xspace}

\def \ie {\emph{i.e.}}
\def \eg {\emph{e.g.}}
\def \etal {\emph{et al.}}

\newcommand{\tit}[1]{\smallbreak\noindent\textbf{#1.}}
\newcommand{\tinytit}[1]{\noindent\textbf{#1.}}

\newcommand{\query}{\operatorname{\bm{q}}}
\newcommand{\doc}{\operatorname{\bm{d}}}
\newcommand{\retQ}{\operatorname{\text{\ours}_{\textbf{Q}}}}
\newcommand{\retD}{\operatorname{\text{\ours}_{\textbf{D}}}}
\newcommand{\retQQ}{\operatorname{\text{\oursnew}_{\textbf{Q}}}}
\newcommand{\retDD}{\operatorname{\text{\oursnew}_{\textbf{D}}}}

\newcommand{\inputH}{\bm{h}_l}
\newcommand{\nextH}{\bm{h}_{l+1}}
\newcommand{\finalH}{\bm{h}_{L}}
\newcommand{\finalHhat}{\overline{\bm{h}}_{L}}
\newcommand{\hatH}{\hat{\bm{h}_l}}

\newcommand{\candidate}{\mathbf{c}_l}
\newcommand{\candidateNext}{\bm{c}_{l+1}}

\newcommand{\genericZ}{\bm{z}^m_l}
\newcommand{\textZ}{\bm{z}^T_l}
\newcommand{\visZ}{\bm{z}^V_l}

\newcommand{\textInputGate}{\bm{i}^T_l}
\newcommand{\visInputGate}{\bm{i}^V_l}
\newcommand{\genericInputGate}{\bm{i}^m_l}
\newcommand{\forgetGate}{\bm{f}_l}

\newcommand{\genericInputW}{\bm{W}^m_i}
\newcommand{\textForgetW}{\bm{W}^T_f}
\newcommand{\visForgetW}{\bm{W}^V_f}

\definecolor{customgray}{gray}{0.35}

\newcommand{\rev}[1]{\textcolor{black}{#1}}

\newcommand{\revv}[1]{\textcolor{black}{#1}}
\begin{document}

\title{Recurrence Meets Transformers for\\Universal Multimodal Retrieval}

\author{Davide Caffagni$^*$, Sara Sarto$^*$, Marcella Cornia, Lorenzo Baraldi, Rita Cucchiara
\thanks{$^*$The first two authors equally contributed to this research.}
\thanks{D. Caffagni, S. Sarto, L. Baraldi, and R. Cucchiara are with the Department of Engineering ``Enzo Ferrari'', University of Modena and Reggio Emilia, Italy (e-mail: \{davide.caffagni, sara.sarto, lorenzo.baraldi, rita.cucchiara\}@unimore.it).}
\thanks{M. Cornia is with the Department of Education and Humanities, University of Modena and Reggio Emilia, Italy (e-mail: marcella.cornia@unimore.it).}
}

\markboth{IEEE Transactions on Pattern Analysis and Machine Intelligence}%
{Caffagni \MakeLowercase{\textit{et al.}}: Recurrence Meets Transformers for Universal Multimodal Retrieval}


\maketitle

\begin{abstract}
With the rapid advancement of multimodal retrieval and its application in LLMs and multimodal LLMs, increasingly complex retrieval tasks have emerged. Existing methods predominantly rely on task-specific fine-tuning of vision-language models and are limited to single-modality queries or documents. In this paper, we propose \oursnew, a unified retrieval model that supports multimodal queries, composed of both images and text, and searches across multimodal document collections where text and images coexist. \oursnew leverages multi-layer representations and a recurrent Transformer architecture with LSTM-inspired gating mechanisms to dynamically integrate information across layers and modalities, capturing fine-grained visual and textual details. \rev{We evaluate \oursnew on the challenging M2KR, M-BEIR, and MMEB benchmarks across different retrieval configurations}. Results demonstrate that \oursnew consistently achieves state-of-the-art performance across diverse settings, while offering faster inference and reduced memory usage compared to prior approaches. When integrated into retrieval-augmented generation pipelines, \oursnew also improves downstream performance on Encyclopedic-VQA and InfoSeek datasets. Our source code and trained models are publicly available at: \url{https://github.com/aimagelab/ReT-2}.
\end{abstract}


\section{Introduction}
\begin{figure}[t]
    \centering
    \includegraphics[width=0.95\linewidth]{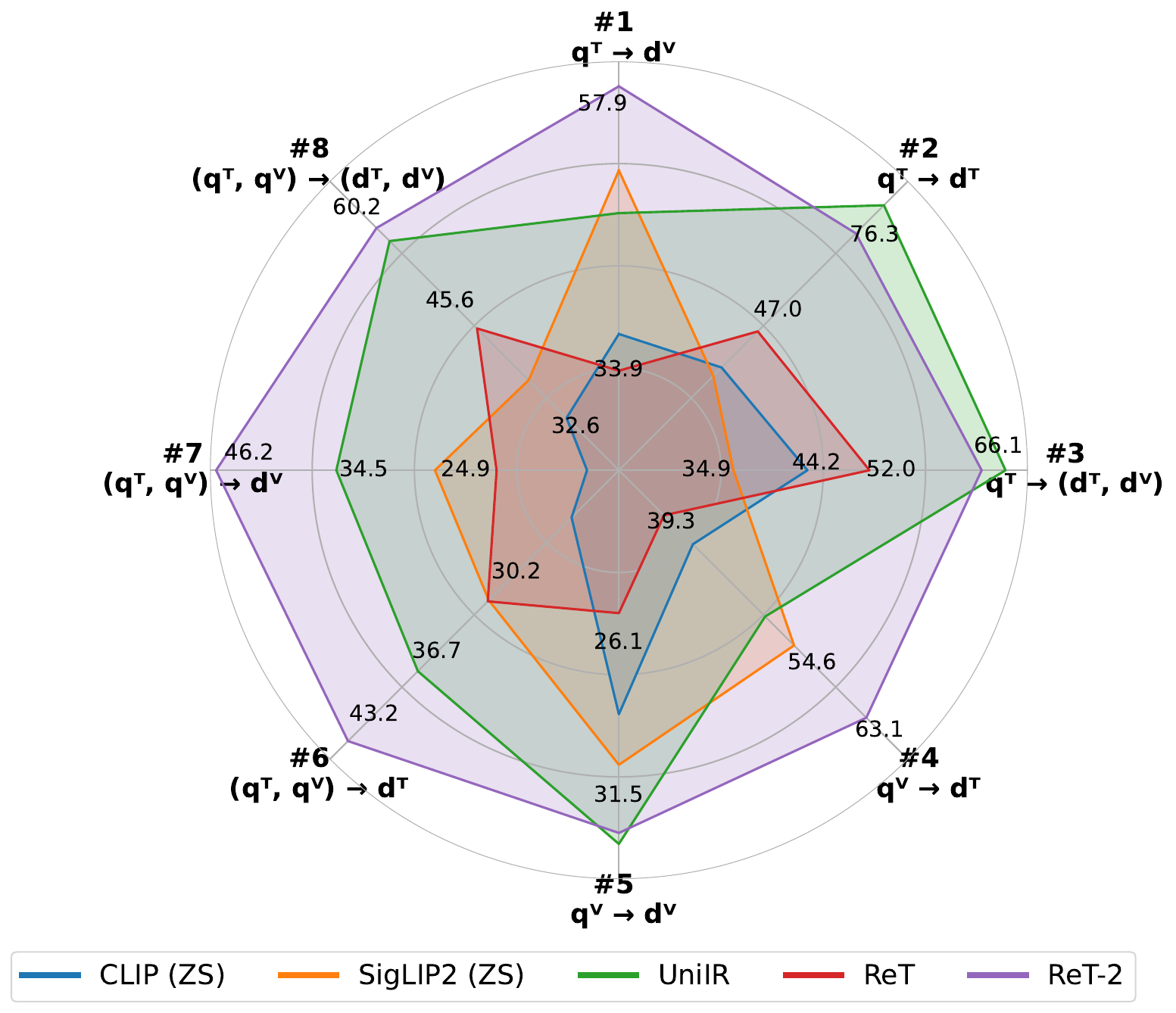}
    \vspace{-0.2cm}
    \caption{In this work, we present \colorbox{ourcolor}{Recurrence-enhanced Transformer (\textbf{\oursnew})}, a novel retrieval approach supporting different tasks and data configurations, from cross-modal image-to-text retrieval -- \ie, \ $q^V \rightarrow d^T$, to multimodal text--image-to-text--image retrieval -- \ie, \ $(q^T,q^V)\rightarrow(d^T,d^V)$.
    The plot shows average results on the M-BEIR benchmark tasks~\cite{wei2024uniir}, highlighting the performance gains of the proposed method over its previous version (\ie, \ours~\cite{caffagni2025recurrence}) and other state-of-the-art methods.}
    \label{fig:first_page}
    \vspace{-0.2cm}
\end{figure}

\IEEEPARstart{I}{nformation} retrieval is a fundamental and challenging task that entails identifying relevant content from large and heterogeneous data collections to satisfy user information needs. Early approaches predominantly focused on unimodal retrieval, where queries and retrievable items belonged to the same modality, such as text or images~\cite{izacard2021unsupervised, zheng2017sift, noh2017large}. In recent years, however, the field has progressively shifted towards multimodal data~\cite{lin2014microsoft,sharma2018conceptual,schuhmann2022laion}, reflecting the growing presence of images, text, and other media in real-world applications. The advent of vision-language models, including CLIP~\cite{radford2021learning}, ALIGN~\cite{jia2021scaling} and other variants~\cite{cherti2023reproducible,zhai2023sigmoid,tschannen2025siglip}, further enabled effective cross-modal retrieval, allowing, for example, natural language queries to retrieve relevant images or vice versa.

At the same time, driven by the advent of Multimodal Large Language Models (MLLMs)~\cite{caffagni2024revolution,liu2024improved, cocchi2025llava,bai2025qwen2} and the growing prominence of visual question answering tasks, there is an increasing demand for retrieval models capable of handling \textit{multimodal queries} and retrieving \textit{multimodal documents}, where multiple modalities coexist both within the query and the items to be retrieved. A typical example involves a query combining an image and a related text question, or specifying which part of the image should be retrieved~\cite{wei2024uniir,lin2024preflmr,chen2023can,mensink2023encyclopedic}. Despite significant progress in multimodal retrieval, existing state-of-the-art methods are largely limited to single-modality queries and documents, and thus fail to fully satisfy the flexibility required by modern applications.

To address these challenges, in this paper, we propose a retrieval approach that natively supports multimodal queries and documents (consisting of both text and images), and that can also handle scenarios with missing modalities from either query or document side. Our approach enables a more general retrieval paradigm (\ie, universal multimodal retrieval), where multiple modalities and diverse retrieval tasks can be accommodated within a single unified framework. Unlike previous approaches that rely on feature fusion from only the last layer of vision-language backbones~\cite{wei2024uniir}, our method exploits multi-layer representations for both modalities. We argue that explicitly incorporating features from shallower layers allows the model to better capture the wide variety of multimodal queries and documents, including fine-grained visual or textual details that are often lost in deeper layers. Moreover, we complement this design with an analysis of layer activations, which allows us to identify and prune redundant layers, thereby reducing computational overhead while simultaneously enhancing robustness.

To achieve these goals, we design a Transformer-based recurrent cell that, at each layer, merges features from the visual and textual backbones with its internal states. Inspired by the gating mechanism of an LSTM~\cite{hochreiter1997long}, our model employs a forget gate to control how much information to retain from shallower layers, while textual and visual input gates modulate the unimodal information flow. This design enables our model to dynamically determine which layers and modalities are most informative for encoding each query or document. \rev{In stark contrast to previous work that combines Transformers and recurrence architectures~\cite{wang2019r, hutchins2022block}, we do not utilize recurrence to model temporal dependencies among tokens. Rather, we exploit a Transformer-based recurrent cell to model the interaction across representations from multiple layers of pre-trained backbones. The recurrent design allows us to save parameters, as multiple layers are processed by the same, shared Transformer-based cell. On the other hand, while in a standard Transformer all the layer representations participate in the residual stream, the input and forget gates of our cell dynamically select which modality and which depth of representation can enter the hidden state.}

Our proposed model, which we call \textbf{\oursnew} (Recurrence-enhanced Transformer), is experimentally evaluated on the challenging M2KR benchmark~\cite{lin2024preflmr}, which integrates a diverse collection of datasets adapted for multimodal retrieval. To further broaden the evaluation, we extend the M2KR benchmark by augmenting the OVEN~\cite{hu2023open}, InfoSeek~\cite{chen2023can}, Encyclopedic-VQA~\cite{mensink2023encyclopedic}, and OKVQA~\cite{marino2019ok} splits to incorporate images within the reference documents.
In addition, we conduct experiments on the M-BEIR benchmark~\cite{wei2024uniir}, focusing particularly on settings where certain modalities are absent. Across more than eight distinct retrieval configurations, including text-to-image, text--image-to-image, and text--image-to-text--image retrieval, \oursnew consistently demonstrates strong and stable performance, as summarized in Fig.~\ref{fig:first_page}. \rev{Across all these experiments, including zero-shot evaluations on the MMEB benchmark~\cite{jiangvlm2vec}, our results highlight not only the effectiveness of our approach in conventional retrieval scenarios, but also its ability to generalize to highly compositional and underexplored multimodal configurations.}

Finally, we demonstrate the utility of \oursnew as a retrieval backbone for retrieval-augmented generation in knowledge-intensive visual question answering. In this setting, many questions can not be answered without retrieving external multimodal knowledge, making retrieval quality a decisive factor. Our experiments, conducted on Encyclopedic-VQA~\cite{mensink2023encyclopedic} and InfoSeek~\cite{chen2023can}, show that \oursnew provides more effective retrieval support compared to alternative retrieval methods, enabling off-the-shelf MLLMs~\cite{cocchi2025llava,bai2025qwen2} to achieve higher answer accuracy without task-specific fine-tuning.

Beyond retrieval accuracy and its effectiveness when employed as retrieval backbone for downstream tasks, we further assess the computational efficiency of \oursnew in comparison to existing state-of-the-art methods. Our analysis reveals that the benefits of \oursnew are not confined to retrieval effectiveness, but also extend to practical efficiency, achieving faster inference and reduced memory usage than competing methods.

In summary, our main contributions are as follows:
\begin{itemize}
    \item We present \oursnew, a unified retrieval model that supports multimodal queries and documents, equipped with a recurrence-enhanced Transformer cell that integrates visual and textual features via LSTM-inspired gating.  
    \item Unlike prior approaches that rely only on final-layer features, we exploit multi-layer representations and introduce a pruning strategy to remove redundant layers, improving both robustness and efficiency.  
    \item \rev{Extensive experiments on the M2KR, M-BEIR, and MMEB benchmarks demonstrate state-of-the-art performance across a wide range of multimodal retrieval tasks.}
    \item We further show that \oursnew boosts retrieval-augmented generation for knowledge-intensive VQA, enabling off-the-shelf MLLMs to achieve higher answer accuracy.  
\end{itemize}

This work is an extended and improved version of our earlier conference paper~\cite{caffagni2025recurrence}. Compared to our previous approach (\ie, \ours), the current work provides a deeper architectural analysis and introduces several architectural modifications that collectively improve both efficiency and robustness. These advances lead to a conceptually simpler yet more effective model, allowing \oursnew to set a stronger foundation for universal multimodal retrieval and its downstream applications.
\section{Related Work}
\label{sec:related}
\tinytit{From Unimodal Retrieval to Cross-Modal Retrieval}
Classical retrieval methods were largely unimodal, focusing on either text-based document search or content-based image retrieval~\cite{izacard2021unsupervised, zheng2017sift, noh2017large}. While effective in their domains, they lacked the ability to bridge modalities. The advent of large-scale vision-language datasets~\cite{sharma2018conceptual,schuhmann2022laion} and dual-encoder models such as CLIP~\cite{radford2021learning} and its variants~\cite{zhai2023sigmoid,sun2023eva,tschannen2025siglip} marked a turning point, enabling contrastive learning to align images and text in a shared embedding space. Despite this progress, these models are typically evaluated on relatively small benchmarks like Flickr30k~\cite{plummer2015flickr30k} and COCO~\cite{lin2014microsoft}, which emphasize simple queries and limit generalization.
Building on these advances, more complex retrieval scenarios have emerged, including composed image retrieval~\cite{liu2021image}, long-text-to-image retrieval~\cite{zhang2024long}, and multimodal query-to-multimodal document retrieval~\cite{chen2023can, hu2023open}. Current methods typically address such tasks through specialized fine-tuning~\cite{miech2021thinking, brown2020smooth, baldrati2022conditioned}, but a universal framework capable of seamlessly accommodating diverse query and document modalities remains an open challenge.

\tit{Universal Multimodal Retrieval}
With a rising demand for multimodal retrieval systems, the ability to handle complex multimodal queries has become essential. This trend has led to the development of specialized benchmarks for multimodal retrieval, supporting diverse tasks and data configurations. For instance, M2KR~\cite{lin2024preflmr} combines several datasets for this task. Similarly, the large-scale M-BEIR benchmark~\cite{wei2024uniir} covers a wide range of domains and image sources, enabling comprehensive evaluation of multimodal retrieval approaches.

The challenge of developing robust multimodal representations remains a foundational question in multimodal learning, driving research toward effective strategies for encoding queries and documents across modalities. UniIR~\cite{wei2024uniir} integrates modalities using features from the last layer of pre-trained models~\cite{radford2021learning,li2022blip}, aiming to build a unified retriever for diverse tasks. Similarly, GENIUS~\cite{kim2025genius} is a flexible generative retrieval framework that converts multimodal inputs into discrete representations and enhances generalization through query-target interpolation. Meanwhile, models like FLMR~\cite{lin2023fine} and PreFLMR~\cite{lin2024preflmr} explore a late-interaction paradigm~\cite{khattab2020colbert}, where multimodal queries and text-only documents are encoded independently into sets of latent tokens, and relevance scores are computed by aggregating token-level similarities. 

With the recent advancements in LLMs~\cite{brown2020language,dubey2024llama}, research has increasingly turned to multimodal models to align visual and textual modalities via visual instruction tuning~\cite{caffagni2024revolution}. Despite these advances, the potential of MLLMs for universal retrieval tasks remains relatively underexplored, with approaches such as LamRA~\cite{liu2025lamra} and MM-Embed~\cite{lin2025mm} attempting to repurpose MLLMs for the task. However, employing MLLMs in this setting typically requires multi-stage finetuning of a large number of parameters, leading to substantial training costs and limited inference efficiency. To address these challenges, recent works explore more efficient strategies, such as PUMA~\cite{lyu2025puma}, which prunes parameters to reduce computational overhead, and JFE~\cite{huang2025joint}, which leverages early visual-textual fusion to enhance cross-modal understanding.

In contrast, \oursnew is designed to efficiently integrate multi-layer visual and textual features thanks to a recurrent-enhanced architecture, achieving performance comparable to MLLM-based models while avoiding their high computational costs.

\tit{Recurrence-Augmented Transformers} Transformer architectures~\cite{vaswani2017attention} have achieved impressive results across diverse domains, from natural language understanding~\cite{brown2020language,wolf2020transformers,touvron2023llama} to computer vision~\cite{dosovitskiy2021image,touvron2021training,zhai2022scaling,khan2022transformers}. However, their quadratic complexity with respect to input length has driven research into alternative designs. One line of work integrates recurrent mechanisms within Transformer models, interleaving Transformer layers with recurrent neural networks to balance attention with sequential processing~\cite{lei2021attention,lei2017simple,bapna2018best}. Other approaches, such as the R-Transformer~\cite{wang2019r}, incorporate local recurrent cells to enable parallel computation. The Block-Recurrent Transformer~\cite{hutchins2022block}, for instance, embeds recurrent dynamics inspired by LSTM cells~\cite{hochreiter1997long} directly within the Transformer framework. \rev{Unlike prior works that primarily use recurrence~to reduce the computational cost of modeling temporal dependencies, in this paper, we exploit recurrence to enable multi-layer and multimodal feature integration, aiming to enhance performance on multimodal retrieval tasks.}
\section{Background}
\label{sec:background}

\tinytit{Problem Formulation}
In our setting, both queries $\query = (q^T, q^V)$ and documents $\doc = (d^T, d^V)$ are structured as paired image-text instances. The textual component of each query, $q^T$, typically comprises an instruction, \eg~``Utilizing the given image, obtain documents that respond to the following question'', followed by a question specific to the associated query image $q^V$. 
The goal is to retrieve documents that are most relevant to the given query. Each document consists of a textual response $d^T$ that addresses the query and may optionally include a corresponding image $d^V$.

This task presents several significant challenges. It demands the ability to interpret fine-grained visual and linguistic cues, establish coherent alignment between multimodal semantics in the query and candidate documents, and perform reasoning across visual and textual modalities. Furthermore, the presence or absence of images in documents introduces variability in the retrieval signal, making the matching process more complex.

\tit{\ours: Recurrent Transformer with Fine-grained Late Interaction}
\ours~\cite{caffagni2025recurrence} is a multimodal retrieval model that introduces a novel Transformer-based recurrent cell designed to fuse multimodal features from both queries and documents, enabling the computation of fine-grained similarity scores between them. The model leverages pre-trained visual and textual backbones, aggregating features across multiple layers to construct rich representations of each modality.
While leveraging a Transformer architecture, \ours incorporates a learnable gating mechanism inspired by LSTMs~\cite{hochreiter1997long} to regulate information flow across layers. At each recurrent step, it combines its internal state with the visual and textual features extracted from the current backbone layer, treating lower-level features as the past. This mechanism enables the model to selectively preserve or discard earlier-layer information, allowing it to emphasize more meaningful high-level features during fusion.

\begin{figure*}[t]
    \centering
    \includegraphics[width=0.98\linewidth]{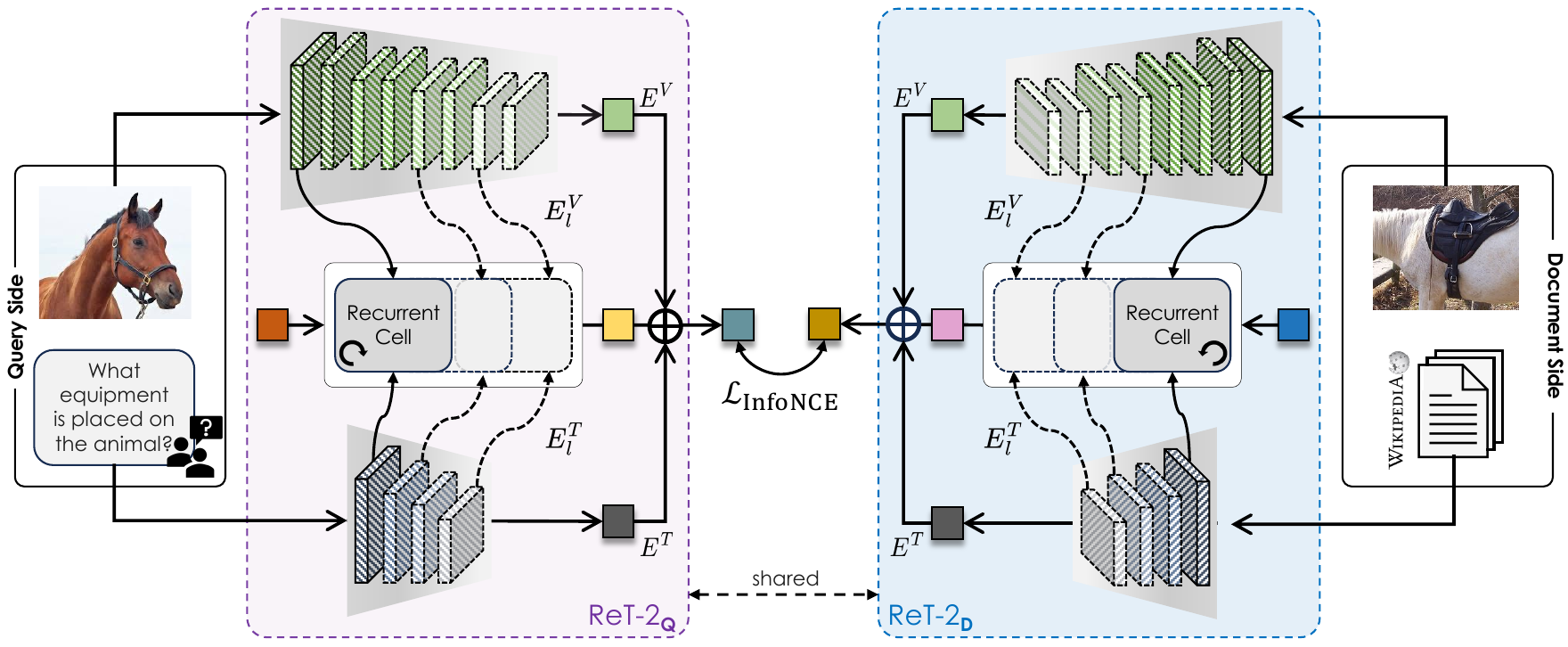}
    \vspace{-0.15cm}
    \caption{Overview of the proposed \colorbox{ourcolor}{Recurrence-enhanced Transformer (\textbf{\oursnew})} for universal multimodal retrieval. 
    }
    \label{fig:model}
    \vspace{-0.2cm}
\end{figure*}

The model comprises two dedicated encoders for queries and documents, $\retQ$ and $\retD$, which share the same architecture but maintain separate learnable parameters that are optimized jointly. Each encoder integrates a recurrent cell with pre-trained visual and textual backbones. Specifically, for each modality $m \in {T, V}$, the unimodal backbone produces a set of activations $E^m(q^m) = {E^m_l(q^m)}_{l=1}^L$, where $E^m_l(q^m) \in \mathbb{R}^{N \times d}$ denotes the features extracted from the $l$-th layer and $L$ is the total number of layers. At each layer $l$, the recurrent cell performs \textit{feature fusion}~\cite{wei2024uniir} over three inputs: the hidden state from the previous step, $\mathbf{h}_l \in \mathbb{R}^{k \times d}$, and the visual and textual representations $\mathbf{E}^V_l$ and $\mathbf{E}^T_l$. For the initial step, the hidden state $\mathbf{h}_0$ is initialized with $k=32$ learnable vectors.

Formally, given a query-document pair $(\query, \doc)$, each side uses distinct learnable input tokens. The final outputs are:
\begin{align}
    \mathbf{Q} & =  \retQ(\query) \in \mathbb{R}^{k \times \overline{d}}\\
    \mathbf{D} & = \retD(\doc)^\intercal \in \mathbb{R}^{\overline{d} \times k},
\end{align}
where $\overline{d}$ is the dimension after projection.

During training, these representations are used to compute a fine-grained late-interaction~\cite{santhanam2022colbertv2} relevance score:
\begin{equation}
\label{eq:fine-grained loss}
s(\textbf{Q}, \textbf{D}) = \sum_{i=1}^k \max_{j=1 \dots k} \textbf{Q}_i \cdot \textbf{D}_j.
\end{equation}
Here, similarity is computed as the dot product between the $i$-th query and $j$-th document token. The $\max$ operator ensures that only the most relevant document tokens contribute to the score of each query token, effectively filtering out locally irrelevant matches.
Training is performed by jointly optimizing both the query and the document encoder with an InfoNCE loss~\cite{radford2021learning}, where global query-document cosine similarities are replaced with the score defined in Eq.~\ref{eq:fine-grained loss}.

\tit{Limitations of ReT}
Our previous work, ReT, demonstrated strong retrieval performance, validating the effectiveness of recurrent multimodal fusion. However, there remains room for improvement in both efficiency and efficacy. \rev{Given the recurrent nature of the architecture, reducing the number of fused layers could lead to faster inference, thereby improving computational efficiency and reducing redundant representations across depth~\cite{dalvi2020analyzing,he2024matters}.} \revv{Beyond efficiency, processing every backbone layer also lengthens the recurrent chain that low-level information must survive and repeatedly exposes the cell's gates to near-identical, residual-stream-redundant evidence across consecutive layers, both of which increase the risk of overwriting or replaying the same fusion decision rather than genuinely refining it (cf. Appendix~\ref{sec:supp_theory}).} Additionally, ReT encodes queries and documents into $32 \times 128$ matrices (\ie, $k\times\overline{d}$). We empirically observe that these matrices suffer from rank collapse~\cite{josephlambda}, where their rows converge to a uniform representation, undermining the purpose of leveraging multiple embeddings to capture diverse nuances of the input. \revv{This is consistent with the theoretical account of rank collapse in self-attention networks~\cite{dong2021attention}, which we relate directly to the recurrent cell in Appendix~\ref{sec:supp_theory}.} This raises the question of whether a single, larger embedding is better than a small embedding matrix for multimodal retrieval.

\section{Proposed Method}
\label{sec:method}
In this section, we introduce an enhanced variant of \ours, referred to as \textbf{\oursnew}, which is specifically designed to address the limitations identified in the original model. \oursnew aims to improve retrieval effectiveness and efficiency when dealing with heterogeneous data sources in large-scale, multimodal collections of entities. A graphical overview of the architecture of our \oursnew model is shown in Fig.~\ref{fig:model}.

\subsection{Overall Architecture}
\label{sec:method_ret}
In our \oursnew model, the architecture retains two dedicated encoders for queries and documents. However, in contrast to the previous version (which employed separate parameter sets optimized jointly), \oursnew introduces a unified encoder architecture with shared weights for both modalities. Specifically, each encoder comprises a recurrent fusion cell coupled with pre-trained, learnable visual and textual backbones. This parameter sharing not only reduces model complexity and reduces overfitting, but also encourages consistent representation learning across queries and documents. \revv{This design rests on the assumption that queries and documents are represented through compatible combinations of visual and textual inputs and must be mapped into a common semantic space for their symmetric dot-product similarity to be meaningful; the same assumption should hold, more broadly, for any symmetric or approximately symmetric dual-encoder retrieval setting.}

In the following, we retain the notation introduced in Sec.~\ref{sec:background} and denote the cross-attention~\cite{vaswani2017attention}  between two matrices $\mathbf{x}$ and $\mathbf{y}$, as $\texttt{Attention}(\mathbf{x}, \mathbf{y})$.

\begin{figure}[t]
    \centering
    \includegraphics[width=0.98\linewidth]{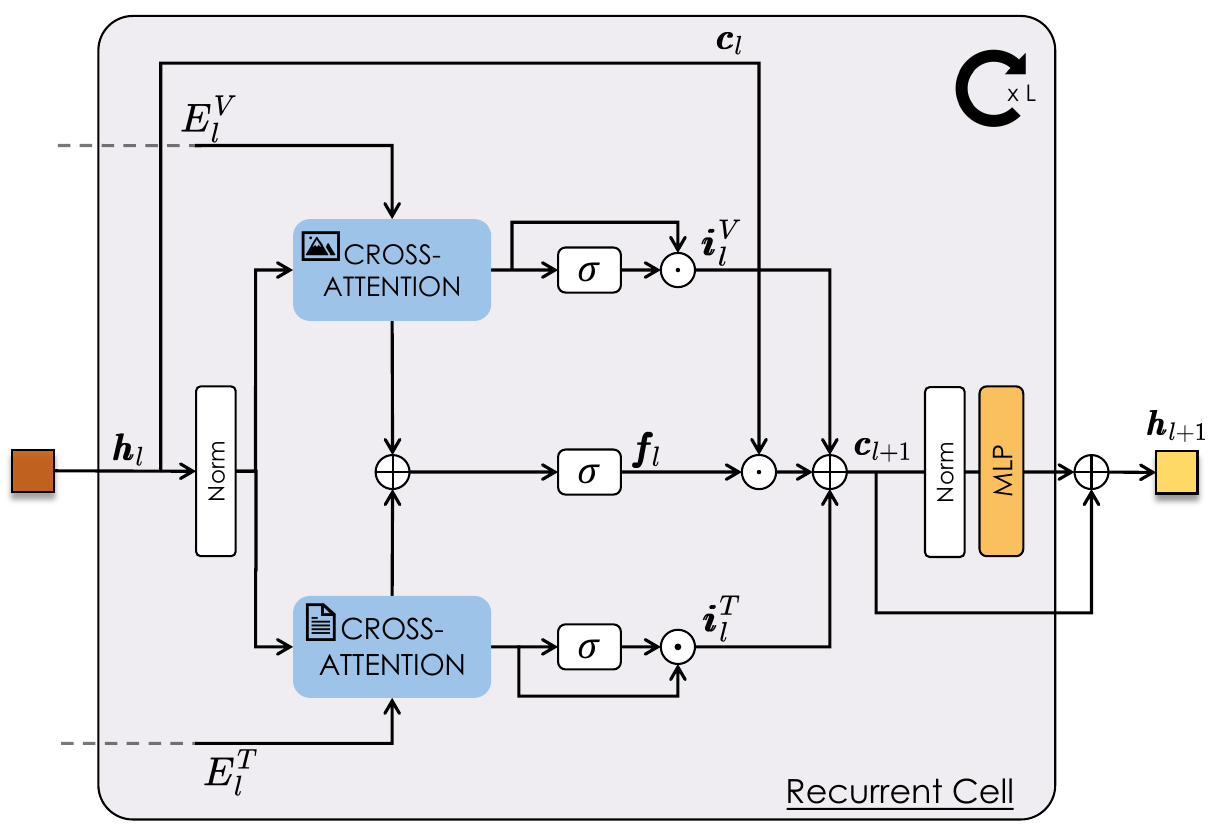}
    \vspace{-0.2cm}
    \caption{Graphical illustration of the proposed recurrent cell for multimodal retrieval, which integrates layer-specific textual and visual features into a matrix-form hidden state.}
    \label{fig:recurrent_block}
    \vspace{-0.3cm}
\end{figure}

\tit{Recurrent Cell}
The architecture of the recurrent cell is illustrated in Fig.~\ref{fig:recurrent_block}. Within the cell, the input hidden state $\inputH$ is processed through three parallel branches. The first branch retains the candidate hidden state $\candidate$ of the recurrent cell. Notably, for layers $l \ge 1$, $\inputH$ encodes accumulated, layer-specific representations of both the image and text. Rather than processing all layers of the visual and textual backbones, we consistently sample three representative layers: one from the lower (early), one from the middle, and one from the upper (final) sections of each backbone. This approach ensures a balanced capture of low-, mid-, and high-level features while maintaining computational efficiency and architectural compatibility across backbones of varying depth.

To effectively incorporate contextual information from both modalities, the remaining two branches perform feature fusion between $\inputH$ and the unimodal visual and textual representations extracted from the $l$-th layer of their respective backbones.

Specifically, we employ two independent cross-attention modules to fuse the normalized input $\hatH$ with the visual and textual representations, respectively, as
\begin{equation}
    \label{eq:unimodal_component}
    \genericZ =  \texttt{Attention}\left(\hatH, \, E^m_l\right),
\end{equation} 
where $m \in {T, V}$ and $\hatH=\texttt{LayerNorm}(\inputH)$.

The outputs of the three branches are combined to compute the updated internal state of the recurrent cell. This state is formed as a gated linear combination of the candidate state $\candidate$ and the outputs from the two feature fusion branches, denoted as $\textZ$ and $\visZ$. The combination is modulated by a set of learnable forget and input gates.

In detail, the forget gate $\forgetGate$ controls the extent to which information from earlier applications of the recurrent cell (corresponding to shallower layers, or the ``past'') is retained in the current step, based on the ongoing multimodal interaction $\genericZ$. In parallel, the input gates $\genericInputGate$ regulate the influence of the unimodal features from the current ($l$-th) layer. This mechanism allows the model to attenuate noisy or less relevant high-level representations when fine-grained visual or textual details (\eg, colors or shapes) are more pertinent to the query.
Formally, the next candidate state is obtained as
\begin{equation}
    \label{eq:final}
    \candidateNext = \candidate \odot \forgetGate + \textZ \odot \textInputGate + \visZ \odot \visInputGate,
\end{equation}
where $\forgetGate$, $\textInputGate$ and $\visInputGate$ indicate the learnable sigmoidal gates. In particular, these are computed as follows:
\begin{equation}
\begin{aligned}
    \forgetGate &= \sigma\left(\textForgetW \cdot \textZ + \visForgetW \cdot \visZ + b_f\right), \\
    \genericInputGate &= \sigma\left(\genericInputW \cdot \genericZ + b_i\right),
\end{aligned}
\label{eq:forget_gate}
\end{equation}
where $\textForgetW, \visForgetW, \genericInputW$ are trainable weight matrices, and $b_f$, $b_i$ are fixed scalar biases.

The updated state $\candidateNext$ undergoes layer normalization and is passed through a residual two-layer feed-forward network to produce the output of the recurrent cell, as
\begin{equation}
    \label{eq:next_input}
    \nextH = \candidateNext +  \texttt{MLP}\left(\texttt{LayerNorm}\left(\candidateNext\right)\right).
\end{equation}

After going through different layers of the backbones, the output from the last iteration of the recurrent cell, $\finalH \in \mathbb{R}^{k \times d}$ (where $k=1$ in our novel formulation), consists of a latent token that serves to compute query-document relevance scores. 
Specifically, the output $\finalH$ is transformed into a different vector space through a linear projection $\textbf{W}_{final} \in \mathbb{R}^{d \times \overline{d}}$, \ie 
\begin{equation}
    \label{eq:pre-output}
    \finalHhat = 
     \finalH \  \cdot \textbf{W}_{final} .
\end{equation}

\tit{Global Feature Injection}
At the output of the recurrent cell, $\finalHhat$ encodes multimodal information that integrates details from multiple levels of abstraction. However, retaining access to the raw global features provides a broader contextual representation of the query or document. To leverage this complementary information, we augment the multimodal representation $\finalHhat$ with the unimodal outputs of the visual and textual backbones, denoted as $E^V$ and $E^T$, respectively. These typically correspond to the \texttt{CLS} visual pooler token and the \texttt{EOS} textual pooler token. The integration is performed by summing the global features with the output of the recurrent cell, obtaining the final representation of the query as
\begin{equation}
    \label{eq:sum}
    \finalHhat = 
     \finalHhat \ + E^V(q^V) + E^T(q^T).
\end{equation}
\revv{The recurrent and global paths are intended to be complementary: the recurrent cell selectively integrates layer-specific cues, which can come at the expense of broader semantic context, while the pooler tokens directly preserve such context. This complementarity is expected to hold whenever the global representation is itself well-trained and not already redundant with the fine-grained evidence accumulated by the recurrent branch, a property we expect to transfer to other architectures that aggregate hierarchical or multi-layer features.}

\tit{\rev{Missing Modalities}}
\rev{In practical use cases, queries and documents are not always multimodal. For instance, a user can formulate a purely textual query, asking to retrieve an image-text pair, as well as an image alone. \oursnew handles these scenarios by using an empty string to overcome a missing text, and a black picture in place of a missing image. Next, the input and forget gate components pertaining to the missing modality (cf. Eq.~\ref{eq:forget_gate}) and the unimodal global features extracted from the placeholder input (cf. Eq.~\ref{eq:sum}) are manually set to zero.}

\subsection{Training Procedure}
Given a query-document pair $(\query,\doc)$, along with a learnable token
in input for both the query and document sides, we denote the corresponding final output  $\finalHhat$ of the query and the document encoders as
\begin{align}
    \mathbf{Q} & = \retQQ(\query) \in \mathbb{R}^{k \times \overline{d}} \\
    \mathbf{D} & = \retDD(\doc)^\intercal \in \mathbb{R}^{\overline{d} \times k},
\end{align}
where $\retQQ=\retDD$ in our shared implementation.

Training is performed by optimizing both the query and the document encoder with the InfoNCE loss~\cite{radford2021learning}, where query-document similarity is computed as the dot-product between the query and the document token.

\begin{figure}[t]
    \centering
    \includegraphics[width=0.975\linewidth]{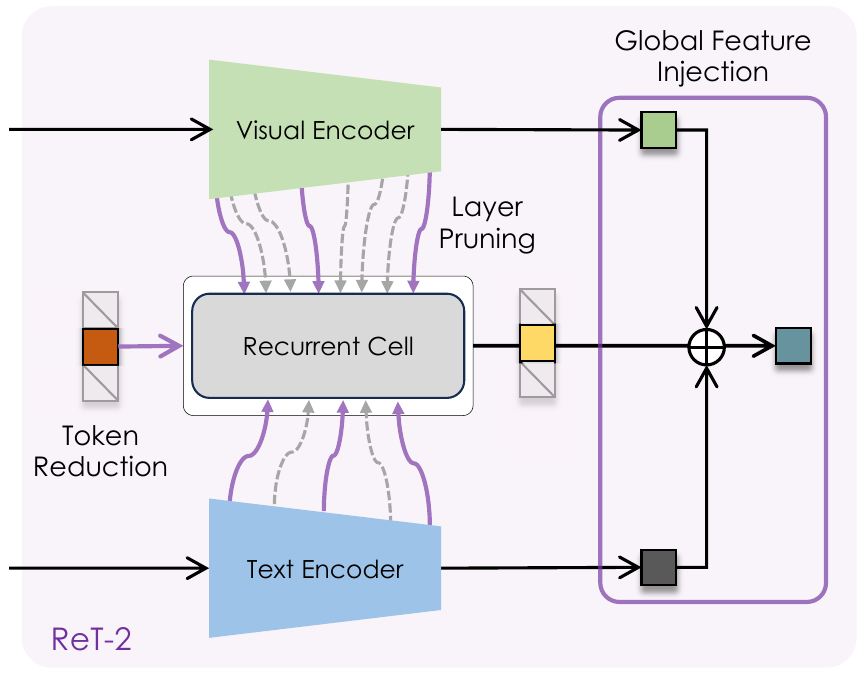}
    \vspace{-0.2cm}
    \caption{Visualization of the differences between the previous method (\ie, \ours) and the newly proposed \oursnew. The new method introduces three key modifications: (i) \textbf{token reduction}, instead of multiple input tokens, only a single token is used; (ii) \textbf{layer pruning}, rather than using all textual and visual layers, we now select only three representative layers (early, middle, and late), independent of the architecture; and (iii) \textbf{global feature injection}, a newly added module that integrates global information to enhance the representation. Highlighted regions indicate the most significant differences.}
    \label{fig:diff}
    \vspace{-0.3cm}
\end{figure}

\subsection{Summary: From \ours to \oursnew}
Overall, \oursnew incorporates several targeted changes to enhance the efficiency, robustness, and simplicity of the original \ours architecture:
\begin{itemize}
    \item \textbf{Shared Query-Document Encoder}: Unlike \ours, which used separate encoders with distinct learnable parameters for queries and documents, \oursnew adopts a shared architecture with tied weights, promoting consistency and reducing model complexity.
    \item \textbf{Token Reduction}: The number of input tokens is reduced from 32 to a single token for each query or document. This design choice addresses the issue of rank collapse observed in the output embeddings and encourages the model to produce more compact and meaningful representations. \revv{The resulting token can be interpreted as a task-specific \texttt{[CLS]}-like representation, recurrently accumulated rather than computed from the raw input.}
    \item \textbf{Simplified Contrastive Objective}: The use of a single token per side eliminates the need for the fine-grained contrastive loss used in \ours. Instead, we apply a standard InfoNCE loss directly over the single fused token from both the query and document, significantly simplifying the retrieval pipeline and improving inference efficiency.
    \item \textbf{Layer Pruning}: Rather than relying on all layers of the textual and visual backbones or attempting to explicitly align architectures with different depths, we always sample three layers: one from the lower (early), one from the middle, and one from the upper (final) part of each backbone. This strategy ensures compatibility and stability, especially when backbones differ in depth.
    \item \textbf{Global Feature Injection}: To enhance contextual understanding, \oursnew integrates global feature representations alongside layer-specific features. This injection of global context helps the model capture general information, further improving retrieval accuracy and robustness.
\end{itemize}

A visual summary of the modifications and improvements implemented in \oursnew is provided in Fig.~\ref{fig:diff}.

\section{Experiments on Multimodal Retrieval}
\label{sec:experiments}

\begin{table}[t]
\centering
\setlength{\tabcolsep}{.5em}
\caption{Selected layers for each backbone in \oursnew. $L$ denotes the depth of each backbone, measured in number of layers.}
\resizebox{0.97\linewidth}{!}{
\begin{tabular}{lcc cc cc}
\toprule
& & \multicolumn{2}{c}{\textbf{Text Encoder}} & & \multicolumn{2}{c}{\textbf{Visual Encoder}} \\
\cmidrule{3-4} \cmidrule{6-7}
\textbf{Backbone} & & $L$ & \textbf{Layer Indices} & & $L$ & \textbf{Layer Indices} \\
\midrule
CLIP ViT-B & & 12 & 3, 7, 11 & & 12 & 3, 7, 11 \\
ColBERTv2 & & 12 & 3, 7, 11 & & - & - \\
CLIP ViT-L & & 12 & 3, 7, 11 & & 24 & 3, 18, 23 \\
SigLIP2 ViT-L & & 24 & 3, 18, 23 & & 24 & 3, 18, 23 \\
OpenCLIP ViT-H & & 24 & 3, 18, 23 & & 32 & 4, 25, 31 \\
\bottomrule
\end{tabular}
}
\label{tab:n_layers}
\vspace{-0.3cm}
\end{table}

\subsection{Datasets and Evaluation Metrics}
\label{sec:datasets_ret}

\rev{We evaluate our models on the M2KR~\cite{lin2024preflmr}, M-BEIR~\cite{wei2024uniir}, and MMEB~\cite{jiangvlm2vec} benchmarks, which provide a diverse, large-scale collection of datasets and task configurations for comprehensive evaluation of multimodal retrieval performance.}

\tit{M2KR Benchmark}
M2KR integrates heterogeneous sources, including  WIT~\cite{srinivasan2021wit}, IGLUE~\cite{bugliarello2022iglue}, KVQA~\cite{shah2019kvqa}, CC3M~\cite{sharma2018conceptual}, 
MSMARCO~\cite{nguyen2016msmarco},
OVEN~\cite{hu2023open}, LLaVA~\cite{liu2024visual}, InfoSeek~\cite{chen2023can}, Encyclopedic-VQA~\cite{mensink2023encyclopedic}, and OKVQA~\cite{marino2019ok}. These datasets span a wide range of domains, enabling robust evaluation of retrieval models under varying degrees of complexity and multimodal reasoning.
To better align with our setting, where both queries and documents are multimodal, we augment the M2KR splits of OVEN, InfoSeek, Encyclopedic-VQA, and OKVQA by enriching the reference documents with associated images~\cite{caffagni2025recurrence}, thereby enabling a more effective evaluation of models that rely on both textual and visual signals during retrieval. In our experiments, we employ training, validation, and test splits used in previous works~\cite{lin2024preflmr,caffagni2025recurrence}.

\tit{M-BEIR Benchmark}
M-BEIR comprises eight retrieval tasks and ten different datasets, with around 1.5M human-authored queries and a pool of 5.6M candidate documents. The benchmark spans diverse sources, including everyday images, fashion products, Wikipedia entries, and news articles. In addition to standard multimodal settings, it includes tasks with missing modalities on either the query or document side, enabling evaluation under incomplete conditions.
To ensure consistency between training and testing, M-BEIR adapts datasets originally designed for different tasks, including OVEN~\cite{hu2023open}, EDIS~\cite{liu2023edis}, CIRR~\cite{liu2021image}, FashionIQ~\cite{wu2021fashion}, COCO~\cite{lin2014microsoft}, Fashion200k~\cite{han2017automatic}, Visual News~\cite{liu2020visual}, and NIGHTS~\cite{fu2023dreamsim}.
Moreover, M-BEIR defines a \textit{global} retrieval scenario, where candidates are retrieved from the full 5.6M pool encompassing all tasks and datasets, and a \textit{local} one, which restricts candidates to the task-specific pool provided by each dataset. In this paper, we report results on the $\text{M-BEIR}_{\text{local}}$ setting, for fair comparison with existing state-of-the-art retrieval models.

\tit{\rev{MMEB Benchmark}}
\rev{The Massive Multimodal Embedding Benchmark (MMEB) comprises 36 datasets spanning four meta-task categories: classification (10 tasks), visual question answering (10 tasks), multimodal retrieval (12 tasks), and visual grounding (4 tasks). In contrast to M2KR and M-BEIR, which focus on retrieval, MMEB enables a broader evaluation of multimodal embedding quality across diverse downstream tasks. We use MMEB solely for evaluation, reporting performance in a zero-shot setting.}

\tit{Evaluation Metrics} 
Following the evaluation protocol of M2KR, we assess model performance using recall@$K$ (\ie, the percentage of queries for which the target document falls within the top-$K$ most similar documents). The value of $K$ is determined based on the experimental setup of each sub-dataset. For VQA splits, we also report the pseudo recall metric, as proposed in~\cite{lin2024preflmr}, which considers a retrieved document relevant whenever it contains the answer. For M-BEIR, we adhere to the original evaluation protocol and report standard recall@$K$ values accordingly (using $K=5$ for most datasets, and $K=10$ for Fashion200k and FashionIQ). \rev{For MMEB, we report recall@$K$ with $K=1$ for all categories, which is equivalent to precision@1 used in the original benchmark.}

\begin{table*}[t]
\centering
\caption{Ablation study results on the M2KR benchmark. All experiments are with CLIP ViT-L for both visual and textual encoders.}
\setlength{\tabcolsep}{.3em}
\resizebox{\linewidth}{!}{%
\begin{tabular}{lc c cc cc cc cc cc c cc cc c cc cc}
\toprule 
& & \multicolumn{1}{c}{\textbf{WIT}} & & \multicolumn{1}{c}{\textbf{IGLUE}} & & \multicolumn{1}{c}{\textbf{KVQA}} & & \multicolumn{1}{c}{\textbf{OVEN}} & & \multicolumn{1}{c}{\textbf{LLaVA}} & & \multicolumn{2}{c} {\textbf{InfoSeek}} & & \multicolumn{2}{c}{\textbf{E-VQA}} & & \multicolumn{2}{c}{\textbf{OKVQA}} & & \\
\cmidrule{3-3} \cmidrule{5-5} \cmidrule{7-7} \cmidrule{9-9} \cmidrule{11-11} \cmidrule{13-14} \cmidrule{16-17} \cmidrule{19-20}
\textbf{Model} & & R@10 & & R@1 & & R@5 & & R@5 & & R@1 & & R@5 & PR@5 & & R@5 & PR@5 & & R@5 & PR@5 & & \textbf{Avg} \\
\midrule
ReT~\cite{caffagni2025recurrence} & & {73.4} & & 81.8 & & {63.5} & & 82.0 & & {79.9} & & {47.0} & {60.5} & & {44.5} & {57.9} & & {20.2} & 66.2 && 61.5 \\
\rowcolor{LightGray}
\hspace{0.4cm}+ global features & & 79.3 & & 81.6 & & 65.8 & & 82.8 & & 82.1 & & 46.6 & 60.8 & & 43.3 & 58.0 & & 17.4 & 64.8 && 62.0 \\
\midrule
+ score fusion (32 tokens) & & 79.5 & & 81.9 & & 66.7 & & 83.4 & & 81.0 & & 42.7 & 57.5 & & 43.4 & 57.1 & & 17.5 & 65.1 & & 61.4 \\
\midrule
+ shared architecture (32 tokens) & & 78.0 & & 83.4 & & 66.7 & & 83.5 & & 84.2 & & 48.0 & 59.9 & & 48.0 & 61.1 & & 13.3 & 62.8 & & 62.6 \\
+ shared architecture (16 tokens) & & 78.4 & & 82.2 & & 66.1 & & 83.6 & & 83.2 & & 47.5 & 60.1 & & 48.4 & 61.7 & & 13.1 & 63.6 & & 62.5 \\
+ shared architecture (8 tokens)  & & 78.1 & & 82.6 & & 62.4 & & 83.5 & & 82.7 & & 47.2 & 60.9 & & 48.4 & 61.1 & & 14.2 & 65.6 & & 62.4 \\
+ shared architecture (4 tokens) & & 78.7 & & 82.2 & & 63.5 & & 82.9 & & 82.3 & & 48.5 & 61.3 & & 47.5 & 60.8 & & 13.5 & 63.1 & & 62.2 \\
+ shared architecture (single token) & & 78.3 & & 81.9 & & 66.5 & & 84.1 & & 84.1 & & 48.2 & 60.5 & & 48.7 & 60.9 & & 12.9 & 65.0 & & 62.8 \\
\rowcolor{LightGray}
+ non-shared architecture (single token) & & 79.8 & & 82.5 & & 67.8 & & 84.4 & & 81.4 & & 46.4 & 59.0 & & 42.6 & 56.7 & & 17.1 & 65.6 && 62.1 \\
\midrule
+ layer pruning & & 77.9 & & 82.2 & & 63.3 & & 84.3 & & 82.9 & & 50.1 & 62.2 & & 47.7 & 60.7 & & 14.6 & 66.0 & & 62.9 \\
\rowcolor{ourcolor}
+ global features (\textbf{\oursnew}) & & 81.1 & & 82.9 & & 72.3 & & 83.1 & & 83.8 & & 48.0 & 61.0 & & 49.7 & 62.6 & & 15.2 & 65.9 & & 64.1 \\
\rowcolor{LightGray}
\hspace{0.4cm}\rev{- recurrent cell} & & \rev{69.5} & & \rev{74.2} & & \rev{59.9} & & \rev{75.8} & & \rev{81.9} & & \rev{37.4} & \rev{52.5} & & \rev{46.6} & \rev{59.0} & & \rev{14.1} & \rev{60.0} & & \rev{57.3} \\
\rowcolor{LightGray}
\hspace{0.4cm}\rev{- shared architecture} & & \rev{75.5} & & \rev{81.1} & & \rev{70.0} & & \rev{78.7} & & \rev{79.1} & & \rev{50.3} & \rev{62.2} & & \rev{35.8} & \rev{51.5} & & \rev{13.9} & \rev{63.5} & & \rev{60.1} \\
\rowcolor{LightGray}
\hspace{0.4cm}\rev{- score fusion (32 tokens)} & & \rev{77.9} & & \rev{81.6} & & \rev{60.9} & & \rev{82.8} & & \rev{81.9} & & \rev{46.2} & \rev{60.2} & & \rev{47.8} & \rev{59.3} & & \rev{12.9} & \rev{62.4} & & \rev{61.3} \\
\rowcolor{LightGray}
\hspace{0.4cm}\rev{- token reduction (32 tokens)} & & \rev{81.3} & & \rev{82.8} & & \rev{70.0} & & \rev{83.1} & & \rev{83.2} & & \rev{48.0} & \rev{62.4} & & \rev{50.0} & \rev{61.5} & & \rev{13.6} & \rev{63.9} & & \rev{63.6} \\
\rowcolor{LightGray}
\hspace{0.4cm}\rev{- layer pruning} & & \rev{80.7} & & \rev{81.8} & & \rev{71.1} & & \rev{82.8} & & \rev{83.5} & & \rev{48.1} & \rev{61.7} & & \rev{49.6} & \rev{61.9} & & \rev{14.8} & \rev{65.6} & & \rev{63.8} \\
\bottomrule
\end{tabular}
}
\label{tab:ablation_large}
\vspace{-0.3cm}
\end{table*}

\subsection{Implementation Details}
In our experiments, we evaluate multiple configurations of both visual and textual backbones. For the visual encoder, we consider CLIP ViT-B/32, CLIP ViT-L/14~\cite{radford2021learning}, SigLIP2 ViT-L/14~\cite{tschannen2025siglip}, and OpenCLIP ViT-H/14~\cite{cherti2023reproducible}. For the textual encoder, we use the corresponding CLIP/SigLIP variants as well as ColBERTv2~\cite{santhanam2022colbertv2}. Following the methodology described in Sec.~\ref{sec:method}, we retain only three representative layers from each backbone, corresponding to early, intermediate, and late stages. The specific layers selected for each configuration are detailed in Table~\ref{tab:n_layers}.

Our models are trained in mixed precision with the Adam optimizer on 4 NVIDIA A100 64GB GPUs for up to 24 hours. When adding global features, we always unfreeze the pooling layer of the backbones, if present. This corresponds to the visual and textual linear projections for CLIP-based and ColBERTv2 models, and to the attention pooling layers for SigLIP2. Following \ours, the recurrent cell of \oursnew operates with a hidden size $d$ equal to 1,024 and with the biases $b_i$ and $b_f$ equal to zero. The dimension of $\textbf{W}_{final}$ (cf. Eq.~\ref{eq:pre-output}) is set to match $d$ with the dimension of the global features. When unfreezing the unimodal backbones, we activate gradient checkpointing, and we downscale their learning rate by 0.05 compared to the recurrent cell for stability. At test time, we index passages using the Faiss library~\cite{johnson2019billion} for fast retrieval.

For M2KR, we use the same training recipe as \ours~\cite{caffagni2025recurrence}, setting the learning rate to $5 \times 10^{-5}$ with a cosine scheduler and a batch size of 512, training for 75k steps. We observe that training further leads to overfitting on some benchmarks, particularly severe on InfoSeek.
For M-BEIR, we train for 20 epochs with a batch size of 768, using the data sampling strategy proposed in~\cite{huang2025joint}. The learning rate is linearly ramped up to $1 \times 10^{-4}$ within the first 300 steps, and then decays accordingly to a cosine schedule.

\subsection{Ablation Studies and Analyses} 
The original \ours model employs 32 input tokens and a dedicated recurrent cell on both the query and document sides. During training, the output tokens are used to compute a fine-grained late-interaction relevance score, following~\cite{khattab2020colbert, santhanam2022colbertv2} (cf. Sec.~\ref{sec:background}). Table~\ref{tab:ablation_large} presents ablation studies supporting the architectural modifications introduced in \oursnew. Results are reported on the M2KR benchmark, using CLIP ViT-L as visual and textual backbones.

\tit{Score Fusion} We first assess the impact of replacing the fine-grained late-interaction relevance score computation with a score fusion strategy. In practice, rather than computing 32 $\times$ 32 dot products for each query-document pair, we sum the rows of the output matrix of \ours before the late-interaction projection to dimension 128, obtaining a single embedding token, typically of a size varying from 768 to 1,024, to compute the query-document similarity via dot product. 
Note that this is equivalent to substituting the $\max{}$ operator in Eq.~\ref{eq:fine-grained loss} with a new summation over $j$ (see~\cite{wei2024uniir} for more details). However, thanks to the distributive property of the dot product, we do not need to compute the $32 \times 32$ similarity matrix explicitly. This shift enables faster and more memory-efficient training, as well as quicker inference retrieval, with minimal change in performance, as the average retrieval score moves from 61.5 to 61.4 -- \ie, \textit{score fusion (32 tokens)}.

\begin{figure*}[t]
    \centering
    \includegraphics[width=0.97\linewidth]{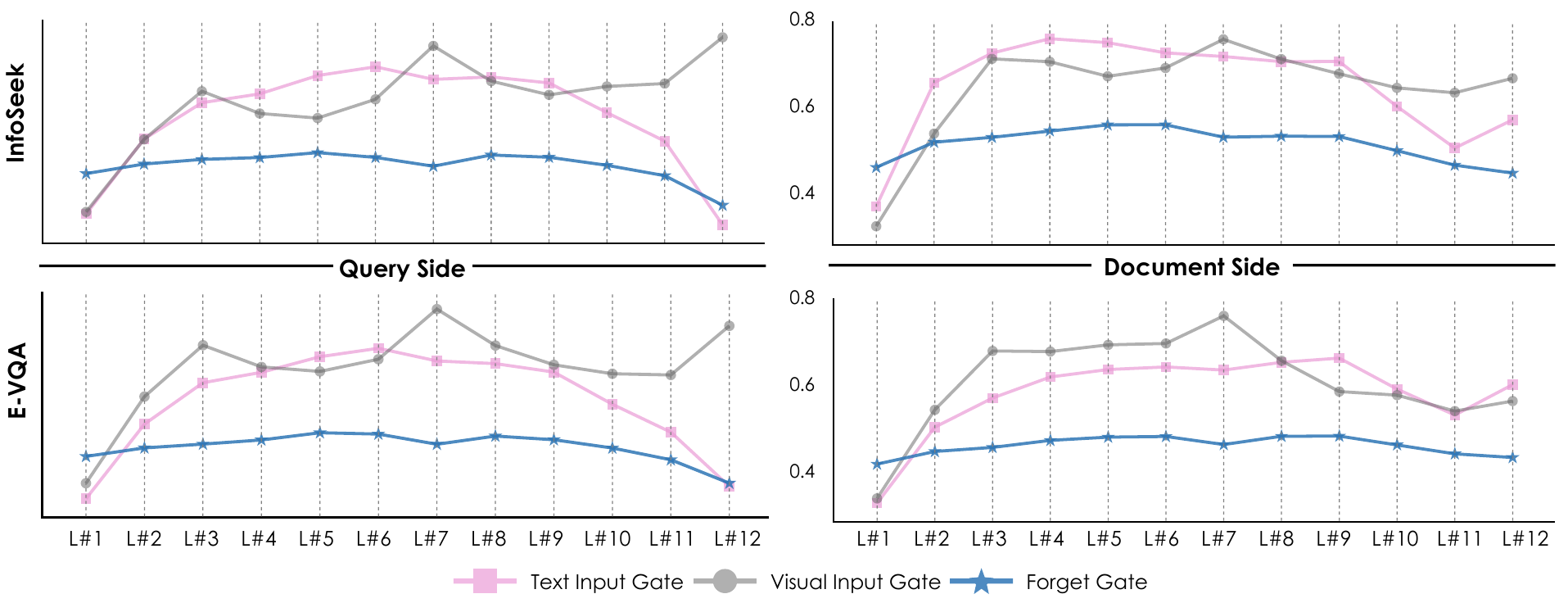}
    \vspace{-0.15cm}
    \caption{Analysis of average gate activation over 2k examples from the InfoSeek and Encyclopedic-VQA test split of the M2KR benchmark.}
    \label{fig:graph_layers}
    \vspace{-0.1cm}
\end{figure*}

\begin{table*}[t]
\centering
\caption{\rev{Performance variation across different layer configurations on the M2KR benchmark. All experiments are with CLIP ViT-L for both visual and textual encoders (\ie, $L=24$ for visual, $L=12$ for textual).}}
\setlength{\tabcolsep}{.3em}
\resizebox{\linewidth}{!}{%
\begin{tabular}{>{\color{black}}l >{\color{black}}c>{\color{black}}c >{\color{black}}c >{\color{black}}c>{\color{black}}c >{\color{black}}c>{\color{black}}c >{\color{black}}c>{\color{black}}c >{\color{black}}c>{\color{black}}c >{\color{black}}c>{\color{black}}c >{\color{black}}c >{\color{black}}c>{\color{black}}c >{\color{black}}c>{\color{black}}c >{\color{black}}c >{\color{black}}c>{\color{black}}c >{\color{black}}c>{\color{black}}c}
\toprule 
& \multicolumn{2}{c}{\rev{\textbf{Layers}}} & & \multicolumn{1}{c}{\rev{\textbf{WIT}}} & & \multicolumn{1}{c}{\rev{\textbf{IGLUE}}} & & \multicolumn{1}{c}{\rev{\textbf{KVQA}}} & & \multicolumn{1}{c}{\rev{\textbf{OVEN}}} & & \multicolumn{1}{c}{\rev{\textbf{LLaVA}}} & & \multicolumn{2}{c}{\rev{\textbf{InfoSeek}}} & & \multicolumn{2}{c}{\rev{\textbf{E-VQA}}} & & \multicolumn{2}{c}{\rev{\textbf{OKVQA}}} & & \\
\cmidrule{2-3} \cmidrule{5-5} \cmidrule{7-7} \cmidrule{9-9} \cmidrule{11-11} \cmidrule{13-13} \cmidrule{15-16} \cmidrule{18-19} \cmidrule{21-22}
\textbf{Model} & \textbf{Text} & \textbf{Vision} & & R@10 & & R@1 & & R@5 & & R@5 & & R@1 & & R@5 & PR@5 & & R@5 & PR@5 & & R@5 & PR@5 & & \textbf{Avg} \\
\midrule
ReT-2 (early only) & 3 & 3 & & 79.3 & & 82.5 & & 71.4 & & 80.9 & & 81.2 & & 49.0 & 62.0 & & 45.1 & 57.9 & & 13.3 & 63.3 & & 62.3 \\
ReT-2 (middle only) & 7 & 18 & & 80.2 & & 83.2 & & 70.4 & & 82.1 & & 80.6 & & 50.2 & 61.9 & & 48.9 & 61.3 & & 13.0 & 63.5 & & 63.2 \\
ReT-2 (late only) & 11 & 23 & & 80.7 & & 83.2 & & 71.0 & & 82.7 & & 80.7 & & 48.4 & 60.4 & & 46.7 & 59.6 & & 12.1 & 63.1 & & 62.6 \\
\midrule
ReT-2 (early+middle) & 3, 7 & 3, 18 & & 80.7 & & 82.8 & & 65.5 & & 83.2 & & 82.9 & & 49.6 & 62.5 & & 49.7 & 61.7 & & 13.9 & 65.0 & & 63.4 \\
ReT-2 (early+late)   & 3, 11 & 3, 23 & & 80.3 & & 83.2 & & 71.1 & & 83.1 & & 83.0 & & 48.0 & 62.4 & & 50.0 & 61.5 & & 14.3 & 64.9 & & 63.8 \\
ReT-2 (middle+late)  & 7, 11 & 18, 23 & & 81.2 & & 82.6 & & 72.0 & & 83.5 & & 83.2 & & 48.6 & 61.1 & & 48.5 & 61.3 & & 15.0 & 64.3 & & 63.8 \\
\midrule
\rowcolor{ourcolor}
\textbf{\oursnew} & 3, 7, 11 & 3, 18, 23 & & 81.1 & & 82.9 & & 72.3 & & 83.1 & & 83.8 & & 48.0 & 61.0 & & 49.7 & 62.6 & & 15.2 & 65.9 & & 64.1 \\
\rowcolor{LightGray}
\hspace{0.4cm}- layer pruning & all & odds & & 80.7 & & 81.8 & & 71.1 & & 82.8 & & 83.5 & & 48.1 & 61.7 & & 49.6 & 61.9 & & 14.8 & 65.6 & & 63.8 \\
\bottomrule
\end{tabular}
}
\label{tab:rebuttal_ablation_layers}
\vspace{-0.3cm}
\end{table*}

\tit{Sharing Weights} Building on the score fusion model, we experiment with sharing the weights between the query and document encoders, essentially setting $\retQ = \retD$. Apart from saving memory during training, switching to a shared architecture -- \ie, \textit{shared architecture (32 tokens)} -- raises the average score to 62.6, with an improvement of +1.2 points compared to having separate encoders. As most substantial gains come from InfoSeek and Encyclopedic-VQA, which present tens to hundreds of questions for the same Wikipedia entity, we credit the shared architecture approach for reducing overfitting on entities seen during training.

\tit{Token Reduction} The next change arises from an analysis of the output matrix of \ours, revealing that it suffers from rank collapse. Empirically, we register the rank collapse score~\cite{josephlambda} of the 32-row matrix generated by \ours when embedding samples from the InfoSeek test split of M2KR. The last recurrent step of \ours outputs 32 $\times$ 1,024 matrices. For them, we register an average rank collapse score of 0.18 when embedding queries and 0.22 when embedding documents. After applying the late-interaction linear projection to 32 $\times$ 128 dimensions, the average rank collapse scores further plummet to 0.09 and 0.11. Ideally, those scores would tend to 1.0, and our analysis indicates that the 32 token embeddings of the output matrix converge to a unified representation. Consequently, the purpose of using multiple token embeddings to represent inputs is questionable. This motivates the exploration of a token reduction strategy, by applying score fusion to a number of tokens equal to 16, 8, 4, and 1 (\ie, no score fusion at all). While reducing the number of tokens initially seems to degrade performance, we register an average improvement of +0.2 points when switching from 32 tokens to a single one -- \ie, \textit{shared architecture (single token)}. Notably, this happens along with a reduction in trainable parameters and less computation, as with a single token, there is no need to apply self-attention in the recurrent cell of \ours. For completeness, we also include the single token version of \ours without sharing weights between the query and document encoders -- \ie, \textit{non-shared architecture (single token)}. \revv{We provide a theoretical account of this rank-collapse behavior, connecting it to self-attention networks~\cite{dong2021attention} and to the role of the forget gate in Eq.~\eqref{eq:final}, in Appendix~\ref{sec:supp_theory}.}

\begin{table*}
\centering
\caption{Experimental results on the M2KR benchmark~\cite{lin2024preflmr}, comparing \oursnew to baselines and competitors when varying the visual backbone. Bold font denotes the best results under the same backbone. The $\dagger$ marker denotes our reproductions.}
\setlength{\tabcolsep}{.35em}
\resizebox{\linewidth}{!}{%
\begin{tabular}{lc c cc cc cc cc cc c cc cc c cc cc}
\toprule 
& & & \multicolumn{1}{c}{\textbf{WIT}} & & \multicolumn{1}{c}{\textbf{IGLUE}} & & \multicolumn{1}{c}{\textbf{KVQA}} & & \multicolumn{1}{c}{\textbf{OVEN}} & & \multicolumn{1}{c}{\textbf{LLaVA}} & & \multicolumn{2}{c}{\textbf{InfoSeek}} & & \multicolumn{2}{c}{\textbf{E-VQA}} & & \multicolumn{2}{c}{\textbf{OKVQA}} & & \\
\cmidrule{4-4} \cmidrule{6-6} \cmidrule{8-8} \cmidrule{10-10} \cmidrule{12-12} \cmidrule{14-15} \cmidrule{17-18} \cmidrule{20-21} 
\textbf{Model} & \textbf{Backbone} & & R@10 & & R@1 & & R@5 & & R@5 & & R@1 & & R@5 & PR@5 & & R@5 & PR@5 & & R@5 & PR@5 && \textbf{Avg} \\
\midrule
CLIP (ZS) & CLIP ViT-B & & 48.9 & & 63.1 & & \textbf{57.8} & & 58.1 & & 33.0 & & 33.6 & 47.4 & & 0.13 & 12.1 & & 0.52 & 49.9 && 36.8 \\
FLMR~\cite{lin2023fine} & CLIP ViT-B & & 23.8 & & - & & 31.9 & & 40.5 & & 56.4 & & - & 47.1 & & - & - & & - & 68.1 && -\\
PreFLMR~\cite{lin2024preflmr} & CLIP ViT-B & & 41.7 & & 57.3 & & 28.6 & & 46.3 & & 67.2 & & 26.0 & 48.8 & & \textbf{55.0} & \textbf{67.9} & & \textbf{27.2} & \textbf{66.1} && 48.4 \\
\ours~\cite{caffagni2025recurrence} & CLIP ViT-B & & 60.1 & & 73.9 & & 26.9 & & 72.9 & & 76.6 & & 30.2 & 48.1 & & 33.0 & 48.9 & & 13.9 & 58.3 & & 49.3 \\
\rowcolor{ourcolor}
\textbf{\oursnew (Ours)} & CLIP ViT-B
& & \textbf{68.3} & & \textbf{76.1} & & 56.6 & & \textbf{73.8} & & \textbf{81.2} & & \textbf{36.9} & \textbf{52.7} & & 36.1 & 52.9 & & 12.0 & 60.7 & & \textbf{55.2} \\
\rowcolor{ourcolorl}
\textbf{\oursnew (Ours)} & CLIP ViT-B {\text{\tiny \faFire}} && \textgray{\textbf{73.7}} && \textgray{\textbf{77.7}} && \textgray{\textbf{66.6}} && \textgray{\textbf{77.3}} && \textgray{\textbf{86.0}} && \textgray{\textbf{38.3}} & \textgray{\textbf{53.8}} && \textgray{42.0} & \textgray{57.6} && \textgray{14.9} & \textgray{62.6}  & & \textgray{\textbf{59.1}} \\
\midrule
CLIP (ZS) & CLIP ViT-L & & 65.9 & & 74.9 & & 73.3 & & 68.5 & & 36.6 & & 48.0 & 58.4 & & 0.17 & 12.0 & & 0.59 & 49.2 && 45.0 \\
PreFLMR~\cite{lin2024preflmr} & CLIP ViT-L & &  60.5 & & 69.2 & & 43.6 & & 59.8 & & 71.8 & & 37.4 & 57.9 & & \textbf{60.9} & \textbf{70.8} & & \textbf{31.4} & \textbf{68.5} && 57.4 \\
\ours~\cite{caffagni2025recurrence} & CLIP ViT-L & & 73.4 & & 81.8 & & 63.5 & & 82.0 & & 79.9 & & 47.0 & 60.5 & & 44.5 & 57.9 & & 20.2 & 66.2 && 61.5 \\
\rowcolor{ourcolor}
\textbf{\oursnew (Ours)} & CLIP ViT-L & & \textbf{81.1} & & \textbf{82.9} & & \textbf{72.3} & & \textbf{83.1} & & \textbf{83.8} & & \textbf{48.0} & \textbf{61.0} & & {49.7} & 62.6 & & 15.2 & 65.9 & & \textbf{64.1} \\
\rowcolor{ourcolorl}
\textbf{\oursnew (Ours)} & CLIP ViT-L {\text{\tiny \faFire}} & & \textgray{\textbf{86.1}} & & \textgray{\textbf{84.4}} & & \textgray{\textbf{78.1}} & & \textgray{\textbf{86.8}} & & \textgray{\textbf{88.6}} & & \textgray{\textbf{49.1}} & \textgray{\textbf{62.3}} & & \textgray{{56.4}} & \textgray{67.4} & & \textgray{20.0} & \textgray{67.8} & & \textgray{\textbf{67.9}} \\
\midrule
SigLIP2 (ZS) & SigLIP2 ViT-L & & 51.9 & & 60.0 & & \textbf{48.4} & & 74.3 & & 41.1 & & 51.4 & 60.4 & & 19.5 & 33.2 & & 6.1 & 50.1 && 45.1 \\
PreFLMR~\cite{lin2024preflmr}$^\dagger$ & SigLIP2 ViT-L & & 68.3 & & \textbf{76.1} & & 39.1 & & 71.5 & & 73.5 & & 42.9 & 59.5 & & \textbf{51.6} & \textbf{64.1} & & \textbf{17.8} & \textbf{70.6} && 57.7 \\
\ours~\cite{caffagni2025recurrence} & SigLIP2 ViT-L & & 65.7 & & 71.8 & & 34.8 & & 81.1 & & 75.1 & & 42.2 & 56.4 & & 35.2 & 51.2 & & 15.4 & 63.3 && 53.8 \\
\rowcolor{ourcolor}
\textbf{\oursnew (Ours)} & SigLIP2 ViT-L & & \textbf{70.3} && 71.2 && 48.2 && \textbf{85.3} && \textbf{81.8} && \textbf{57.1} & \textbf{65.5} && 44.5 & 58.1 && 10.8 & 61.5 && \textbf{59.5} \\
\rowcolor{ourcolorl}
\textbf{\oursnew (Ours)} & SigLIP2 ViT-L {\text{\tiny \faFire}} & & \textgray{\textbf{80.6}} && \textgray{\textbf{79.4}} && \textgray{\textbf{61.8}} && \textgray{\textbf{88.8}} && \textgray{\textbf{89.4}} && \textgray{\textbf{59.7}} & \textgray{\textbf{67.7}} && \textgray{\textbf{51.6}} & \textgray{63.5} && \textgray{\textbf{21.2}} & \textgray{70.5} && \textgray{\textbf{66.7}} \\
\midrule
OpenCLIP (ZS) & OpenCLIP ViT-H  & & 74.2 & & 78.2 & & \textbf{68.0} & & 78.4 & & 45.3 & & \textbf{53.2} & 61.3 & & 20.8 & 33.3 & & 7.3 & 63.9 && 53.1 \\
PreFLMR~\cite{lin2024preflmr} & OpenCLIP ViT-H & & 60.5 & & 71.2 & & 39.4 & & 61.5 & & 72.3 & & 39.2 & 59.5 & & \textbf{62.5} & \textbf{71.7} & & \textbf{30.2} & \textbf{68.1} && 57.8 \\
\ours~\cite{caffagni2025recurrence} & OpenCLIP ViT-H & & 71.4 & & 80.0 & & 59.3 & & 83.0 & & 79.8 & & 47.3 & 60.7 & & 44.8 & 57.8 & & {18.2} & 63.4 && 60.5 \\
\rowcolor{ourcolor}
\textbf{\oursnew (Ours)} & OpenCLIP ViT-H  & & \textbf{80.2} & & \textbf{82.3} & & 66.2 & & \textbf{83.3} & & \textbf{86.1} & & 52.8 & \textbf{63.1} & & 45.9 & 59.3 & & 14.4 & 64.0 & & \textbf{63.4} \\
\rowcolor{ourcolorl}
\textbf{\oursnew (Ours)}  & OpenCLIP ViT-H {\text{\tiny \faFire}} & & \textgray{\textbf{85.5}} & & \textgray{\textbf{84.2}} & & \textgray{\textbf{75.8}} & & \textgray{\textbf{88.4}} & & \textgray{\textbf{91.1}} & & \textgray{\textbf{58.0}} & \textgray{\textbf{66.7}} & & \textgray{58.9} & \textgray{69.3} & & \textgray{{18.3}} & \textgray{65.1} & & \textgray{\textbf{69.2}} \\
\bottomrule
\end{tabular}
}
\label{tab:m2kr}
\vspace{-0.3cm}
\end{table*}

\tit{Layer Pruning} Driven by the computational constraints of \ours, primarily due to the recurrent cell being applied to a predefined number of layers ranging from 12 to 16, we explore a layer pruning strategy to improve efficiency. In detail, we sample a total of three layers, corresponding to the early, middle, and late stages of both the visual and textual backbone\footnote{Because in Table~\ref{tab:ablation_large} \oursnew is paired with CLIP ViT-L/14, it follows that we employ the third, eighteenth, and second last layer from the visual backbone, and the third, seventh, and second last layer from the textual backbone. We refer to Table~\ref{tab:n_layers} for the layer selection in backbones with different depths.}. This strategy guarantees to preserve information from different abstraction levels, and it has been recently proven effective for the visual-language alignment of MLLMs~\cite{lin2025multi}. 
Our choice is further supported by an empirical analysis of the average gate activations of \ours, conducted on the InfoSeek and Encyclopedic-VQA test splits of M2KR. As shown in Fig.~\ref{fig:graph_layers}, the visual input gate exhibits three prominent activation peaks, aligning with the selected layer groups. On the other hand, the textual input gate has a smoother behavior, peaking mainly across early-to-middle stages, thus highlighting the importance of including low-level textual features. Quantitative experimental results validate the effectiveness of this pruning strategy: not only does it preserve retrieval performance, but it also yields a +0.1 points improvement in accuracy. \revv{Beyond this empirical evidence, Appendix~\ref{sec:supp_theory} provides a theoretical motivation for this design, arguing that a shorter recurrent chain is easier to retain over~\cite{bengio1994learning,pascanu2013difficulty,hochreiter1997long}, and that consecutive backbone layers, being connected by residual paths, tend to carry redundant evidence~\cite{voita2019bottom,gromov2024unreasonable,sun2024transformer}, exposing the cell's gates to correlated -- rather than complementary -- decisions when no pruning is applied.}

\tit{Global Feature Injection} Finally, we incorporate global features, obtaining our final \oursnew model. In detail, we apply score fusion by summing the multimodal, multilayer token coming from the recurrent cell with the pooler token of the visual backbone and the one from the textual backbone. This raises the average score to 64.1, with a +2.6 points improvement over \ours. For fairness of comparison, we apply global feature injection to \ours as well (\ie, gray row). In this setting, the pooler tokens are first projected to dimension 128 and then concatenated to the 32 tokens of the recurrent cell. We highlight that, even in this scenario, global features raise the performance of \ours, while still falling behind \oursnew. \rev{We also point out that the favorable impact of adding global features plays a synergistic role with the multi-layer, multimodal features generated by the recurrent cell. Indeed, if we drop the recurrent cell and only optimize the pre-trained backbones to generate better global features, the performance lowers considerably -- \ie, - \textit{recurrent cell}.}

\tit{\rev{Per-Component Contributions}}
\rev{To clarify the role of each design choice, we repeat the ablation studies presented so far, this time starting from our final configuration, removing one component at a time independently (Table~\ref{tab:ablation_large}, last four rows). First, using dedicated encoders for queries and documents (\ie, - \textit{shared architecture}) drops the overall score by -4.0 points. Second, replacing the score fusion paradigm with the original fine-grained late-interaction scheme of \ours (\ie, - \textit{score fusion (32 tokens)}) drops the average score by -2.8 points. Third, setting the number of learnable tokens back to 32 (\ie, - \textit{token reduction (32 tokens)}) yields a modest drop of 0.5 points, suggesting limited benefit from multiple tokens despite the additional computational cost. Finally, applying the recurrent cell to all layers (\ie, - \textit{layer pruning}) rather than choosing a single layer from the early, middle, and late sections of the backbones shows comparable results (-0.3 points), justifying our layer pruning design to save time and computation.}

\tit{\rev{Layer Selection}} \rev{To validate the robustness of our layer selection strategy, we additionally evaluate alternative layer configurations in Table~\ref{tab:rebuttal_ablation_layers}. In particular, we evaluate models that use only shallow (early), middle, or deep (late) layers, as well as combinations of these (\eg, early+middle, early+late, and middle+late). The results show that relying on a single depth consistently underperforms, while combining layers improves performance. Note that, using a single layer is equivalent to \textit{bypassing the recurrence at all}, as the network showed in Fig.~\ref{fig:recurrent_block} runs only once. These results confirm that the chosen layers are meaningful, providing complementary information from multiple depths of the network.
}

\smallskip
\noindent In summary, our final model, \oursnew, fuses multimodal and multi-layer features into a single learnable token, shares parameters between the query and document encoders, and incorporates global features. This design achieves superior performance without relying on the computationally expensive fine-grained contrastive loss, and is adopted as the final model for all subsequent experiments.

\begin{table*}
\centering
\caption{Experimental results on the M2KR benchmark~\cite{lin2024preflmr}, comparing \oursnew to baselines and competitors when employing ColBERTv2~\cite{santhanam2022colbertv2} as textual backbone. $\dagger$ indicates our reproductions.}
\setlength{\tabcolsep}{.32em}
\resizebox{0.92\linewidth}{!}{%
\begin{tabular}{lc c cc cc cc cc cc c cc cc c cc cc}
\toprule 
& & & \multicolumn{1}{c}{\textbf{WIT}} & & \multicolumn{1}{c}{\textbf{IGLUE}} & & \multicolumn{1}{c}{\textbf{KVQA}} & & \multicolumn{1}{c}{\textbf{OVEN}} & & \multicolumn{1}{c}{\textbf{LLaVA}} & & \multicolumn{2}{c}{\textbf{InfoSeek}} & & \multicolumn{2}{c}{\textbf{E-VQA}} & & \multicolumn{2}{c}{\textbf{OKVQA}} & & \\
\cmidrule{4-4} \cmidrule{6-6} \cmidrule{8-8} \cmidrule{10-10} \cmidrule{12-12} \cmidrule{14-15} \cmidrule{17-18} \cmidrule{20-21} 
\textbf{Model} & \textbf{Backbone} & & R@10 & & R@1 & & R@5 & & R@5 & & R@1 & & R@5 & PR@5 & & R@5 & PR@5 & & R@5 & PR@5 && \textbf{Avg} \\
\midrule
PreFLMR~\cite{lin2024preflmr} & CLIP ViT-L & &  60.5 & & 69.2 & & 43.6 & & 59.8 & & 71.8 & & 37.4 & 57.9 & & \textbf{60.9} & \textbf{70.8} & & \textbf{31.4} & \textbf{68.5} && 57.4 \\
\ours~\cite{caffagni2025recurrence} & CLIP ViT-L && 73.9 && 79.3 && 48.6 && 79.6 && 79.6 && 40.0 & 58.9 && 43.4 & 59.0 && 19.0 & 64.1 && 58.7 \\
\rowcolor{ourcolor}
\textbf{\oursnew (Ours)} & CLIP ViT-L & & \textbf{78.6} & & \textbf{80.3} & & \textbf{48.8} & & \textbf{81.2} & & \textbf{80.3} & & \textbf{50.9} & \textbf{64.9} & & 47.1 & 62.1 & & 14.8 & 62.1 && \textbf{61.0} \\
\rowcolor{ourcolorl}
\textbf{\oursnew (Ours)} & CLIP ViT-L {\text{\tiny \faFire}} & & \textgray{\textbf{81.9}} & & \textgray{\textbf{81.0}} & & \textgray{\textbf{62.9}} & & \textgray{\textbf{83.7}} & & \textgray{\textbf{84.8}} & & \textgray{\textbf{52.1}} & \textgray{\textbf{66.2}} & & \textgray{55.1} & \textgray{67.8} & & \textgray{16.7} & \textgray{64.2} & & \textgray{\textbf{65.1}} \\
\midrule
PreFLMR~\cite{lin2024preflmr}$^\dagger$ & SigLIP2 ViT-L & & 68.3 & & 76.1 & & 39.1 & & 71.5 & & 73.5 & & 42.9 & 59.5 & & \textbf{51.6} & \textbf{64.1} & & \textbf{17.8} & \textbf{70.6} && 57.7 \\
\ours~\cite{caffagni2025recurrence} & SigLIP2 ViT-L & & 65.7 & & 71.8 & & 34.8 && 81.8 & & 75.1 & & 42.2 & 56.4 & & 35.2 & 51.2 & & 15.4 & 63.3 && 53.9 \\
\rowcolor{ourcolor}
\textbf{\oursnew (Ours)} & SigLIP2 ViT-L & & \textbf{78.9} & & \textbf{79.1} & & \textbf{48.6} & & \textbf{84.4} & & \textbf{83.0} & & \textbf{53.7} & \textbf{66.3} & & 49.1 & 63.2 & & 15.2 & 61.9 && \textbf{62.1} \\
\rowcolor{ourcolorl}
\textbf{\oursnew (Ours)} & SigLIP2 ViT-L {\text{\tiny \faFire}} & & \textgray{\textbf{82.7}} & & \textgray{\textbf{82.5}} & & \textgray{\textbf{56.4}} & & \textgray{\textbf{86.6}} & & \textgray{\textbf{86.1}} & & \textgray{\textbf{59.4}} & \textgray{\textbf{68.6}} & & \textgray{\textbf{56.5}} & \textgray{\textbf{68.6}} & & \textgray{17.3} & \textgray{67.4} & & \textgray{\textbf{66.5}} \\
\bottomrule
\end{tabular}
}
\label{tab:m2kr_colbert}
\vspace{-0.3cm}
\end{table*}

\begin{table*}
\centering
\caption{\rev{Results on the M2KR benchmark when training \oursnew with the 3-stage pipeline proposed in~\cite{lin2024preflmr}. Our original single-stage version is reported in gray. All models employ CLIP ViT-L/14 as visual backbone which is kept frozen in all training stages.}}
\setlength{\tabcolsep}{.3em}
\resizebox{\linewidth}{!}{%
\begin{tabular}{>{\color{black}}l>{\color{black}}c >{\color{black}}c >{\color{black}}c >{\color{black}}c >{\color{black}}c>{\color{black}}c >{\color{black}}c>{\color{black}}c >{\color{black}}c>{\color{black}}c >{\color{black}}c>{\color{black}}c >{\color{black}}c>{\color{black}}c >{\color{black}}c >{\color{black}}c>{\color{black}}c >{\color{black}}c>{\color{black}}c >{\color{black}}c >{\color{black}}c>{\color{black}}c >{\color{black}}c>{\color{black}}c}
\toprule 
& & & & & \multicolumn{1}{c}{\rev{\textbf{WIT}}} & & \multicolumn{1}{c}{\rev{\textbf{IGLUE}}} & & \multicolumn{1}{c}{\rev{\textbf{KVQA}}} & & \multicolumn{1}{c}{\rev{\textbf{OVEN}}} & & \multicolumn{1}{c}{\rev{\textbf{LLaVA}}} & & \multicolumn{2}{c}{\rev{\textbf{InfoSeek}}} & & \multicolumn{2}{c}{\rev{\textbf{E-VQA}}} & & \multicolumn{2}{c}{\rev{\textbf{OKVQA}}} & & \\
\cmidrule{6-6} \cmidrule{8-8} \cmidrule{10-10} \cmidrule{12-12} \cmidrule{14-14} \cmidrule{16-17} \cmidrule{19-20} \cmidrule{22-23}
\textbf{Model} & \textbf{Textual Backbone} & & \textbf{GPU Hrs} & & R@10 & & R@1 & & R@5 & & R@5 & & R@1 & & R@5 & PR@5 & & R@5 & PR@5 & & R@5 & PR@5 & & \textbf{Avg} \\
\midrule
PreFLMR~\cite{lin2024preflmr} & ColBERTv2 {\text{\tiny \faFire}} & & 864 & & 60.5 & & 69.2 & & 43.6 & & 59.8 & & 71.8 & & 37.4 & 57.9 & & 60.9 & 70.8 & & \textbf{31.4} & 68.5 & & 57.4 \\
\rowcolor{ourcolor}
\textbf{\oursnew (Ours)} & ColBERTv2 {\text{\tiny \faFire}} & & 260 & & 77.9 & & 79.3 & & 57.0 & & 81.6 & & 80.9 & & \textbf{51.7} & \textbf{65.7} & & \textbf{62.0} & \textbf{71.8} & & 25.3 & 69.1 & & 65.7 \\
\rowcolor{ourcolor}
\textbf{\oursnew (Ours)} & CLIP ViT-L {\text{\tiny \faFire}} & & 212 & & \textbf{81.9} & & \textbf{83.7} & & \textbf{75.0} & & \textbf{84.4} & & \textbf{84.9} & & 50.6 & 64.6 & & 61.9 & 70.3 & & 26.4 & \textbf{72.8} & & \textbf{68.8} \\
\midrule
\rowcolor{LightGray}
\textbf{\oursnew (Ours)} & CLIP ViT-L & & 80 & & \textgray{81.1} & & \textgray{82.9} & & \textgray{72.3} & & \textgray{83.1} & & \textgray{83.8} & & \textgray{48.0} & \textgray{61.0} & & \textgray{49.7} & \textgray{62.6} & & \textgray{15.2} & \textgray{65.9} & & \textgray{64.1} \\
\bottomrule
\end{tabular}
}
\label{tab:preflmr}
\vspace{-0.4cm}
\end{table*}

\subsection{Comparison with the State of the Art}
\tinytit{Results on the M2KR Benchmark}
Table~\ref{tab:m2kr} presents a comparison of our proposed method, \oursnew, against a zero-shot CLIP baseline and other retrieval approaches. These include FLMR~\cite{lin2023fine} and PreFLMR~\cite{lin2024preflmr}, two multimodal retrieval models trained on M2KR. Both models adopt a multimodal query and a text-only document setting. FLMR relies on the \texttt{CLS} token for image representation, whereas PreFLMR enriches visual information using patch embeddings from the penultimate layer, capturing more fine-grained features. For reference, we also report results from our earlier model, \ours~\cite{caffagni2025recurrence}. We also include a variant of \oursnew in which the visual and textual backbones are unfrozen during training (\faFire).

Across all datasets and backbones, \oursnew consistently outperforms the original \ours. For example, on WIT with a frozen CLIP ViT-L backbone, \oursnew achieves a substantial gain of +7.7 points over \ours (81.1 vs. 73.4). When compared to other state-of-the-art methods, \oursnew achieves the best average performance in most settings, with the only exceptions being Encyclopedic-VQA and OKVQA, where PreFLMR slightly outperforms it. In this regard, we notice that PreFLMR employs a three-stage training pipeline, with the second stage being dedicated to Encyclopedic-VQA and the third stage entailing a careful balancing and resampling of each sub-dataset. In contrast, our \oursnew models are trained in a single-stage run on the entire M2KR dataset.
The trainable variant of \oursnew (\faFire) further boosts performance -- for instance, with the SigLIP2 backbone, the trainable version delivers an average improvement of +7.2 points. Similar trends are observed across all backbone architectures: CLIP ViT-B shows improvement from 55.2 to 59.1, CLIP ViT-L from 64.1 to 67.9, and OpenCLIP ViT-H from 63.4 to 69.2.
Finally, scaling the unfrozen visual backbone also correlates with stronger retrieval results: average performance increases from 59.1 with CLIP ViT-B, to 67.9 with CLIP ViT-L, and 69.2 with OpenCLIP ViT-H. In contrast, when the backbones are frozen, we observe a similar trend to that reported in both \ours and PreFLMR: the larger OpenCLIP ViT-H underperforms relative to the smaller CLIP ViT-L, suggesting that the benefits of scaling depend on the dataset and experimental setting. 

To provide a fairer comparison with the original PreFLMR model, which uses ColBERTv2~\cite{santhanam2022colbertv2} as its textual backbone, in Table~\ref{tab:m2kr_colbert} we also report the results obtained when replacing the textual backbone in both \ours and \oursnew with ColBERTv2. This evaluation is conducted using both CLIP and SigLIP2 ViT-L visual backbones, ensuring consistency and comparability across architectures. 
As it can be seen, the performance trends remain consistent: \oursnew continues to outperform both the original \ours and PreFLMR, even when matched on backbone architecture. The largest improvements are again observed with the trainable variant of \oursnew, yielding average gains of +6.4 and +12.6 points over \ours when using CLIP and SigLIP2 ViT-L, respectively.

\rev{Additionally, since \oursnew models are trained in a single run on all the data from M2KR, in Table~\ref{tab:preflmr} we explore the more complex training scheme proposed by PreFLMR~\cite{lin2024preflmr}. This consists of a 3-stage training: first, it begins with an extensive pre-training on the full M2KR data, followed by a fine-tuning on in-domain data from Encyclopedic-VQA~\cite{mensink2023encyclopedic}; finally, a new training is run over the full M2KR data, this time with a careful balancing of the data sampled from each dataset. In particular, data from OKVQA are up-sampled 10 times, while data from Encyclopedic-VQA is upsampled twice. Moreover, the text encoder is kept frozen during the first pre-training run, while it is free during the subsequent stages. We train variants of \oursnew following these settings, matching the total amount of training samples seen by PreFLMR at each stage. As shown, while the gain on all datasets are preserved, the fine-tuning on Encyclopedic-VQA data greatly increases the performance on this dataset, surpassing the results of PreFLMR. Similarly, the scores on OKVQA are now much closer to PreFLMR in terms of R@5, and even better concerning PR@5, probably caused by the upsample of OKVQA data in the final training stage. These results indicate that the training mixture plays a key role in optimizing multimodal retrievers. Nevertheless, our single-stage training strategy (\ie, last gray row) remains simpler, avoids dataset rebalancing, and reduces training cost, while still achieving state-of-the-art performance. Fine-tuning on in-domain data can further benefit domain-specific scenarios.}

\begin{table*}[t]
\centering
\caption{Experimental results on the $\text{M-BEIR}_{\text{local}}$ benchmark~\cite{wei2024uniir}. $\dagger$ indicates our reproductions, and gray denotes MLLM-based methods.
}
\setlength{\tabcolsep}{.24em}
\resizebox{\linewidth}{!}{%
\begin{tabular}{lccc cccccccccccccccc cc}
\toprule 
& & & & \multicolumn{3}{c}{\textbf{\#1}} & \multicolumn{1}{c}{\textbf{\#2}} & \multicolumn{2}{c}{\textbf{\#3}} & \multicolumn{3}{c}{\textbf{\#4}} & \multicolumn{1}{c}{\textbf{\#5}} & \multicolumn{2}{c}{\textbf{\#6}} & \multicolumn{2}{c}{\textbf{\#7}} & \multicolumn{2}{c}{\textbf{\#8}} \\
\cmidrule(lr){5-7} \cmidrule(lr){8-8} \cmidrule(lr){9-10} \cmidrule(lr){11-13} \cmidrule(lr){14-14} \cmidrule(lr){15-16} \cmidrule(lr){17-18} \cmidrule(lr){19-20} 
&& \textbf{Backbone} && \textbf{VN} & \textbf{COCO} & \textbf{F200k} & \textbf{WQA} & \textbf{EDIS} & \textbf{WQA} & \textbf{VN} & \textbf{COCO} & \textbf{F200k} & \textbf{NIGHTS} & \textbf{OVEN} & \textbf{InfoSeek} & \textbf{FIQ} & \textbf{CIRR} & \textbf{OVEN} & \textbf{InfoSeek} & & \textbf{Avg} \\
\midrule
CLIP (ZS) & & CLIP ViT-L && 43.4 & 61.1 & 6.6 & 36.2 & 43.3 & 45.1 & 41.3 & 79.0 & 7.7 & 26.1 & 24.2 & 20.5 & 7.0 & 13.2 & 38.8 & 26.4 & & 32.5 \\
SigLIP2 (ZS) & & SigLIP2 ViT-L && 40.0 & 77.5 & 34.8 & 33.7 & 27.3 & 42.5 & 40.4 & 88.1 & 35.3 & 28.4 & 30.0 & 30.2 & 20.4 & 29.3 & 41.9 & 34.3 & & 39.6 \\
PreFLMR~\cite{lin2024preflmr} && CLIP ViT-L && - & - & - & 68.1 & 21.8 & 37.6 & 0.1 & 8.1 & 0.0 & - & 19.9 & 21.7 & - & - & 27.4 & 23.5 & & - \\
\midrule
\ours~\cite{caffagni2025recurrence} && CLIP ViT-L && 23.2 & 66.3 & 12.3 & 47.0 & 47.1 & 56.9 & 23.0 & 85.5 & 9.5 & 21.5 & 39.0 & 21.4 & 10.6 & 27.1 & 57.3 & 33.9 & & 36.3 \\
\ours~\cite{caffagni2025recurrence} && CLIP ViT-L \text{\tiny \faFire} && 24.2 & 72.8 & 14.5 & 54.3 & 48.5 & 65.6 & 24.1 & 87.6 & 15.7 & 25.6 & 37.5 & 20.2 & 13.0 & 37.2 & 56.3 & 35.2 & & 39.5 \\
GENIUS~\cite{kim2025genius} & & CLIP ViT-L \text{\tiny \faFire} & & 27.4 & 78.0 & 16.2 & 44.6 & 44.3 & 60.6 & 28.4 & 91.1 & 16.3 & 30.2 & 41.9 & 20.7 & 19.3 & 39.5 & 52.5 & 30.1 && 40.1 \\
UniIR~\cite{wei2024uniir} && BLIP ViT-L \text{\tiny \faFire} && 23.4 & 79.7 & 26.1 & 80.0 & 50.9 & 79.8 & 22.8 & 89.9 & 28.9 & \textbf{33.0} & 41.0 & 22.4 & 29.2 & 52.2 & 55.8 & 33.0 & & 46.8 \\
UniIR~\cite{wei2024uniir} && CLIP ViT-L \text{\tiny \faFire} && 42.6 & {81.1} & 18.0 & 84.7 & 59.4 & 78.7 & 43.1 & 92.3 & 18.3 & 32.0 & 45.5 & 27.9 & 24.4 & 44.6 & 67.6 & 48.9 & & 50.6 \\
UniIR~\cite{wei2024uniir}$^\dagger$ && SigLIP2 ViT-L \text{\tiny \faFire} && 29.4 & 78.1 & 21.6 & 75.3 & 49.9 & 77.6 & 33.0 & 91.1 & 44.5 & 29.5 & 52.9 & 27.9 & 33.1 & 54.0 & 71.2 & \textbf{50.7} & & 51.2 \\
\rowcolor{ourcolor} 
\textbf{\oursnew (Ours)} && CLIP ViT-L  {\text{\tiny \faFire}} && \textbf{47.3} & 80.2 & 21.1 & \textbf{86.0} & \textbf{56.7} & \textbf{80.2} & \textbf{46.8} & 91.6 & 22.7 & 31.5 & 48.7 & 27.5 & 23.8 & 44.3 & 69.1 & 47.0 & & 51.5 \\
\rowcolor{ourcolor} 
\textbf{\oursnew (Ours)} && SigLIP2 ViT-L {\text{\tiny \faFire}} && 38.9 & \textbf{84.8} & \textbf{50.0} & 76.3 & 53.7 & 78.4 & 42.0 & \textbf{95.0} & \textbf{52.2} & 31.5 & \textbf{54.1} & \textbf{32.3} & \textbf{35.3} & \textbf{57.1} & \textbf{72.1} & 48.3 && \textbf{56.4} \\
\midrule
\midrule
\rowcolor{LightGray}
MM-Embed~\cite{lin2025mm} & & LLaVA-NeXT-7B
& & 41.0 & 71.3 & 17.1 & 95.9 & 68.8 & 85.0 & 41.3 & 90.1 & 18.4 & 32.4 & 42.1 & 42.3 & 25.7 & 50.0 & 64.1 & 57.7 & & 52.7 \\
\rowcolor{LightGray}
JFE~\cite{huang2025joint} && PaliGemma-3B && 34.6 & 78.5 & 37.2 & 88.7 & 54.3 & 82.4 & 33.1 & 90.0 & 36.9 & 27.8 & 46.0 & 35.6 & 31.8 & 54.0 
& 72.7 & 61.1 & & 54.0 \\
\rowcolor{LightGray}
PUMA~\cite{lyu2025puma} && Qwen2-VL-7B && 35.7 & 79.5 & 25.8 & 86.2 & 35.2 & 90.1 & 29.0 & 31.4 & 58.2 & 78.4 & 52.7 & 48.3 & 30.6 & 49.9 & 74.0 & 65.2 & & 54.4\\
\rowcolor{LightGray}
LamRA~\cite{liu2025lamra} & & Qwen2-VL-7B & & 41.6 & 81.5 & 28.7 & 86.0 & 62.6 & 81.2 & 39.6 & 90.6 & 30.4 & 32.1 & 54.1 & 52.1 & 33.2 & 53.1 & 76.2 & 63.3 & & 56.6 \\   
\bottomrule
\end{tabular}
}
\label{tab:mbeir}
\vspace{-0.4cm}
\end{table*}

\begin{table}[t]
\centering
\caption{\rev{Experimental results on the MMEB benchmark~\cite{jiangvlm2vec} in a zero-shot setting (no training on the MMEB split).}}
\setlength{\tabcolsep}{.35em}
\resizebox{\linewidth}{!}{%
\begin{tabular}{>{\color{black}}l >{\color{black}}c >{\color{black}}c >{\color{black}}c >{\color{black}}c >{\color{black}}c >{\color{black}}c >{\color{black}}c >{\color{black}}c >{\color{black}}c >{\color{black}}c}
\toprule
\textbf{Model} & & \textbf{Classification} & & \textbf{VQA} & & \textbf{Retrieval} & & \textbf{Grounding} & & \textbf{All} \\
\midrule
CLIP~\cite{radford2021learning}  & & 42.8 & & 9.1 & & 53.0 & & 51.8 & & 37.8 \\
BLIP-2~\cite{li2023blip}  & & 27.0 & & 4.2 & & 33.9 & & 47.0 & & 25.2 \\
MagicLens~\cite{zhang2024magiclens}  & & 38.8 & & 8.3 & & 35.4 & & 26.0 & & 27.8 \\
E5-V~\cite{jiang2024e5} & & 21.8 & & 4.9 & & 11.5 & & 19.0 & & 13.3 \\
UniIR~\cite{wei2024uniir} & & 44.3 & & 16.2 & & 61.8 & & 62.2 & & 44.3 \\
\midrule
\rowcolor{ourcolor}
\textbf{\oursnew} (CLIP) & & 44.2 & & 16.3 & & 64.3 & & 62.8 & & 45.0 \\
\rowcolor{ourcolor}
\textbf{\oursnew} (SigLIP2) & & \textbf{55.3} & & \textbf{16.7} & & \textbf{64.5} & & \textbf{59.2} & & \textbf{48.1} \\
\bottomrule
\end{tabular}
}
\label{tab:mmeb}
\vspace{-0.4cm}
\end{table}

\tit{Results on M-BEIR Benchmark}
In Table~\ref{tab:mbeir}, we further evaluate the generalization capability of our proposed approach on $\text{M-BEIR}_{\text{local}}$. 
The benchmark comprises eight distinct tasks, each presenting different modality configurations and challenges\footnote{A detailed description of each task is provided in Appendix~\ref{sec:supp_details}.}.
In this setting, we compare \oursnew with zero-shot baselines and competitors like UniIR~\cite{wei2024uniir}, GENIUS~\cite{kim2025genius}, and the previous version of our model (\ie, \ours). Specifically, UniIR proposes strategies for encoding multimodal queries and documents, by leveraging pre-trained models like CLIP and BLIP~\cite{li2022blip} to integrate different modalities. In this table, we also include our reproduction of UniIR using the SigLIP2 backbone to ensure a fair and consistent comparison. GENIUS, on the other hand, is a versatile generative retrieval framework that discretizes multimodal inputs. As additional competitors, we include retrieval models based on MLLMs, such as MM-Embed~\cite{lin2025mm}, JFE~\cite{huang2025joint}, PUMA~\cite{lyu2025puma}, and LamRA~\cite{liu2025lamra}. Due to their significantly larger model sizes and parameter counts, these methods are not directly comparable to ours. 

The results show that \oursnew, using both the CLIP and SigLIP2 ViT-L backbones, significantly outperforms not only the original \ours version but also all other competitors. For instance, the SigLIP2 variant of \oursnew achieves a notable improvement of +5.2 points over UniIR using the same backbone.
Remarkably, despite being smaller in size and not relying on an LLM, the variant of \oursnew based on SigLIP2 delivers the best overall performance compared to nearly all MLLM-based competitors, falling just short of the LamRA model, which achieves only a +0.2-points average improvement.

\tit{\rev{Results on MMEB Benchmark}} \rev{We further evaluate \oursnew on MMEB~\cite{jiangvlm2vec} which spans diverse multimodal tasks across four different categories: classification, visual question answering, retrieval, and visual grounding. We focus on \oursnew models trained on M-BEIR, as it better reflects the heterogeneous settings of MMEB. For comparison, we consider models reported in~\cite{jiangvlm2vec} that are not specifically fine-tuned on MMEB, including CLIP~\cite{radford2021learning}, BLIP-2~\cite{li2023blip}, MagicLens~\cite{zhang2024magiclens}, E5-V~\cite{jiang2024e5}, and UniIR~\cite{wei2024uniir}. Table~\ref{tab:mmeb} report the results averaged on each category. As shown, \oursnew with the CLIP backbone already outperforms all competing methods, achieving a score on the retrieval category of 64.3 against 61.8 of UniIR, which represents the strongest baseline. Switching to the SigLIP2 backbone further boosts the retrieval score to 64.5, and raises the overall average to 48.1, confirming that the gains observed on M2KR and M-BEIR generalize to this broader benchmark.}

\begin{table}[t]
\caption{Comparison of training resources and inference times between \oursnew and competing methods.}
\label{tab:training_inference}
\centering
\setlength{\tabcolsep}{.25em}
\resizebox{\linewidth}{!}{%
\begin{tabular}{l cccc c ccc cc}
\toprule 
 & \multicolumn{4}{c}{\textbf{Training Info}} & & \multicolumn{3}{c}{\textbf{Inference Time (ms)}}\\
\cmidrule(lr){2-5} \cmidrule(lr){7-9}
\textbf{Model} & \multicolumn{2}{c}{Backbones} & \#GPUs & Hrs & & Forward & Retrieval & All $\downarrow$ & & \#Tokens \\
\midrule
CLIP (ZS) & T \SnowflakeChevron & V \SnowflakeChevron & - & - & & 18.6 & 0.7 & 19.3 & & 1 \\
SigLIP2 (ZS) & T \SnowflakeChevron & V \SnowflakeChevron & - & - & & 19.2 & 0.8 & 20.0 & & 1 \\
\midrule
PreFLMR~\cite{lin2024preflmr} & T \faFire & V \SnowflakeChevron & 4 & 864 & & 32.7 & 406.1 & 438.8 & & 320 \\
UniIR~\cite{wei2024uniir} & T \faFire & V \faFire & 8 & 72 & & 23.8 & 0.8 & 24.6 & & 1  \\
LamRA~\cite{liu2025lamra} & \multicolumn{2}{c}{MLLM \faFire} & 16 & N/A & & 52.7 & 1.5 & 54.2 && 1 \\
\midrule
\ours~\cite{caffagni2025recurrence}  & T \SnowflakeChevron & V \SnowflakeChevron & 4 & 80 & & 31.4 & 3.5 & 34.9 & & 32 \\
\rowcolor{ourcolor}
\textbf{\oursnew (Ours)} & T \SnowflakeChevron & V \SnowflakeChevron & 4 & 80 & & 26.8 & 0.8 & 27.6 & & 1 \\
\rowcolor{ourcolorl}
\textbf{\oursnew (Ours)} & T \faFire & V \faFire & \textgray{4} & \textgray{160} & & \textgray{26.8} & \textgray{0.8} & \textgray{27.6} & & \textgray{1} \\
\bottomrule
\end{tabular}
}
\vspace{-.4cm}
\end{table}

\subsection{Computational Analysis}
In Table~\ref{tab:training_inference}, we provide a computational analysis of \oursnew and competitors in terms of resource demand for training and inference speed. The analysis employs a subset of the InfoSeek dataset comprising 100k image-text passages and 4.7k image-text queries. For CLIP ViT-L and SigLIP2, which we include as baselines for image-text retrieval, we mask out text on the query side and images on the document side. For \ours and PreFLMR, we follow the implementation in~\cite{khattab2020colbert} to index passages, enabling efficient fine-grained late-interaction retrieval through GPU acceleration. This implementation runs the forward pass of the models in full precision, so we stick with full precision to measure the forward time of all the models. An exception is LamRA, which we run in half precision to account for the additional memory requirements due to its 7B MLLM backbone.
For the other methods, we build a \texttt{GpuIndexFlat} using the Faiss library.  All experiments are run on a single NVIDIA A100 GPU (64GB of VRAM).

Notably, \oursnew benefits from the introduced layer pruning strategy and the use of a single input token to embed queries and documents, resulting in significantly faster forward and retrieval times compared to \ours and PreFLMR, which rely on the more computationally intensive fine-grained late-interaction paradigm.
Compared with UniIR, \oursnew demonstrates competitive retrieval speed while generally requiring equal or lower training resources, depending on whether the unimodal backbones are trained together with the recurrent retrieval cell or kept frozen.
It is worth noting that LamRA takes nearly twice the forward and retrieval time of \oursnew, not to mention the additional storage required for saving embeddings of size 3,584 rather than 768 as in our model. Ultimately, the decision to rely on MLLMs rather than smaller encoders based on vision-language foundation models is a trade-off between performance and efficiency.

\section{Experiments on Retrieval-Augmented VQA}
As a more realistic use case, we evaluate our approach for retrieval-augmented generation in knowledge-intensive VQA, where an off-the-shelf MLLM must answer visual questions requiring detailed knowledge of a specific entity (\eg, the subject of a Wikipedia page). Since such questions are often unanswerable without external knowledge, we assess the effectiveness of \oursnew in retrieving relevant context to help the MLLM answer the questions correctly.

\subsection{Datasets and Evaluation Metrics}
\tinytit{Encyclopedic-VQA Dataset~\cite{mensink2023encyclopedic}} It contains visual questions related to a Wikipedia entity. The test set counts 5,750 questions, of which 1,000 are two-hop questions, meaning that two Wikipedia pages should be retrieved sequentially, with the correct answer lying in the second one. Results are evaluated in terms of accuracy, with an answer being counted as correct if its BEM score~\cite{bulian2022tomayto} with respect to the ground-truth is higher than 0.5. The official knowledge base consists of 2M Wikipedia pages, each divided into several sections, possibly attached with an image. As we are interested in the multimodal retrieval task, we split each Wikipedia page into multiple image-text documents, with the text being the content of the section. Concerning the visual component, we have three scenarios: we select the image directly linked to the specific section; if unavailable, we fall back to the first image associated with the entire Wikipedia page, often corresponding to the first picture appearing in the web page; if neither option exists, we omit the image, obtaining a text-only document. Following this protocol, we collect 15.9M documents in total.

\tit{InfoSeek Dataset~\cite{chen2023can}} Similarly, InfoSeek entails visual questions about Wikipedia entities. The test set annotations have not been publicly released, so we report the performance on the 73,620 questions of the validation set. A question can be of type \textit{string}, \textit{numeric}, or \textit{time}, and is marked as either \textit{unseen question} or \textit{unseen entity}, based on the given question or the referring Wikipedia entity being not present in the training set. The evaluation metric is the harmonic mean between the accuracy on the unseen question and unseen entity splits, both computed with an exact matching criterion. The official knowledge base of InfoSeek contains as many as 6M Wikipedia pages. However, only a subset of them is typically used for evaluation. Thus, to be consistent with prior research, we stick with the 100k pages in the knowledge base proposed in previous works~\cite{caffagni2024wiki,cocchi2025augmenting}. Different from Encyclopedic-VQA, these pages are not divided into sections, so we follow \ours~\cite{caffagni2025recurrence} and split each page into chunks of 100 words with the format \texttt{Title: [WikiTitle]; Content: [...]}. If an image is available for a given Wikipedia page, we attach it to all of its text chunks, creating image-text documents.
This process builds up a knowledge base of 1.02M documents.

\subsection{Implementation Details}
We run experiments with two different MLLMs, namely LLaVA-MORE-8B~\cite{cocchi2025llava}, built upon LLaMA-3.1-8B~\cite{dubey2024llama}, and the more recent Qwen2.5-VL-7B~\cite{bai2025qwen2}. In all experiments, we prompt\footnote{The exact prompt used in our experiments is reported in Appendix~\ref{sec:supp_details}.} the MLLM with the text content of the top-$k$ retrieved documents, using $k$ equal to 3. Because InfoSeek relies on exact matching to evaluate answers, we prepend its prompt with 3-shot examples, one for each question type, to teach the MLLM how to format the answer.
Generation is done through beam search decoding, using a beam size of 5 for LLaVA-MORE and a beam size of 3 for Qwen2.5-VL, limiting the number of generated tokens to 20 due to memory constraints.

For retrieval, we build image-text queries with the image being taken directly from the visual question, while we prepend the question with an instruction taken from the M2KR templates~\cite{lin2024preflmr}. 
Finally, in Encyclopedic-VQA, we do not differentiate between single and two-hop questions, running a single-step retrieval even in a two-hop context.

\begin{table}[t]
\centering
\caption{\rev{VQA accuracy scores on the Encyclopedic-VQA test set and the InfoSeek validation set comparing off-the-shelf MLLMs with different retrieval models.}}
\setlength{\tabcolsep}{.2em}
\resizebox{\linewidth}{!}{
\begin{tabular}{lcc c cc c cc}
\toprule
& & & \multicolumn{2}{c}{\textbf{E-VQA}} & & \multicolumn{3}{c}{\textbf{InfoSeek}} \\
\cmidrule{4-5} \cmidrule{7-9}
 \textbf{Model} & \textbf{Retrieval Model} & & Single-Hop & All & & Un-Q & Un-E & All \\
\midrule
BLIP-2~\cite{li2023blip} & - & &  12.6 & 12.4 & & 12.7 & 12.3 & 12.5 \\
InstructBLIP~\cite{dai2023instructblip} & - & &  11.9 & 12.0 & & 8.9 & 7.4 & 8.1 \\
LLaVA-1.5-7B~\cite{liu2024improved} & - & & 16.3 & 16.9 & & 9.6 & 9.4 & 9.5 \\
\midrule
LLaVA-MORE-8B~\cite{cocchi2025llava} & - & & 13.8 & 14.9 & & 8.9 & 8.0 & 8.4 \\
LLaVA-MORE-8B~\cite{cocchi2025llava} & CLIP ViT-L & & 17.9 & 19.0 & & 14.5 & 13.6 & 14.1 \\
LLaVA-MORE-8B~\cite{cocchi2025llava} & SigLIP2 ViT-L & & 17.5 & 18.6 & & 16.0 & 15.1 & 15.5 \\
LLaVA-MORE-8B~\cite{cocchi2025llava} & PreFLMR~\cite{lin2024preflmr} & & 27.8 & 26.9 & & 13.0 & 11.7 & 12.3 \\
LLaVA-MORE-8B~\cite{cocchi2025llava} & \ours~\cite{caffagni2025recurrence} & & 21.9 & 21.8 & & 21.1 & 15.0 & 17.5 \\
LLaVA-MORE-8B~\cite{cocchi2025llava} & UniIR~\cite{wei2024uniir} & & 16.9 & 18.2 & & \textbf{25.1} & 18.8 & 21.5 \\
\rowcolor{ourcolor}
LLaVA-MORE-8B~\cite{cocchi2025llava} & \textbf{\oursnew (Ours)} & & \textbf{28.5} & \textbf{27.1} & & 24.3 & \textbf{21.5} & \textbf{22.8} \\
\midrule
Qwen2.5-VL-7B~\cite{bai2025qwen2} & - & & 19.8 & 19.7 & & 18.6 & 18.1 & 18.3 \\
Qwen2.5-VL-7B~\cite{bai2025qwen2} & CLIP ViT-L & & 19.5 & 20.4 & & 18.7 & 17.9 & 18.3 \\
Qwen2.5-VL-7B~\cite{bai2025qwen2} & SigLIP2 ViT-L & & 20.1 & 20.9 & & 19.8 & 19.5 & 19.7 \\
Qwen2.5-VL-7B~\cite{bai2025qwen2} & PreFLMR~\cite{lin2024preflmr} & & \textbf{34.4} & \textbf{33.0} & & 18.0 & 15.8 & 16.8 \\
Qwen2.5-VL-7B~\cite{bai2025qwen2} & \ours~\cite{caffagni2025recurrence} & & 26.6 & 26.2 & & 24.5 & 17.9 & 20.7 \\
Qwen2.5-VL-7B~\cite{bai2025qwen2} & UniIR~\cite{wei2024uniir} & & 18.6 & 19.2 & & \textbf{29.0} & 22.4 & 25.3 \\
\rowcolor{ourcolor}
Qwen2.5-VL-7B~\cite{bai2025qwen2} & \textbf{\oursnew (Ours)} & & 33.5 & 31.6 & & 27.9 & \textbf{25.1} & \textbf{26.4} \\
\bottomrule
\end{tabular}
  }
\label{tab:rag}
\vspace{-0.2cm}
\end{table}

\subsection{Experimental Results}
For these experiments, we compare \oursnew against other multimodal retrievers, such as \ours, UniIR, and PreFLMR, and also include the results of the original CLIP and SigLIP2 models as baselines. Note that for all considered models, we employ their best-performing configurations in this setting. \rev{The experimental results are reported in Table~\ref{tab:rag}, which also includes off-the-shelf MLLMs without retrieval as references, namely BLIP-2~\cite{li2023blip}, InstructBLIP~\cite{dai2023instructblip}, and LLaVA-1.5-7B~\cite{liu2024improved}, as well as the two MLLMs used in our experiments (\ie, LLaVA-MORE-8B~\cite{cocchi2025llava} and Qwen2.5-VL-7B~\cite{bai2025qwen2}).}

\begin{table}[t]
\centering
\caption{\rev{VQA accuracy scores on the Encyclopedic-VQA test set and InfoSeek validation set comparing task-specific models equipped with different retrieval modules.}}
\setlength{\tabcolsep}{.22em}
\resizebox{\linewidth}{!}{
\begin{tabular}{>{\color{black}}l >{\color{black}}c>{\color{black}}c >{\color{black}}c >{\color{black}}c>{\color{black}}c >{\color{black}}c >{\color{black}}c>{\color{black}}c>{\color{black}}c}
\toprule
& \multicolumn{2}{c}{\rev{\textbf{Retrieval}}} & & \multicolumn{2}{c}{\rev{\textbf{E-VQA}}} & & \multicolumn{3}{c}{\rev{\textbf{InfoSeek}}} \\
\cmidrule{2-3} \cmidrule{5-6} \cmidrule{8-10}
\textbf{Model} & \textbf{Model} & \textbf{Size} & & Single-Hop & All & & Un-Q & Un-E & All \\
\midrule
Wiki-LLaVA-7B~\cite{caffagni2024wiki} & CLIP+Contr. & 0.6B & & 17.7 & 20.3 & & 30.1 & 27.8 & 28.9 \\
EchoSight-8B~\cite{yan2024echosight} & EVA-CLIP & 8B & & 26.4 & 24.9 & & 30.0 & 30.7 & 30.4 \\
VLM-PRF~\cite{hong2025knowledge} & EVA-CLIP & 8B & & 40.1 & 39.2 & & 43.5 & 42.1 & 42.5 \\
\midrule
\rowcolor{LightGray}
ReflectiVA-3B~\cite{cocchi2025augmenting} & EVA-CLIP & 8B & & \textgray{33.7} & \textgray{35.2} & & \textgray{39.6} & \textgray{38.1} & \textgray{38.9} \\
& PreFLMR~\cite{lin2024preflmr} & 0.5B & & 28.5 & 31.6 & & 32.7 & 29.0 & 30.8 \\
& UniIR~\cite{wei2024uniir} & 0.4B & & 36.9 & 38.3 & & \textbf{36.9} & 32.1 & 34.3 \\
\rowcolor{ourcolor}
& \textbf{\oursnew (Ours)} & 0.5B & & \textbf{38.7} & \textbf{40.1} & & 36.5 & \textbf{33.5} & \textbf{34.9} \\
\midrule
\rowcolor{LightGray}
ReAG-3B~\cite{compagnoni2025reag} & EVA-CLIP & 8B & & \textgray{41.3} & \textgray{42.9} & & \textgray{43.7} & \textgray{42.9} & \textgray{43.3} \\
& PreFLMR~\cite{lin2024preflmr} & 0.5B & & 28.5 & 31.4 & & 34.0 & 30.7 & 32.3 \\
& UniIR~\cite{wei2024uniir} & 0.4B & & 38.0 & 39.2 & & 41.3 & 36.8 & 38.9 \\
\rowcolor{ourcolor}
& \textbf{\oursnew (Ours)} & 0.5B & & \textbf{38.7} & \textbf{40.0} & & \textbf{43.2} & \textbf{39.9} & \textbf{41.5} \\
\midrule
\rowcolor{LightGray}
ReflectiVA-7B~\cite{cocchi2025augmenting} & EVA-CLIP & 8B & & \textgray{36.8} & \textgray{36.8} & & \textgray{43.5} & \textgray{44.3} & \textgray{43.9} \\
& PreFLMR~\cite{lin2024preflmr} & 0.5B & & 28.7 & 31.5 & & 35.4 & 32.9 & 34.1 \\
& UniIR~\cite{wei2024uniir} & 0.4B & & 36.4 & \textbf{38.9} & & 37.8 & 36.7 & 37.2 \\
\rowcolor{ourcolor}
& \textbf{\oursnew (Ours)} & 0.5B & & \textbf{36.5} & 38.2 & & \textbf{38.5} & \textbf{37.6} & \textbf{38.0} \\
\midrule
\rowcolor{LightGray}
ReAG-7B~\cite{compagnoni2025reag} & EVA-CLIP & 8B & & \textgray{44.9} & \textgray{47.0} & & \textgray{48.3} & \textgray{46.2} & \textgray{47.2} \\
& PreFLMR~\cite{lin2024preflmr} & 0.5B & & 28.4 & 31.6 & & 39.8 & 35.4 & 37.5 \\
& UniIR~\cite{wei2024uniir} & 0.4B & & 42.6 & 44.1 & & 46.1 & 40.6 & 43.2 \\
\rowcolor{ourcolor}
& \textbf{\oursnew (Ours)} & 0.5B & & \textbf{42.9} & \textbf{44.7} & & \textbf{47.6} & \textbf{43.5} & \textbf{45.5} \\
\bottomrule
\end{tabular}
}
\label{tab:rag_2}
\vspace{-0.2cm}
\end{table}

First of all, we observe that in general, both LLaVA-MORE and Qwen2.5-VL benefit from retrieval-augmented generation, demonstrating the challenge posed by Encyclopedic-VQA and InfoSeek. When LLaVA-MORE is used as the generator, \oursnew stands out as the best multimodal retriever across both benchmarks, even outscoring PreFLMR on Encyclopedic-VQA, despite PreFLMR having undergone a dedicated training stage on that dataset. This suggests that at a large scale, the fine-grained late-interaction mechanism may be exposed to the size of the knowledge base more severely than single-token retrieval. Switching to Qwen2.5-VL, the results are better than LLaVA-MORE, testifying the superior capabilities of this more recent MLLM. In this context, \ours falls slightly behind PreFLMR on Encyclopedic-VQA, but compensates for that by confirming itself as the best retriever on InfoSeek, scoring 9.6 points higher than PreFLMR, which even underperforms compared to Qwen2.5-VL without retrieval. Overall, these results confirm the effectiveness of our approach, showing that off-the-shelf MLLMs can achieve competitive performance in knowledge-intensive VQA without task-specific fine-tuning.

\subsection{\rev{\oursnew as Retriever for Task-Specific Methods}} \rev{In the experiments of Table~\ref{tab:rag} on general-purpose MLLMs, the considered retrieval models are specifically designed to directly provide the MLLM with a chunk of text from the knowledge base. In contrast, state-of-the-art methods for knowledge-intensive VQA~\cite{yan2024echosight,cocchi2025augmenting,compagnoni2025reag,caffagni2024wiki} typically involve fine-tuning the MLLM and, in many cases, a two-stage retrieval pipeline. In this setting, the first stage retrieves multimodal candidate documents, while the second stage refines the selection by extracting the most relevant textual passages. As a result, the role of the retriever is primarily to identify the entity referred to by the multimodal query. This entity is usually associated with a document (\eg, a full Wikipedia page), which is then processed by an additional module for filtering and refinement. For instance, ReflectiVA~\cite{cocchi2025augmenting} employs the MLLM itself to select relevant passages from the retrieved document, whereas ReAG~\cite{compagnoni2025reag} introduces a secondary LLM (\ie, the critic) to filter irrelevant content. Moreover, these models are typically fine-tuned on the target datasets to better exploit the retrieved knowledge.}

\rev{Accordingly, while previous experiments evaluate retrievers based on their ability to return \textit{fine-grained} passages likely to contain the answer, here we instead assess their effectiveness as \textit{coarse} search engines, retrieving broadly relevant documents for subsequent refinement by downstream components. Table~\ref{tab:rag_2} reports the results of this analysis, where we compare \oursnew, PreFLMR, and UniIR as coarse retrieval engines within two state-of-the-art pipelines, ReflectiVA and ReAG, both based on Qwen2.5-VL~\cite{bai2025qwen2} at the 3B and 7B scales. We follow the original implementation settings and knowledge bases for both Encyclopedic-VQA (2M Wikipedia pages) and InfoSeek (100k Wikipedia pages). For each query, the top-$k$ Wikipedia pages are retrieved based on the page summary and the associated primary image, when available (with $k=20$ following~\cite{compagnoni2025reag}). As a reference, we report the performance of the original models equipped with EVA-CLIP 8B as retriever, which is nearly~16$\times$ larger than \oursnew in terms of parameters and~requires approximately 3.5$\times$ more storage for embeddings. Under this setting, \oursnew consistently outperforms PreFLMR and UniIR across both pipelines and model scales, achieving performance comparable to significantly larger retrieval backbones and further demonstrating its effectiveness in downstream retrieval scenarios.}

\section{Conclusion}
\label{sec:conclusion}
In this work, we introduced \oursnew, a recurrent Transformer-based retrieval model that unifies multimodal queries and documents within a single framework. By combining multi-layer visual and textual representations through a gated recurrent cell, \oursnew achieves robust retrieval performance across diverse multimodal settings, as shown on the M2KR and M-BEIR benchmarks. Furthermore, it proves to be a powerful retrieval component for retrieval-augmented generation, enabling off-the-shelf MLLMs to achieve superior accuracy on knowledge-intensive VQA tasks. Our analysis also shows that \oursnew offers not only accuracy improvements but also significant efficiency gains, with faster inference and reduced memory usage compared to existing methods. Overall, we believe that leveraging multi-layer features with recurrent integration offers a promising direction toward more scalable, robust, and practical multimodal retrieval systems.

\section*{Acknowledgments}
This work has been partially supported by the EU Horizon project ELLIOT (GA No. 101214398) and by the EuroHPC JU projects MINERVA (GA No. 101182737) and IT4LIA (GA No. 101234224). We also acknowledge the CINECA award under the ISCRA initiative, for the availability of high-performance computing resources.

\bibliographystyle{IEEEtran}
\bibliography{bibliography}

@string{nips      = {Advances in Neural Information Processing Systems}}

@string{iccv      = {Proceedings of the IEEE/CVF International Conference on Computer Vision}}

@string{eccv      = {Proceedings of the European Conference on Computer Vision}}

@string{cvpr      = {Proceedings of the IEEE/CVF Conference on Computer Vision and Pattern Recognition}}

@string{cvprw     = {Proceedings of the IEEE/CVF Conference on Computer Vision and Pattern Recognition Workshops}}

@string{iclr      = {Proceedings of the International Conference on Learning Representations}}

@string{icml      = {Proceedings of the International Conference on Machine Learning}}

@string{acmmm     = {Proceedings of the ACM International Conference on Multimedia}}

@string{mir       = {Proceedings of the ACM International Workshop on Multimedia Information Retrieval}}

@string{aaai      = {Proceedings of the Conference on Artificial Intelligence}}

@string{emnlp     = {Proceedings of the Conference on Empirical Methods in Natural Language Processing}}

@string{acl       = {Proceedings of the Annual Meeting of the Association for Computational Linguistics}}

@string{naacl    = {Proceedings of the Conference of the North American Chapter of the Association for Computational Linguistics: Human Language Technologies}}

@STRING{ieeetpami = {IEEE Transactions on Pattern Analysis and Machine Intelligence}}

@string{cvpr     = {CVPR}}

@string{cvprw    = {CVPR Workshops}}

@string{nips     = {NeurIPS}}

@string{iccv     = {ICCV}}

@string{iccvw    = {ICCV Workshops}}

@string{eccv     = {ECCV}}

@string{iclr     = {ICLR}}

@string{acmmm    = {ACM Multimedia}}

@string{icml     = {ICML}}

@string{aaai     = {AAAI}}

@string{emnlp    = {EMNLP}}

@string{naacl    = {NAACL}}

@string{acl      = {ACL}}

@string{sigir     = {ACM SIGIR}}

@string{ieeetpami  = {IEEE Trans. PAMI}}

@inproceedings{jia2021scaling,
  title={{Scaling up visual and vision-language representation learning with noisy text supervision}},
  author={Jia, Chao and Yang, Yinfei and Xia, Ye and Chen, Yi-Ting and Parekh, Zarana and Pham, Hieu and Le, Quoc and Sung, Yun-Hsuan and Li, Zhen and Duerig, Tom},
  booktitle=icml,
  year={2021},
}

@article{huang2025joint,
  title={{Joint Fusion and Encoding: Advancing Multimodal Retrieval from the Ground Up}},
  author={Huang, Lang and Wu, Qiyu and Miao, Zhongtao and Yamasaki, Toshihiko},
  journal={arXiv preprint arXiv:2502.20008},
  year={2025}
}

@inproceedings{zhang2024long,
  title={{Long-CLIP: Unlocking the Long-Text Capability of CLIP}},
  author={Zhang, Beichen and Zhang, Pan and Dong, Xiaoyi and Zang, Yuhang and Wang, Jiaqi},
  booktitle=eccv,
  year={2024},
}

@inproceedings{miech2021thinking,
  title={{Thinking Fast and Slow: Efficient Text-to-Visual Retrieval With Transformers}},
  author={Miech, Antoine and Alayrac, Jean-Baptiste and Laptev, Ivan and Sivic, Josef and Zisserman, Andrew},
  booktitle=cvpr,
  year={2021}
}

@inproceedings{brown2020smooth,
  title={{Smooth-AP: Smoothing the Path Towards Large-Scale Image Retrieval}},
  author={Brown, Andrew and Xie, Weidi and Kalogeiton, Vicky and Zisserman, Andrew},
  booktitle=eccv,
  year={2020},
}

@inproceedings{baldrati2022conditioned,
  title={{Conditioned and Composed Image Retrieval Combining and Partially Fine-Tuning CLIP-Based Features}},
  author={Baldrati, Alberto and Bertini, Marco and Uricchio, Tiberio and Del Bimbo, Alberto},
  booktitle=cvpr,
  year={2022}
}

@inproceedings{lin2025multi,
  title={{Multi-Layer Visual Feature Fusion in Multimodal LLMs: Methods, Analysis, and Best Practices}},
  author={Lin, Junyan and Chen, Haoran and Fan, Yue and Fan, Yingqi and Jin, Xin and Su, Hui and Fu, Jinlan and Shen, Xiaoyu},
  booktitle=cvpr,
  year={2025}
}

@article{lyu2025puma,
  title={{PUMA: Layer-Pruned Language Model for Efficient Unified Multimodal Retrieval with Modality-Adaptive Learning}},
  author={Lyu, Yibo and Shao, Rui and Chen, Gongwei and Zhu, Yijie and Guan, Weili and Nie, Liqiang},
  journal=acmmm,
  year={2025}
}

@inproceedings{plummer2015flickr30k,
  title={{Flickr30k Entities: Collecting Region-to-Phrase Correspondences for Richer Image-to-Sentence Models}},
  author={Plummer, Bryan A and Wang, Liwei and Cervantes, Chris M and Caicedo, Juan C and Hockenmaier, Julia and Lazebnik, Svetlana},
  booktitle=iccv,
  year={2015}
}

@inproceedings{cocchi2025llava,
  title={{LLaVA-MORE: A Comparative Study of LLMs and Visual Backbones for Enhanced Visual Instruction Tuning}},
  author={Cocchi, Federico and Moratelli, Nicholas and Caffagni, Davide and Sarto, Sara and Baraldi, Lorenzo and Cornia, Marcella and Cucchiara, Rita},
  booktitle=iccvw,
  year={2025}
}

@inproceedings{lin2014microsoft,
Author = {Lin, Tsung-Yi and Maire, Michael and Belongie, Serge and Hays, James and Perona, Pietro and Ramanan, Deva and Doll{\'a}r, Piotr and Zitnick, C Lawrence},
Booktitle = eccv,
Title = {{Microsoft COCO: Common Objects in Context}},
Year = {2014}
}

@article{dubey2024llama,
  title={{The Llama 3 Herd of Models}},
  author={Dubey, Abhimanyu and Jauhri, Abhinav and Pandey, Abhinav and Kadian, Abhishek and Al-Dahle, Ahmad and Letman, Aiesha and Mathur, Akhil and Schelten, Alan and Yang, Amy and Fan, Angela and others},
  journal={arXiv preprint arXiv:2407.21783},
  year={2024}
}

@inproceedings{lin2024preflmr,
  title={{PreFLMR: Scaling Up Fine-Grained Late-Interaction Multi-modal Retrievers}},
  author={Lin, Weizhe and Mei, Jingbiao and Chen, Jinghong and Byrne, Bill},
  booktitle=acl,
  year={2024}
}

@inproceedings{lin2023fine,
  title={{Fine-grained Late-interaction Multi-modal Retrieval for Retrieval Augmented Visual Question Answering}},
  author={Lin, Weizhe and Chen, Jinghong and Mei, Jingbiao and Coca, Alexandru and Byrne, Bill},
  booktitle=nips,
  year={2023}
}

@inproceedings{wei2024uniir,
  title={{UniIR: Training and Benchmarking Universal Multimodal Information Retrievers}},
  author={Wei, Cong and Chen, Yang and Chen, Haonan and Hu, Hexiang and Zhang, Ge and Fu, Jie and Ritter, Alan and Chen, Wenhu},
  booktitle=eccv,
  year={2024}
}

@inproceedings{zhai2023sigmoid,
  title={{Sigmoid Loss for Language Image Pre-Training}},
  author={Zhai, Xiaohua and Mustafa, Basil and Kolesnikov, Alexander and Beyer, Lucas},
  booktitle=iccv,
  year={2023}
}

@inproceedings{li2023blip,
  title={{BLIP-2: Bootstrapping Language-Image Pre-training with Frozen Image Encoders and Large Language Models}},
  author={Li, Junnan and Li, Dongxu and Savarese, Silvio and Hoi, Steven},
  booktitle=icml,
  year={2023},
}

@inproceedings{chen2023can,
  title={{Can Pre-trained Vision and Language Models Answer Visual Information-Seeking Questions?}},
  author={Chen, Yang and Hu, Hexiang and Luan, Yi and Sun, Haitian and Changpinyo, Soravit and Ritter, Alan and Chang, Ming-Wei},
  booktitle=emnlp,
  year={2023}
}

@inproceedings{hu2023open,
  title={{Open-domain Visual Entity Recognition: Towards Recognizing Millions of Wikipedia Entities}},
  author={Hu, Hexiang and Luan, Yi and Chen, Yang and Khandelwal, Urvashi and Joshi, Mandar and Lee, Kenton and others},
  booktitle=cvpr,
  year={2023}
}

@inproceedings{mensink2023encyclopedic,
  title={{Encyclopedic VQA: Visual questions about detailed properties of fine-grained categories}},
  author={Mensink, Thomas and Uijlings, Jasper and Castrejon, Lluis and Goel, Arushi and Cadar, Felipe and Zhou, Howard and Sha, Fei and Araujo, Andr{\'e} and Ferrari, Vittorio},
  booktitle=iccv,
  year={2023}
}

@article{johnson2019billion,
  title={{Billion-Scale Similarity Search with GPUs}},
  author={Johnson, Jeff and Douze, Matthijs and J{\'e}gou, Herv{\'e}},
  journal={IEEE Trans. on Big Data},
  volume={7},
  number={3},
  pages={535--547},
  year={2019},
  publisher={IEEE}
}

@inproceedings{li2022blip,
  title={{BLIP: Bootstrapping Language-Image Pre-training for Unified Vision-Language Understanding and Generation}},
  author={Li, Junnan and Li, Dongxu and Xiong, Caiming and Hoi, Steven},
  booktitle=icml,
  year={2022},
}

@inproceedings{bapna2018best,
  title={{The Best of Both Worlds: Combining Recent Advances in Neural Machine Translation}},
  author={Bapna, Ankur and Foster, George and Jones, Llion and Hughes, Macduff and Johnson, Melvin and Chen, Mia and Schuster, Mike and Parmar, Niki J and others},
  booktitle=acl,
  year={2018}
}

@inproceedings{lei2017simple,
  title={{Simple recurrent units for highly parallelizable recurrence}},
  author={Lei, Tao and Zhang, Yu and Wang, Sida I and Dai, Hui and Artzi, Yoav},
  booktitle=emnlp,
  year={2017}
}

@inproceedings{lei2021attention,
  title={{When attention meets fast recurrence: Training language models with reduced compute}},
  author={Lei, Tao},
  booktitle=emnlp,
  year={2021}
}

@article{wang2019r,
  title={{R-Transformer: Recurrent Neural Network Enhanced Transformer}},
  author={Wang, Zhiwei and Ma, Yao and Liu, Zitao and Tang, Jiliang},
  journal={arXiv preprint arXiv:1907.05572},
  year={2019}
}

@inproceedings{hutchins2022block,
  title={{Block-Recurrent Transformers}},
  author={Hutchins, DeLesley and Schlag, Imanol and Wu, Yuhuai and Dyer, Ethan and Neyshabur, Behnam},
  booktitle=nips,
  year={2022}
}

@inproceedings{marino2019ok,
  title={{OK-VQA: A Visual Question Answering Benchmark Requiring
External Knowledge}},
  author={Marino, Kenneth and Rastegari, Mohammad and Farhadi, Ali and Mottaghi, Roozbeh},
  booktitle=cvpr,
  year={2019}
}

@inproceedings{khattab2020colbert,
  title={{ColBERT: Efficient and Effective Passage Search via Contextualized Late Interaction over BERT}},
  author={Khattab, Omar and Zaharia, Matei},
  booktitle=sigir,
  year={2020}
}

@inproceedings{santhanam2022colbertv2,
  title={{ColBERTv2: Effective and Efficient Retrieval via Lightweight Late Interaction}},
  author={Santhanam, Keshav and Khattab, Omar and Saad-Falcon, Jon and Potts, Christopher and Zaharia, Matei},
  booktitle=naacl,
  year={2022}
}

@inproceedings{radford2021learning,
  title={{Learning Transferable Visual Models From Natural Language Supervision}},
  author={Radford, Alec and Kim, Jong Wook and Hallacy, Chris and Ramesh, Aditya and Goh, Gabriel and Agarwal, Sandhini and Sastry, Girish and Askell, Amanda and Mishkin, Pamela and Clark, Jack and Krueger, Gretchen and Sutskever, Ilya},
  booktitle=icml,
  year={2021},
}

@inproceedings{vaswani2017attention,
  title={{Attention Is All You Need}},
  author={Vaswani, Ashish and Shazeer, Noam and Parmar, Niki and Uszkoreit, Jakob and Jones, Llion and Gomez, Aidan N and Kaiser, {\L}ukasz and Polosukhin, Illia},
  booktitle=nips,
  year={2017}
}

@article{hochreiter1997long,
  title={{Long Short-Term Memory}},
  author={Hochreiter, Sepp and Schmidhuber, J{\"u}rgen},
  journal={Neural Computation},
  volume={9},
  year={1997},
}

@inproceedings{dosovitskiy2021image,
  title={{An Image is Worth 16x16 Words: Transformers for Image Recognition at Scale}},
  author={Dosovitskiy, Alexey and Beyer, Lucas and Kolesnikov, Alexander and Weissenborn, Dirk and Zhai, Xiaohua and Unterthiner, Thomas and Dehghani, Mostafa and Minderer, Matthias and Heigold, Georg and Gelly, Sylvain and Uszkoreit, Jakob and Houlsby, Neil},
  booktitle=iclr,
  year={2021}
}

@inproceedings{touvron2021training,
  title={Training data-efficient image transformers \& distillation through attention},
  author={Touvron, Hugo and Cord, Matthieu and Douze, Matthijs and Massa, Francisco and Sablayrolles, Alexandre and J{\'e}gou, Herv{\'e}},
  booktitle=icml,
  year={2021},
}

@inproceedings{brown2020language,
  title={{Language models are few-shot learners}},
  author={Brown, Tom and Mann, Benjamin and Ryder, Nick and Subbiah, Melanie and Kaplan, Jared D and Dhariwal, Prafulla and Neelakantan, Arvind and Shyam, Pranav and Sastry, Girish and Askell, Amanda and others},
  booktitle=nips,
  year={2020}
}

@article{touvron2023llama,
  title={{LLaMA: Open and Efficient Foundation Language Models}},
  author={Touvron, Hugo and Lavril, Thibaut and Izacard, Gautier and Martinet, Xavier and Lachaux, Marie-Anne and Lacroix, Timoth{\'e}e and Rozi{\`e}re, Baptiste and Goyal, Naman and Hambro, Eric and Azhar, Faisal and others},
  journal={arXiv preprint arXiv:2302.13971},
  year={2023}
}

@inproceedings{cherti2023reproducible,
  title={{Reproducible scaling laws for contrastive language-image Learning}},
  author={Cherti, Mehdi and Beaumont, Romain and Wightman, Ross and Wortsman, Mitchell and Ilharco, Gabriel and Gordon, Cade and Schuhmann, Christoph and Schmidt, Ludwig and Jitsev, Jenia},
  booktitle=cvpr,
  year={2023}
}

@inproceedings{wolf2020transformers,
  title={{Transformers: State-of-the-Art Natural Language Processing}},
  author={Wolf, Thomas and Debut, Lysandre and Sanh, Victor and Chaumond, Julien and Delangue, Clement and Moi, Anthony and Cistac, Pierric and Rault, Tim and Louf, R{\'e}mi and Funtowicz, Morgan and others},
  booktitle=emnlp,
  year={2020}
}

@inproceedings{zhai2022scaling,
  title={{Scaling Vision Transformers}},
  author={Zhai, Xiaohua and Kolesnikov, Alexander and Houlsby, Neil and Beyer, Lucas},
  booktitle=cvpr,
  year={2022}
}

@article{khan2022transformers,
  title={{Transformers in Vision: A Survey}},
  author={Khan, Salman and Naseer, Muzammal and Hayat, Munawar and Zamir, Syed Waqas and Khan, Fahad Shahbaz and Shah, Mubarak},
  journal={ACM CSUR},
  volume={54},
  number={10s},
  pages={1--41},
  year={2022},
}

@inproceedings{srinivasan2021wit,
  title={{WIT: Wikipedia-based Image Text Dataset for Multimodal Multilingual Machine Learning}},
  author={Srinivasan, Krishna and Raman, Karthik and Chen, Jiecao and Bendersky, Michael and Najork, Marc},
  booktitle=sigir,
  year={2021}
}

@inproceedings{sharma2018conceptual,
  title={{Conceptual Captions: A Cleaned, Hypernymed, Image Alt-text Dataset For Automatic Image Captioning}},
  author={Sharma, Piyush and Ding, Nan and Goodman, Sebastian and Soricut, Radu},
  booktitle=acl,
  year={2018}
}

@inproceedings{bugliarello2022iglue,
  title={{IGLUE: A Benchmark for Transfer Learning Across Modalities, Tasks, and Languages}},
  author={Bugliarello, Emanuele and Liu, Fangyu and Pfeiffer, Jonas and Reddy, Siva and Elliott, Desmond and Ponti, Edoardo Maria and Vuli{\'c}, Ivan},
  booktitle=icml,
  year={2022},
}

@inproceedings{schuhmann2022laion,
  title={{LAION-5B: An open large-scale dataset for training next generation image-text models}},
  author={Schuhmann, Christoph and Beaumont, Romain and Vencu, Richard and Gordon, Cade and Wightman, Ross and Cherti, Mehdi and Coombes, Theo and Katta, Aarush and Mullis, Clayton and Wortsman, Mitchell and others},
  booktitle=nips,
  year={2022}
}

@inproceedings{shah2019kvqa,
  title={{KVQA: Knowledge-Aware Visual Question Answering}},
  author={Shah, Sanket and Mishra, Anand and Yadati, Naganand and Talukdar, Partha Pratim},
  booktitle=aaai,
  year={2019}
}

@article{izacard2021unsupervised,
  title={{Unsupervised Dense Information Retrieval with Contrastive Learning}},
  author={Izacard, Gautier and Caron, Mathilde and Hosseini, Lucas and Riedel, Sebastian and Bojanowski, Piotr and Joulin, Armand and Grave, Edouard},
  journal={arXiv preprint arXiv:2112.09118},
  year={2021}
}

@article{zheng2017sift,
  title={{SIFT meets CNN: A decade survey of instance retrieval}},
  author={Zheng, Liang and Yang, Yi and Tian, Qi},
  journal=ieeetpami,
  volume={40},
  number={5},
  pages={1224--1244},
  year={2017},
}

@article{liu2024visual,
  title={{Visual Instruction Tuning}},
  author={Liu, Haotian and Li, Chunyuan and Wu, Qingyang and Lee, Yong Jae},
  journal=nips,
  year={2024}
}

@inproceedings{noh2017large,
  title={Large-scale image retrieval with attentive deep local features},
  author={Noh, Hyeonwoo and Araujo, Andre and Sim, Jack and Weyand, Tobias and Han, Bohyung},
  booktitle=cvpr,
  year={2017}
}

@inproceedings{nguyen2016msmarco,
  title={{MS MARCO: A Human Generated MAchine Reading COmprehension Dataset}},
  author={Nguyen, Tri and Rosenberg, Mir and Song, Xia and Gao, Jianfeng and Tiwary, Saurabh and Majumder, Rangan and Deng, Li},
  booktitle=nips,
  year={2016}
}

@article{sun2023eva,
  title={{EVA-CLIP: Improved Training Techniques for CLIP at Scale}},
  author={Sun, Quan and Fang, Yuxin and Wu, Ledell and Wang, Xinlong and Cao, Yue},
  journal={arXiv preprint arXiv:2303.15389},
  year={2023}
}

@inproceedings{caffagni2024wiki,
  title={{Wiki-LLaVA: Hierarchical Retrieval-Augmented Generation for Multimodal LLMs}},
  author={Caffagni, Davide and Cocchi, Federico and Moratelli, Nicholas and Sarto, Sara and Cornia, Marcella and Baraldi, Lorenzo and Cucchiara, Rita},
  booktitle=cvprw,
  year={2024}
}

@inproceedings{yan2024echosight,
  title={{EchoSight: Advancing Visual-Language Models with Wiki Knowledge}},
  author={Yan, Yibin and Xie, Weidi},
  booktitle={{EMNLP Findings}},
  year={2024}
}

@inproceedings{caffagni2024revolution,
  title={{The Revolution of Multimodal Large Language Models: A Survey}},
  author={Caffagni, Davide and Cocchi, Federico and Barsellotti, Luca and Moratelli, Nicholas and Sarto, Sara and Baraldi, Lorenzo and Cornia, Marcella and Cucchiara, Rita},
  booktitle={{ACL Findings}},
  year={2024}
}

@inproceedings{caffagni2025recurrence,
  title={{Recurrence-Enhanced Vision-and-Language Transformers for Robust Multimodal Document Retrieval}},
  author={Caffagni, Davide and Sarto, Sara and Cornia, Marcella and Baraldi, Lorenzo and Cucchiara, Rita},
  booktitle=cvpr,
  year={2025}
}

@inproceedings{liu2025lamra,
  title={{LamRA: Large Multimodal Model as Your Advanced Retrieval Assistant}},
  author={Liu, Yikun and Zhang, Yajie and Cai, Jiayin and Jiang, Xiaolong and Hu, Yao and Yao, Jiangchao and Wang, Yanfeng and Xie, Weidi},
  booktitle=cvpr,
  year={2025}
}

@inproceedings{lin2025mm,
  title={{MM-Embed: Universal Multimodal Retrieval with Multimodal LLMs}},
  author={Lin, Sheng-Chieh and Lee, Chankyu and Shoeybi, Mohammad and Lin, Jimmy and Catanzaro, Bryan and Ping, Wei},
  booktitle=iclr,
  year={2025}
}

@inproceedings{kim2025genius,
  title={{GENIUS: A Generative Framework for Universal Multimodal Search}},
  author={Kim, Sungyeon and Zhu, Xinliang and Lin, Xiaofan and Bastan, Muhammet and Gray, Douglas and Kwak, Suha},
  booktitle=cvpr,
  year={2025}
}

@inproceedings{liu2024improved,
  title={{Improved Baselines with Visual Instruction Tuning}},
  author={Liu, Haotian and Li, Chunyuan and Li, Yuheng and Lee, Yong Jae},
  booktitle=cvpr,
  year={2024}
}

@inproceedings{dai2023instructblip,
  title={{InstructBLIP: Towards General-purpose Vision-Language Models with Instruction Tuning}},
  author={Dai, Wenliang and Li, Junnan and Li, Dongxu and Tiong, Anthony and Zhao, Junqi and Wang, Weisheng and Li, Boyang and Fung, Pascale N and Hoi, Steven},
  booktitle=nips,
  year={2023}
}

@article{bai2025qwen2,
  title={{Qwen2.5-VL Technical Report}},
  author={Bai, Shuai and Chen, Keqin and Liu, Xuejing and Wang, Jialin and Ge, Wenbin and Song, Sibo and Dang, Kai and Wang, Peng and Wang, Shijie and Tang, Jun and others},
  journal={arXiv preprint arXiv:2502.13923},
  year={2025}
}

@inproceedings{cocchi2025augmenting,
  title={{Augmenting Multimodal LLMs with Self-Reflective Tokens for Knowledge-based Visual Question Answering}},
  author={Cocchi, Federico and Moratelli, Nicholas and Cornia, Marcella and Baraldi, Lorenzo and Cucchiara, Rita},
  booktitle=cvpr,
  year={2025}
}

@inproceedings{dalvi2020analyzing,
  title={{Analyzing Redundancy in Pretrained Transformer Models}},
  author={Dalvi, Fahim and Sajjad, Hassan and Durrani, Nadir and Belinkov, Yonatan},
  booktitle=emnlp,
  year={2020}
}

@article{he2024matters,
  title={{What Matters in Transformers? Not All Attention is Needed}},
  author={He, Shwai and Sun, Guoheng and Shen, Zheyu and Li, Ang},
  journal={arXiv preprint arXiv:2406.15786},
  year={2024}
}

@inproceedings{josephlambda,
  title={{Lambda-Skip Connections: the Architectural Component that Prevents Rank Collapse}},
  author={Joseph, Federico Arangath and Sieber, Jerome and Zeilinger, Melanie and Alonso, Carmen Amo},
  booktitle=iclr,
  year=2025
}

@inproceedings{bulian2022tomayto,
  title={{Tomayto, Tomahto. Beyond Token-level Answer Equivalence for Question Answering Evaluation}},
  author={Bulian, Jannis and Buck, Christian and Gajewski, Wojciech and B{\"o}rschinger, Benjamin and Schuster, Tal},
  booktitle=emnlp,
  year={2022}
}

@article{tschannen2025siglip,
  title={{SigLIP 2: Multilingual Vision-Language Encoders with Improved Semantic Understanding, Localization, and Dense Features}},
  author={Tschannen, Michael and Gritsenko, Alexey and Wang, Xiao and Naeem, Muhammad Ferjad and Alabdulmohsin, Ibrahim and Parthasarathy, Nikhil and Evans, Talfan and others},
  journal={arXiv preprint arXiv:2502.14786},
  year={2025}
}

@inproceedings{liu2020visual,
  title={{Visual News: Benchmark and Challenges in News Image Captioning}},
  author={Liu, Fuxiao and Wang, Yinghan and Wang, Tianlu and Ordonez, Vicente},
  booktitle=emnlp,
  year={2021}
}

@inproceedings{han2017automatic,
  title={{Automatic Spatially-Aware Fashion Concept Discovery}},
  author={Han, Xintong and Wu, Zuxuan and Huang, Phoenix X and Zhang, Xiao and Zhu, Menglong and Li, Yuan and Zhao, Yang and Davis, Larry S},
  booktitle=iccv,
  year={2017}
}

@inproceedings{chang2022webqa,
  title={{WebQA: Multihop and Multimodal QA}},
  author={Chang, Yingshan and Narang, Mridu and Suzuki, Hisami and Cao, Guihong and Gao, Jianfeng and Bisk, Yonatan},
  booktitle=cvpr,
  year={2022}
}

@inproceedings{liu2023edis,
  title={{EDIS: Entity-Driven Image Search over Multimodal Web Content}},
  author={Liu, Siqi and Feng, Weixi and Fu, Tsu-jui and Chen, Wenhu and Wang, William Yang},
  booktitle=emnlp,
  year={2023}
}

@inproceedings{fu2023dreamsim,
  title={{DreamSim: Learning New Dimensions of Human Visual Similarity using Synthetic Data}},
  author={Fu, Stephanie and Tamir, Netanel and Sundaram, Shobhita and Chai, Lucy and Zhang, Richard and Dekel, Tali and Isola, Phillip},
  booktitle=nips,
  year={2023}
}

@inproceedings{wu2021fashion,
  title={{Fashion IQ: A New Dataset Towards Retrieving Images by Natural Language Feedback}},
  author={Wu, Hui and Gao, Yupeng and Guo, Xiaoxiao and Al-Halah, Ziad and Rennie, Steven and Grauman, Kristen and Feris, Rogerio},
  booktitle=cvpr,
  year={2021}
}

@inproceedings{liu2021image,
  title={{Image Retrieval on Real-Life Images With Pre-Trained Vision-and-Language Models}},
  author={Liu, Zheyuan and Rodriguez-Opazo, Cristian and Teney, Damien and Gould, Stephen},
  booktitle=iccv,
  year={2021}
}

@inproceedings{zhang2024magiclens,
  title={{MagicLens: Self-Supervised Image Retrieval with Open-Ended Instructions}},
  author={Zhang, Kai and Luan, Yi and Hu, Hexiang and Lee, Kenton and Qiao, Siyuan and Chen, Wenhu and Su, Yu and Chang, Ming-Wei},
  booktitle=icml,
  year={2024},
}

@article{jiang2024e5,
  title={{E5-V: Universal Embeddings with Multimodal Large Language Models}},
  author={Jiang, Ting and Song, Minghui and Zhang, Zihan and Huang, Haizhen and Deng, Weiwei and Sun, Feng and Zhang, Qi and others},
  journal={arXiv preprint arXiv:2407.12580},
  year={2024}
}

@inproceedings{compagnoni2025reag,
  title={{ReAG: Reasoning-Augmented Generation for Knowledge-based Visual Question Answering}},
  author={Compagnoni, Alberto and Morini, Marco and Sarto, Sara and Cocchi, Federico and Caffagni, Davide and Cornia, Marcella and Baraldi, Lorenzo and Cucchiara, Rita},
  booktitle=cvpr,
  year={2026}
}

@inproceedings{jiangvlm2vec,
  title={{VLM2Vec: Training Vision-Language Models for Massive Multimodal Embedding Tasks}},
  author={Jiang, Ziyan and Meng, Rui and Yang, Xinyi and Yavuz, Semih and Zhou, Yingbo and Chen, Wenhu},
  booktitle=iclr,
  year={2025}
}

@inproceedings{hong2025knowledge,
  title={{Knowledge-based Visual Question Answer with Multimodal Processing, Retrieval and Filtering}},
  author={Hong, Yuyang and Gu, Jiaqi and Yang, Qi and Fan, Lubin and Wu, Yue and Wang, Ying and Ding, Kun and Xiang, Shiming and Ye, Jieping},
  booktitle=nips,
  year={2025}
}

@inproceedings{dong2021attention,
  title={Attention is not all you need: pure attention loses rank doubly exponentially with depth},
  author={Dong, Yihe and Cordonnier, Jean Baptiste and Loukas, Andreas},
  booktitle=icml,
  year={2021},
}

@inproceedings{voita2019bottom,
  title={{The Bottom-up Evolution of Representations in the Transformer: A Study with Machine Translation and Language Modeling Objectives}},
  author={Voita, Elena and Sennrich, Rico and Titov, Ivan},
  booktitle=emnlp,
  year={2019}
}

@inproceedings{gromov2024unreasonable,
  title={{The Unreasonable Ineffectiveness of the Deeper Layers}},
  author={Gromov, Andrey and Tirumala, Kushal and Shapourian, Hassan and Glorioso, Paolo and Roberts, Daniel A.},
 booktitle=iclr,
  year={2025}
}

@inproceedings{sun2024transformer,
  title={{Transformer Layers as Painters}},
  author={Sun, Qi and Pickett, Marc and Nain, Aakash Kumar and Jones, Llion},
  booktitle=aaai,
  year={2025}
}

@article{bengio1994learning,
  title={{Learning Long-Term Dependencies with Gradient Descent is Difficult}},
  author={Bengio, Yoshua and Simard, Patrice and Frasconi, Paolo},
  journal={IEEE Transactions on Neural Networks},
  year={1994}
}

@inproceedings{pascanu2013difficulty,
  title={{On the Difficulty of Training Recurrent Neural Networks}},
  author={Pascanu, Razvan and Mikolov, Tomas and Bengio, Yoshua},
  booktitle=icml,
  year={2013}
}

\begin{IEEEbiography}
[{\includegraphics[width=1in,clip,keepaspectratio]{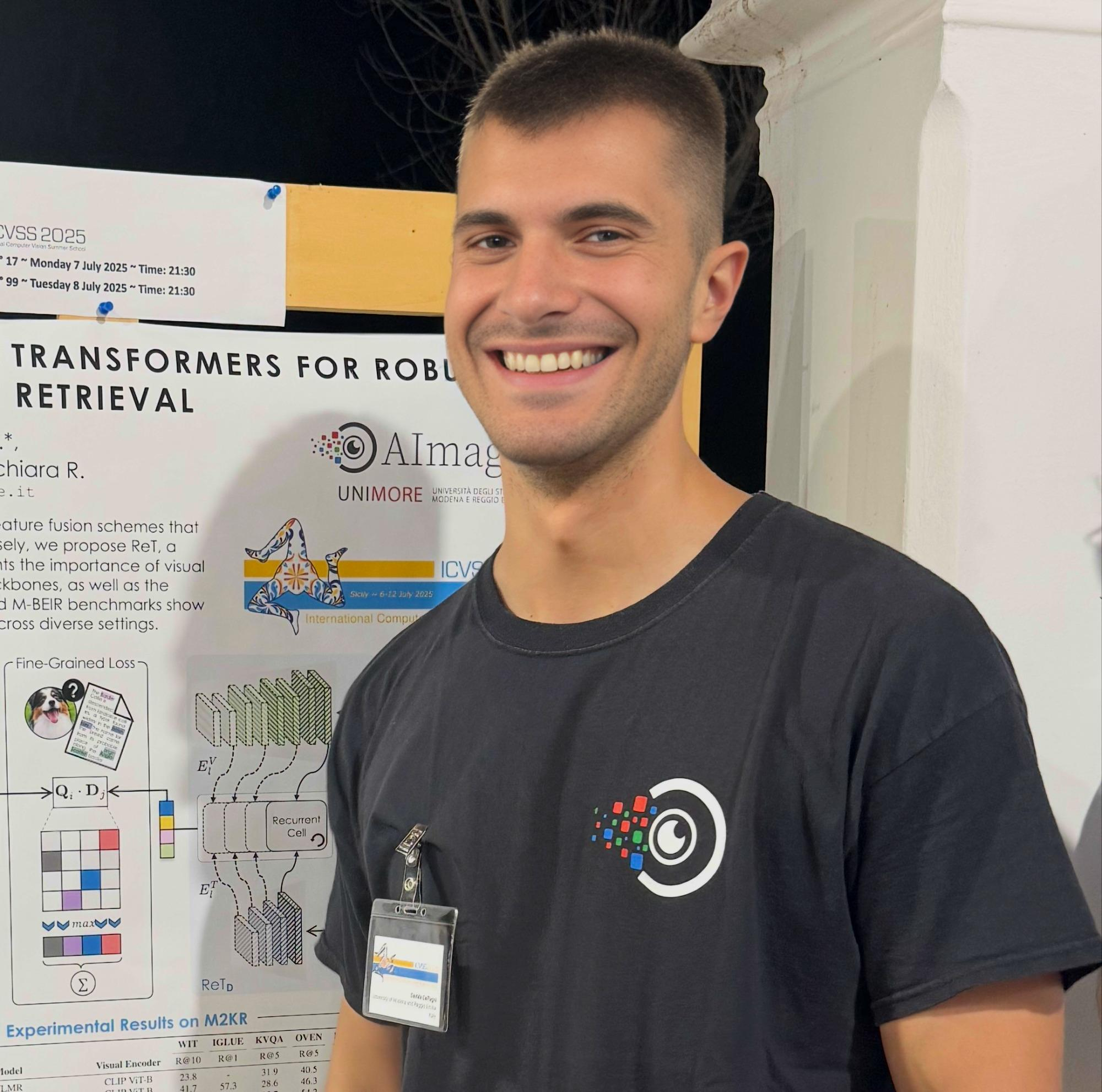}}]{Davide Caffagni} received the M.Sc. degree in Computer Engineering cum laude from the University of Modena and Reggio Emilia in 2023. He is currently pursuing a PhD in Information and Communication Technologies (ICT) at the University of Modena and Reggio Emilia. His research topics include Computer Vision and Natural Language Processing, with a focus on image captioning, multimodal large language models, and multimodal retrieval.
\end{IEEEbiography}

\begin{IEEEbiography}[{\includegraphics[width=1in,height=1.25in,clip,keepaspectratio]{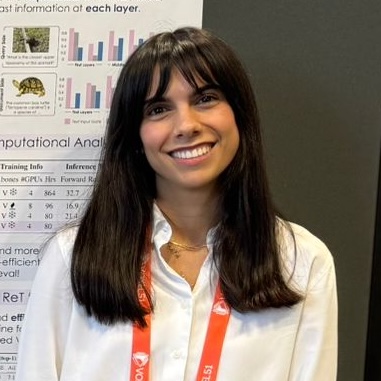}}]{Sara Sarto} received the Ph.D. degree cum laude in ICT from the University of Modena and Reggio Emilia in 2026. She is currently a Postdoctoral Researcher at the Department of Engineering ``Enzo Ferrari'', University of Modena and Reggio Emilia. She was an Applied Scientist Intern at Amazon in 2025.  Her research interests include vision-and-language models and multimodal learning, focusing on cross-modal and multimodal retrieval.
\end{IEEEbiography}

\begin{IEEEbiography}[{\includegraphics[width=1in,height=1.25in,clip,keepaspectratio]{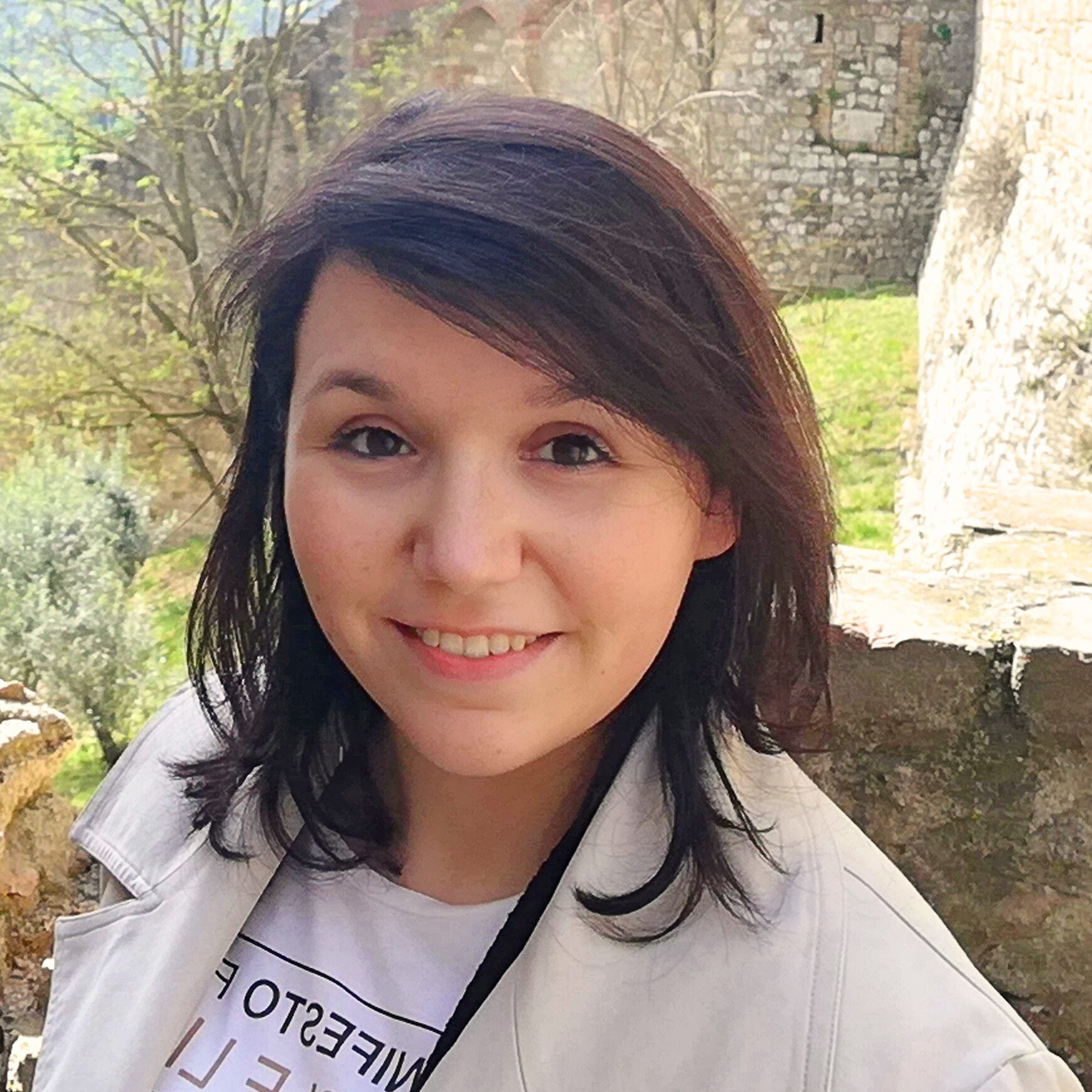}}]{Marcella Cornia} received the Ph.D. degree cum laude in ICT from the University of Modena and Reggio Emilia in 2020. She is currently an Associate Professor with the Department of Education and Humanities, University of Modena and Reggio Emilia. She has authored or coauthored more than 100 publications in scientific journals and international conference proceedings. Her research interests include vision-and-language tasks, generative AI, and multimodal learning. She is member of ELLIS and regularly serves as Area Chair/Reviewer for international conferences and journals.
\end{IEEEbiography}

\begin{IEEEbiography}[{\includegraphics[width=1in,height=1.25in,clip,keepaspectratio]{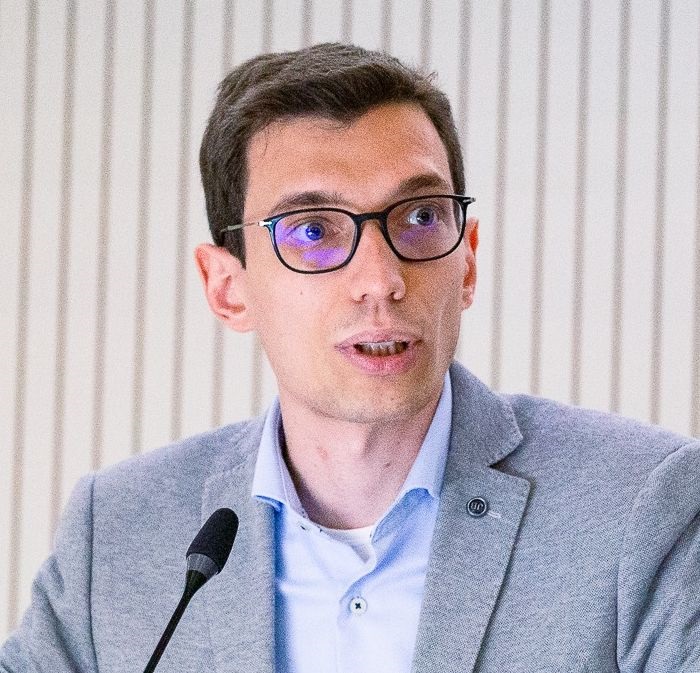}}]{Lorenzo Baraldi} received the Ph.D. degree cum laude in ICT from the University of Modena and Reggio Emilia in 2018. He is currently an Associate Professor with the Department of Engineering ``Enzo Ferrari'', University of Modena and Reggio Emilia. He was a Research Intern at Facebook AI Research (FAIR) in 2017. He has authored or coauthored more than 130 publications in scientific journals and international conference proceedings. His research interests include video understanding, deep learning, and multimedia. He is an ELLIS Scholar and regularly serves as Area Chair/Reviewer for international conferences and journals.
\end{IEEEbiography}

\begin{IEEEbiography}[{\includegraphics[width=1in,height=1.25in,clip,keepaspectratio]{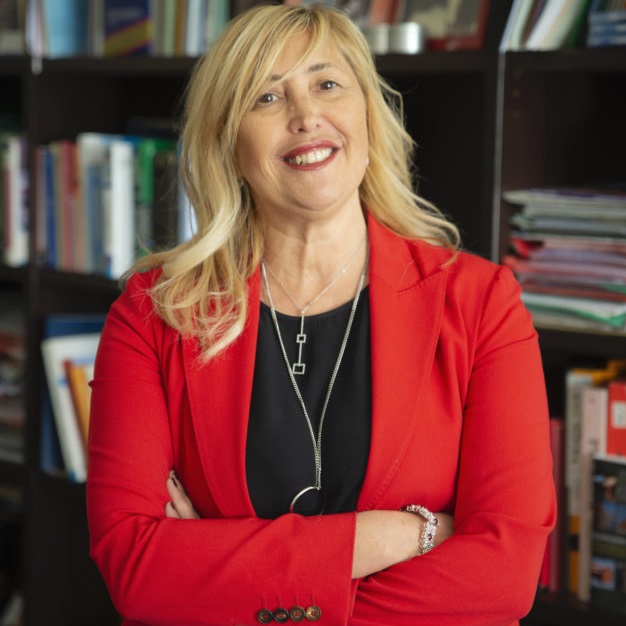}}]{Rita Cucchiara} received the Ph.D. degree in Computer Engineering from the University of Bologna in 1992. She is currently a Full Professor of Computer Engineering and the Rector of the University of Modena and Reggio Emilia, where she also heads the AImageLab Laboratory. She has authored or coauthored more than 600 papers in journals and international proceedings, and has been a coordinator of several projects in computer vision and pattern recognition. She also is ELLIS Fellow, director of the ELLIS Unit of Modena (Unimore), and Coordinator of the Unimore Unit of the National Ph.D. School in AI for Society.
\end{IEEEbiography}

\clearpage
\newpage
\setcounter{page}{1}
\appendices

\section{Additional Implementation Details}
\label{sec:supp_details}

\tinytit{Additional Details on M2KR}
We report in Table~\ref{tab:m2kr_dataset_detail} the detailed composition of the M2KR benchmark. While the MSMARCO dataset (highlighted in gray in the table) is part of the original benchmark, we opted not to include it in our experiments due to its entirely textual nature (\ie, $q^T \to d^T$). Other datasets, namely WIT, KVQA, IGLUE, and CC3M, are designed to evaluate the ability of the models to identify relevant documents given an input image (\ie,  $(q^V, q^T) \to d^T$).
While CC3M is incorporated into the M2KR training set to enhance scene understanding, it is excluded from validation and test splits since its original focus is caption generation rather than retrieval. The IGLUE benchmark, a subset of WIT, is retained to ensure comparability with prior work. Finally, KVQA, originally introduced as a knowledge-based VQA task, is adapted to fit our setting.

In this setting, each item is paired with an instruction sampled from predefined templates. For instance, WIT and IGLUE use instructions like \texttt{\{Image\} Please describe the document that corresponds to this image}; KVQA uses \texttt{\{Image\} Provide a brief description of the image and the relevant details of the person in the image}; and CC3M uses templates similar to \texttt{\{Image\} Describe the image concisely}. For a detailed enumeration of all possible instruction templates, we refer the reader to the original M2KR paper~\cite{lin2024preflmr}.

The task $(q^V, q^T) \to (d^V, d^T)$ requires joint understanding of both images and text for accurate retrieval. In this setting, we use datasets such as OVEN, LLaVA, OKVQA, InfoSeek, and Encyclopedic-VQA. Training, validation, and test samples are downsampled from the original datasets. In the original M2KR benchmark, these datasets were not fully multimodal -- \ie, the document side did not include visual input. To better align with our setting, we augment these splits by enriching the reference documents with images, as described in Sec.~\ref{sec:datasets_ret}. This dataset version is available in our repository.

\tit{Additional Details on M-BEIR}
The M-BEIR benchmark consists of eight multimodal retrieval tasks spanning ten datasets from diverse domains and image sources.  In particular, it standardizes training and evaluation by repurposing diverse datasets. 
Image-caption datasets (COCO~\cite{lin2014microsoft}, Fashion200k~\cite{han2017automatic}, and Visual News~\cite{liu2020visual}) are adapted by treating captions as queries, while NIGHTS~\cite{fu2023dreamsim} addresses visual similarity in nighttime scenes. It also includes retrieval-based VQA datasets (InfoSeek~\cite{chen2023can}, WebQA~\cite{chang2022webqa}), where documents are relevant if they contain the answer.
These tasks are designed to evaluate performance under both missing-modality and fully multimodal scenarios. For example, Task \#2 involves purely textual inputs on both the query and document sides, whereas Task \#8 is fully multimodal, with both queries and documents containing visual and textual information. Other tasks, such as \#3 and \#7, introduce asymmetry by providing multimodal inputs on only one side -- either the query or the document -- testing the ability of the models to handle missing modalities. Detailed dataset splits are reported in Table~\ref{tab:mbeir_dataset_detail}.

\begin{table}[t]
\centering
\caption{Summary of the M2KR benchmark~\cite{lin2024preflmr}. For each dataset, we report the number of training, validation, and test samples, along with the size of the document pool (split into training and validation/test). \colorbox{ourcolor}{Purple color} highlights datasets augmented with document-side images, while \colorbox{LightGray}{gray} denotes datasets excluded from both training and test. The last row reflects only the datasets used in our experiments.}
\resizebox{\linewidth}{!}{  
\setlength{\tabcolsep}{.32em}{
  \begin{tabular}{lcc ccc c c}
    \toprule
    & & \multicolumn{3}{c}{\textbf{Query}} && \textbf{Document} \\
    \cmidrule{3-5}  \cmidrule{7-7} 
    \textbf{Task} & \textbf{Dataset} & \textbf{\# Train} & \textbf{\# Val} & \textbf{\# Test} && \textbf{\# Pool}\\
    \midrule
    \rowcolor{LightGray}
    \multirow{1}{*}{$q^T \to d^T$} & \textgray{MSMARCO~\cite{nguyen2016msmarco}} & \textgray{400k} & \textgray{6.9k} & \textgray{5.1k} && \textgray{8.8M/200k} \\
     \midrule
    \multirow{4}{*}{$(q^V, q^T) \to d^T$} & WIT~\cite{srinivasan2021wit} & 2.8M & 20.1k & 5.1k && 4.1M/40k \\
     & IGLUE~\cite{bugliarello2022iglue} & - & - & 685 && -/1k \\
     & KVQA~\cite{shah2019kvqa} & 16k & 13.4k & 5.1k && 16.3k/4.6k \\
     & CC3M~\cite{sharma2018conceptual} & 595k & - & - && 595k/- \\
     \midrule
     \rowcolor{ourcolor}
     \cellcolor{white} & OVEN~\cite{hu2023open} & 339k & 20k & 5.1k && 10k/3.1k \\
     \rowcolor{ourcolor}
     \cellcolor{white} & OKVQA~\cite{marino2019ok} & 9k & 5k & 5k && 110k/110k \\
     \rowcolor{ourcolor}
     \cellcolor{white} & InfoSeek~\cite{chen2023can} & 76k & - & 4.7k && 100k/100k \\
     \rowcolor{ourcolor}
     \cellcolor{white}\multirow{-4}{*}{$(q^V, q^T) \to (d^V, d^T)$} & E-VQA~\cite{mensink2023encyclopedic} & 167k & 9.8k & 3.7k && 50k/50k \\
     \midrule
     \midrule
    2 tasks & 8 datasets & 4.8M & 68.3k & 29.4k && 4.98M/308k \\
    \bottomrule
  \end{tabular}
}
}
\label{tab:m2kr_dataset_detail}
\vspace{-0.3cm}
\end{table}

\begin{table}[t]
\centering
\caption{Summary of the M-BEIR benchmark~\cite{wei2024uniir}. For each dataset, we report the number of training, validation, and test samples, along with the size of the document pool.}
\resizebox{\linewidth}{!}{  
\setlength{\tabcolsep}{.25em}{
  \begin{tabular}{lcc ccc c c}
    \toprule
    & & \multicolumn{3}{c}{\textbf{Query}} && \multicolumn{1}{c}{\textbf{Document}} \\
    \cmidrule{3-5}  \cmidrule{7-7} 
    \textbf{Task} & \textbf{Dataset} & \textbf{\# Train} & \textbf{\# Val} & \textbf{\# Test} && \textbf{\# Pool}\\
    \midrule
    \multirow{3}{*}{\textbf{\#1: }$q^T \to d^V$} & VN~\cite{liu2020visual} & 99k & 20k & 20k && 542k \\
     & COCO~\cite{lin2014microsoft} & 100k & 24.8k & 24.8k && 5k \\
     & F200k~\cite{han2017automatic}  & 15k & 1.7k & 1.7k && 201k \\
     \midrule
     \multirow{1}{*}{\textbf{\#2: }$q^T \to d^T$} & WQA~\cite{chang2022webqa} & 16k & 1.7k & 2.4k && 544k \\
     \midrule
     \multirow{2}{*}{\textbf{\#3: }$q^T \to (d^V, d^T)$} & EDIS~\cite{liu2023edis} & 26k & 3.2k & 3.2k && 1M \\
     & WQA~\cite{chang2022webqa} & 17k & 1.7k & 2.5k && 403k \\
     \midrule
     \multirow{3}{*}{\textbf{\#4: }$q^V \to d^T$} & VN~\cite{liu2020visual} & 100k & 20k & 20k && 537k \\
     & COCO~\cite{lin2014microsoft} & 113k & 5k & 5k && 25k \\
     & F200k~\cite{han2017automatic} & 15k & 4.8k & 4.8k && 61k \\
     \midrule
     \multirow{1}{*}{\textbf{\#5: }$q^V \to d^V$} & NIGHTS~\cite{fu2023dreamsim} & 16k & 2k & 2k && 40k \\
     \midrule
     \multirow{2}{*}{\textbf{\#6: }$(q^V, q^T) \to d^T$} & OVEN~\cite{hu2023open} & 150k & 50k & 50k && 676k \\
     & InfoSeek~\cite{chen2023can} & 141k & 11k & 11k && 611k \\
     \midrule
     \multirow{2}{*}{\textbf{\#7: }$(q^V, q^T) \to d^V$} & FIQ~\cite{wu2021fashion} & 16k & 2k & 6k && 74k \\
     & CIRR~\cite{liu2021image} & 26k & 2k & 4k && 21k \\
     \midrule
     \multirow{2}{*}{\textbf{\#8: }$(q^V, q^T) \to (d^V, d^T)$} & OVEN~\cite{hu2023open} & 157k & 14.7k & 14.7k && 335k \\
     & InfoSeek~\cite{chen2023can} & 143k & 17.6k & 17.6k && 481k \\
     \midrule
     \midrule
    8 tasks & 10 datasets & 1.1M & 182k & 190k && 5.6M \\
    \bottomrule
  \end{tabular}
}
}
\label{tab:mbeir_dataset_detail}
\vspace{-0.3cm}
\end{table}

\tit{Retrieval-Augmented VQA}
In our experiments, we prompt the MLLM with the top-$K$ documents retrieved using \oursnew. The prompt used to generate answers is defined as follows:

\resizebox{0.96\linewidth}{!}{
\begin{minipage}{\linewidth}
\vspace{0.25cm}
\noindent\texttt{\{Image\} Given the context, answer the question based on the image.}

\noindent\texttt{Question: \{Question\}}

\noindent\texttt{Context:}

\noindent \texttt{\#\# \{C$_\text{1}$\}} 

\noindent \texttt{\#\# \{...\}}

\noindent \texttt{\#\# \{C$_K$\}}

\noindent\texttt{If the context does not help with the question, try to answer it anyway. Do not generate anything but the short answer.}

\noindent\texttt{Short answer:}
\vspace{0.3cm}
\end{minipage}
}
where \texttt{C$_K$} is replaced with the text content of the top-$K$ retrieved documents.

\section{\revv{Theoretical Motivation of Token Reduction and Layer Pruning}}
\label{sec:supp_theory}

\revv{This appendix provides the full derivations and literature grounding for the token reduction and layer pruning design choices introduced in Sec.~\ref{sec:method} and empirically validated in Sec.~\ref{sec:experiments} (cf. Table~\ref{tab:ablation_large} and Table~\ref{tab:rebuttal_ablation_layers}).}

\subsection{\revv{Token Reduction and Rank Collapse}}
\label{sec:supp_rank_collapse}

\revv{To understand why reducing the number of learnable tokens from $k=32$ to $k=1$ does not sacrifice representational capacity, we relate the internal state of the recurrent cell to the rank-collapse phenomenon described by Dong~\etal~\cite{dong2021attention} for pure self-attention networks (SANs). This analysis further accompanies the empirical rank-collapse measurements reported in Sec.~\ref{sec:experiments}.} 

\revv{Reinstating, for this analysis, the general $k$-token notation used for \ours in Sec.~\ref{sec:background}, let $\mathbf{H}_l \in \mathbb{R}^{k \times d}$ denote the stacked hidden state of the recurrent cell at layer $l$ (which reduces to the single row vector $\inputH$ used elsewhere in the paper once $k=1$). Following~\cite{dong2021attention}, $\mathbf{H}_l$ can be decomposed into a component shared across all $k$ rows and a residual capturing their differences:}
\begin{equation}
    \mathbf{H}_l = \mathbf{1}\bar{\mathbf{h}}_l^\top + \mathbf{R}_l, \qquad \bar{\mathbf{h}}_l = \frac{1}{k}\mathbf{1}^\top \mathbf{H}_l, \qquad \mathbf{R}_l = \Pi \mathbf{H}_l,
    \label{eq:supp_decomp}
\end{equation}
\revv{where $\Pi = \mathbf{I}_k - \frac{1}{k}\mathbf{1}\mathbf{1}^\top$ is the centering matrix. The term $\mathbf{1}\bar{\mathbf{h}}_l^\top$ has rank at most one and identical rows, whereas $\mathbf{R}_l$ contains all the information that distinguishes one token from another; $\mathbf{R}_l = \mathbf{0}$ exactly when all $k$ tokens have collapsed onto the same representation.}

\revv{Dong~\etal~show that, in a pure SAN, repeated self-attention contracts $\mathbf{R}_l$ at a doubly-exponential rate with depth, and that this contraction is only counteracted by skip connections and MLPs, which slow but do not reverse it. The recurrent cell of \ours repeatedly mixes the same $k=32$ latent rows of $\mathbf{H}_l$ across every processed layer of the backbone (\ie, self-attention among the $k$ tokens, followed by two cross-attention operations against $\mathbf{E}^T_l$ and $\mathbf{E}^V_l$, cf. Eq.~\eqref{eq:unimodal_component}), so the number of recurrent steps plays a role analogous to the depth of a SAN with respect to the contraction of $\mathbf{R}_l$. Three structural properties of the cell reinforce this contraction:}
\begin{itemize}
    \item \revv{The cross-attention update in Eq.~\eqref{eq:unimodal_component} uses $\mathbf{H}_l$ (or its normalized form $\hatH$) to produce the queries. If two rows $\mathbf{h}_{l,i} \approx \mathbf{h}_{l,j}$, they yield near-identical query vectors, near-identical attention distributions over the shared context $\mathbf{E}^m_l$, and therefore near-identical outputs.}
    \item \revv{The gates and the feed-forward network in Eq.~\eqref{eq:forget_gate}--\eqref{eq:next_input} are shared across all $k$ token positions. If two rows are identical, these shared, position-independent transformations map them to identical outputs.}
    \item \revv{The residual path connecting $\candidate$ to $\candidateNext$ in Eq.~\eqref{eq:final} is modulated by the forget gate $\forgetGate$, which is confined to $[0,1]$ (Eq.~\eqref{eq:forget_gate}). Because $\forgetGate$ is trained to actively suppress stale or irrelevant information, it weakens a residual connection that would otherwise counteract collapse. We verified this by removing $\forgetGate$ from Eq.~\eqref{eq:final}, which restores an unmodulated residual path. This measurably increases token diversity, but it also removes the cell's ability to forget outdated content, which we found to hurt retrieval performance.}
\end{itemize}

\subsection{\revv{Layer Pruning and Recurrence Length}}
\label{sec:supp_layer_pruning}

\revv{Our layer pruning strategy restricts the recurrent cell to attend one early, one middle, and one late layer of each backbone, rather than all $L$ layers. This is motivated by two complementary mechanisms, one concerning the length of the recurrence, the other the statistical redundancy of its inputs.}

\tit{\revv{(i) Recurrence Length and Memory Retention}}
\revv{The recurrent cell of \oursnew follows an LSTM-style additive-forget update (Eq.~\eqref{eq:final}): at each of the processed layers, the previous candidate state is attenuated by the sigmoidal forget gate $\forgetGate \in [0,1]$ and combined with fresh, layer-specific evidence. Unrolling Eq.~\eqref{eq:final} shows that the contribution of an early layer $l_0$ to the final state $\finalH$ is multiplicatively discounted by the product $\prod_{l=l_0}^{L-1}\forgetGate$ of every forget gate encountered afterwards. Because each factor lies in $(0,1)$, this product can only remain non-negligible if the gates learn to stay close to $1$ consistently across the entire remaining chain, which becomes harder as $L$ grows, mirroring the classical difficulty of learning long-range dependencies with recurrent models~\cite{bengio1994learning,pascanu2013difficulty}. Gating, as introduced by LSTMs~\cite{hochreiter1997long},  still requires the gates to be correctly trained over the whole unrolled chain, which is  easier over the shorter sequences used after layer pruning than over longer sequences.}

\tit{\revv{(ii) Residual-stream Redundancy Across Depth}}
\revv{Both the visual and textual backbones used in \oursnew are pre-trained Transformers, in which every layer updates the representation through an additive residual connection, $\mathbf{E}^m_{l+1} = \mathbf{E}^m_l + \Delta^m_l$. As the update $\Delta^m_l$ is small relative to $\mathbf{E}^m_l$~\cite{voita2019bottom,gromov2024unreasonable,sun2024transformer}, consecutive layers end up representing highly similar information. Gromov~\etal~\cite{gromov2024unreasonable} show directly, via layer-wise cosine similarity, that deep and middle layers are often near-redundant with their immediate neighbors.}

\revv{This redundancy has a direct consequence for the recurrent cell: the gates $\forgetGate$ and $\genericInputGate$ at layer $l$ are themselves computed from $\genericZ$, which depends on $\mathbf{E}^m_l$ (Eq.~\eqref{eq:forget_gate}). If $\mathbf{E}^m_l \approx \mathbf{E}^m_{l+1}$, the gates computed at these two consecutive steps will also be near-identical -- including any miscalibration. Concretely, if the cell under- or over-estimates the relevance of some cue at layer $l$ it is likely to reproduce the same misjudgment at layer $l+1$, since it is presented with virtually the same evidence. Two such consecutive steps do not behave as two independent opportunities to refine the internal state; they behave as one decision applied twice, consuming recurrence capacity without adding genuinely new information and, in the worst case, reinforcing a gating error.}

\revv{Running the recurrent cell over all $L$ layers therefore generates an error accumulation risk, which is lowered by selecting a subset of layers via layer pruning. Under this reading, layer pruning can be understood as a way of matching the recurrent cell's limited processing budget to the non-uniform rate at which the backbone actually accumulates new information across depth.}

\section{Qualitative Results}
\label{sec:supp_qualitatives}
\begin{figure*}[t]
    \centering
    \resizebox{\linewidth}{!}{
    \setlength{\tabcolsep}{.5em}
    \begin{tabular}{p{4.2cm} p{4.2cm} p{4.2cm} p{4.2cm} p{0.1mm}}
    \toprule
    \multicolumn{2}{c}{\textbf{WIT}} & \multicolumn{2}{c}{\textbf{IGLUE}} \\
    \midrule
    \centering\includegraphics[width=0.9\linewidth,height=0.7\linewidth]{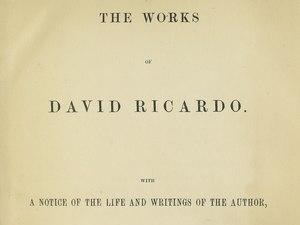} &
    \centering\includegraphics[width=0.9\linewidth,height=0.7\linewidth]{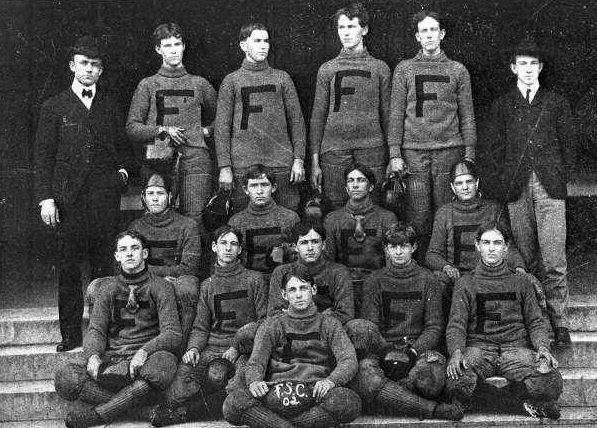} & 
    \centering\includegraphics[width=0.9\linewidth,height=0.7\linewidth]{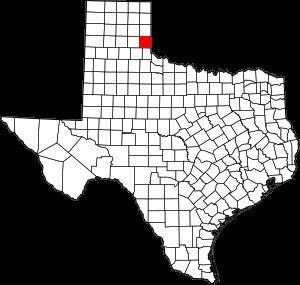} &
    \centering\includegraphics[width=0.9\linewidth,height=0.7\linewidth]{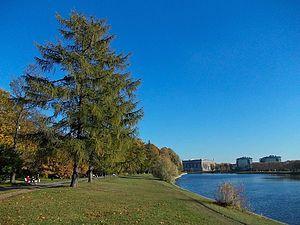} & \\
    \textit{\footnotesize Could you elucidate the document associated with this image?} &  \textit{\footnotesize Identify the document that this image pertains to.} &  \textit{\footnotesize Please give information on the document that goes with this image.} & \textit{\footnotesize Could you elucidate the document associated with this image?} \\
    \midrule
    \footnotesize\textbf{PreFLMR~\cite{lin2024preflmr}}: title: Lilith, The Legend of the First Woman hierarchical section title: Lilith, The Legend of the First Woman caption [...] & 
    \footnotesize\textbf{PreFLMR~\cite{lin2024preflmr}}: title: Pomona-Pitzer Sagehens / History caption reference description: Members of the Pomona football team in 1907 caption [...] & 
    \footnotesize\textbf{PreFLMR~\cite{lin2024preflmr}}: title: Yoakum County, Texas hierarchical section title: Yoakum County, Texas caption reference description: Location [...] & 
    \footnotesize\textbf{PreFLMR~\cite{lin2024preflmr}}: title: Tomaszów Mazowiecki section title: Sulejowski Reservoir hierarchical section title: Tomaszów Mazowiecki [...] \\

    \midrule
    \footnotesize\textbf{\ours~\cite{caffagni2025recurrence}}: title: The Marble Faun hierarchical section title: The Marble Faun caption reference description: First edition title page caption [...] & 
    \footnotesize\textbf{\ours~\cite{caffagni2025recurrence}}: title: 1900 Western University of Pennsylvania football team caption attribution description: English: The 1900 Pittsburgh [...] & 
    \footnotesize\textbf{\ours~\cite{caffagni2025recurrence}}: title: Camp County, Texas hierarchical section title: Camp County, Texas caption reference description: Location within the [...] & 
    \footnotesize\textbf{\ours~\cite{caffagni2025recurrence}}:  title: Tomaszów Mazowiecki section title: Sulejowski Reservoir hierarchical section title: Tomaszów Mazowiecki [...] \\
    
    \midrule
    \footnotesize\textbf{\colorbox{ourcolor}{\oursnew (Ours):}} title: David Ricardo section title: Publications hierarchical section title: David Ricardo / Publications caption [...] & 
    \footnotesize\textbf{\colorbox{ourcolor}{\oursnew (Ours):}} title: List of Florida State University athletes caption reference description: Florida State's first football team, ``The Eleven'' [...] & 
    \footnotesize\textbf{\colorbox{ourcolor}{\oursnew (Ours):}} title: Collingsworth County, Texas hierarchical section title: Collingsworth County, Texas caption reference description: [...] & 
    \footnotesize\textbf{\colorbox{ourcolor}{\oursnew (Ours):}} title: Yelagin Island section title: Current use hierarchical section title: Yelagin Island / Current use caption reference description [...] \\
    \\
    \midrule
    \multicolumn{2}{c}{\textbf{KVQA}} & \multicolumn{2}{c}{\textbf{LLAVA}} \\
    \midrule
    \centering\includegraphics[width=0.9\linewidth,height=0.7\linewidth]{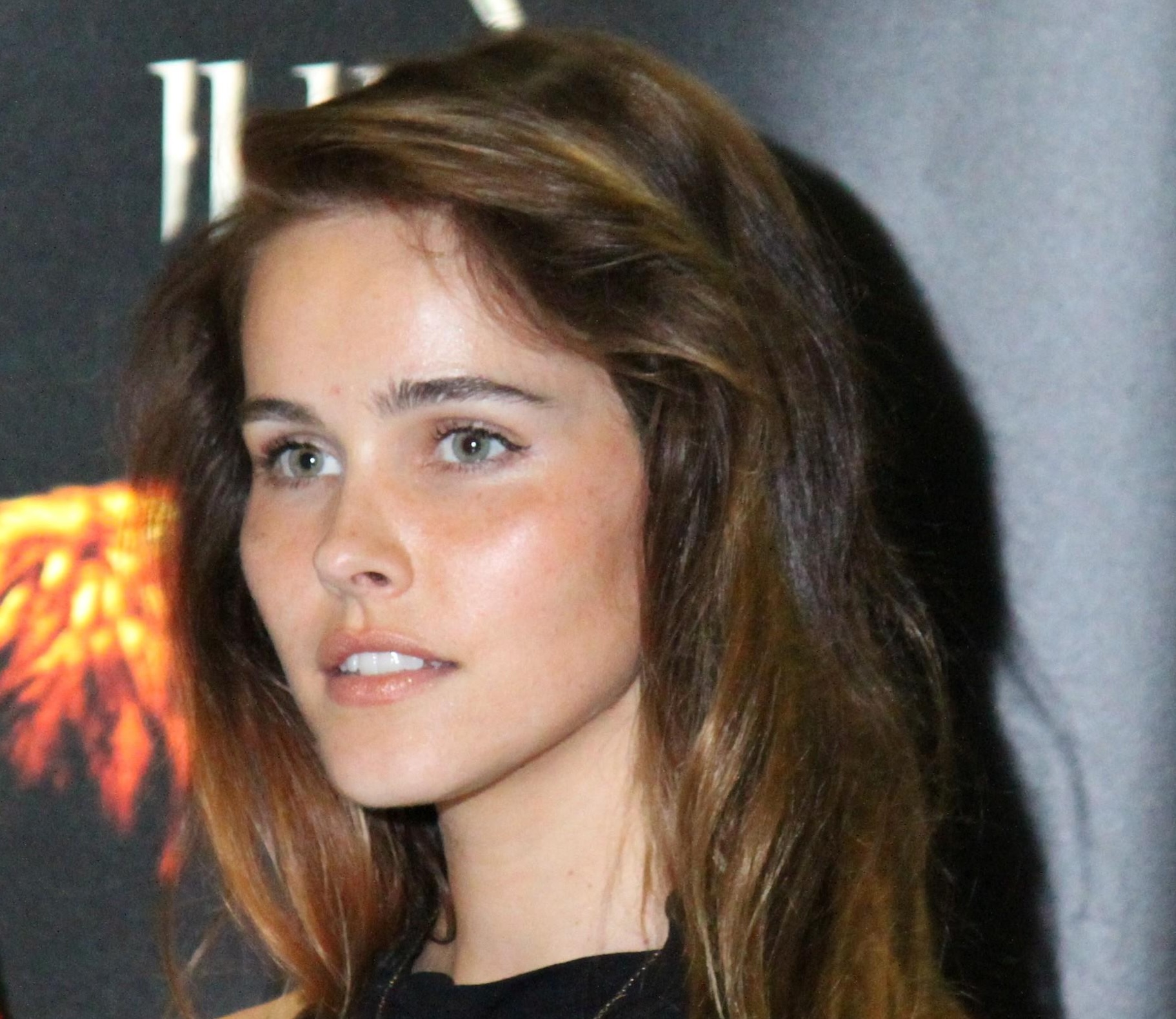} &
    \centering\includegraphics[width=0.9\linewidth,height=0.7\linewidth]{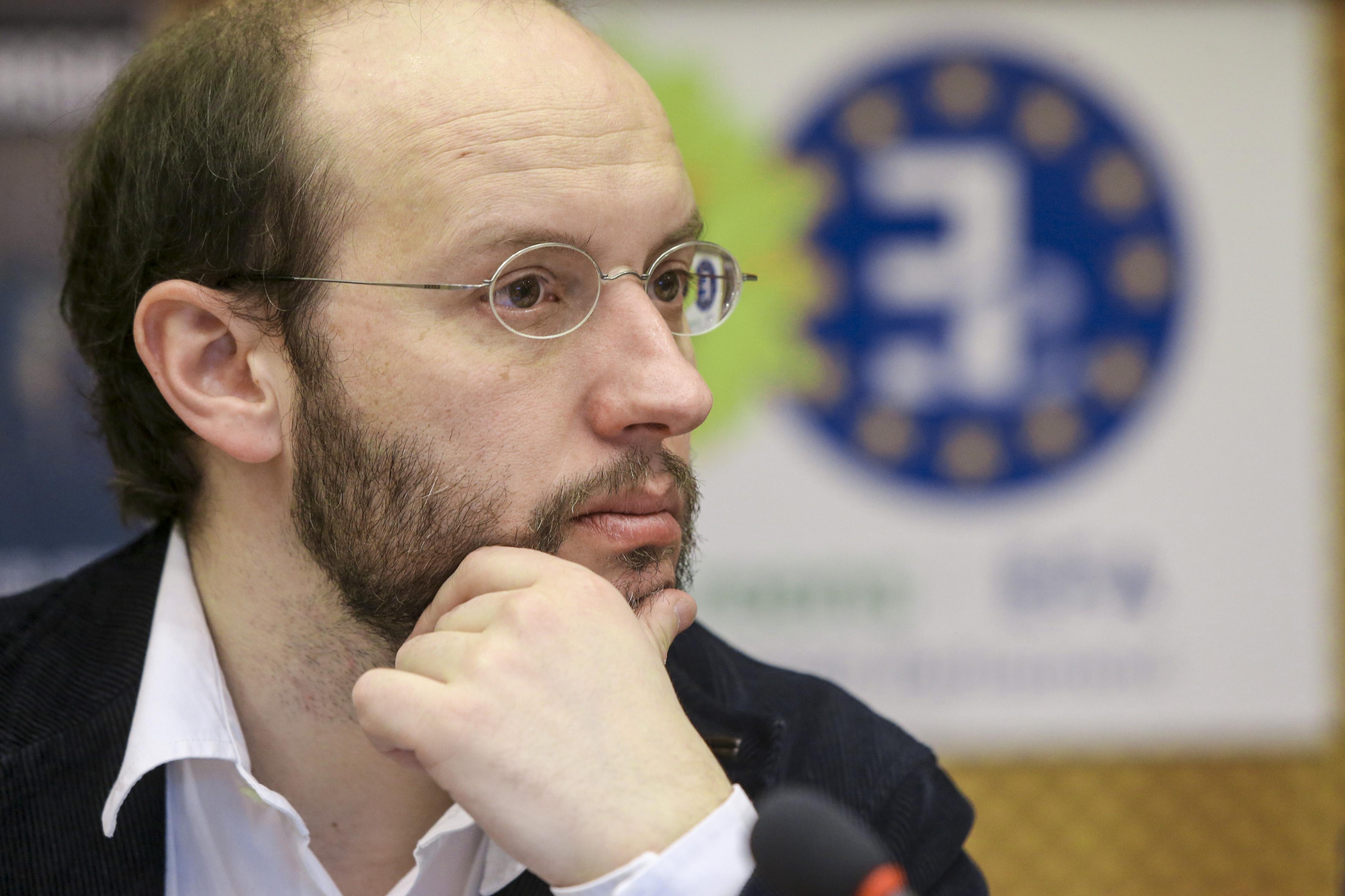} & 
    \centering\includegraphics[width=0.9\linewidth,height=0.7\linewidth]{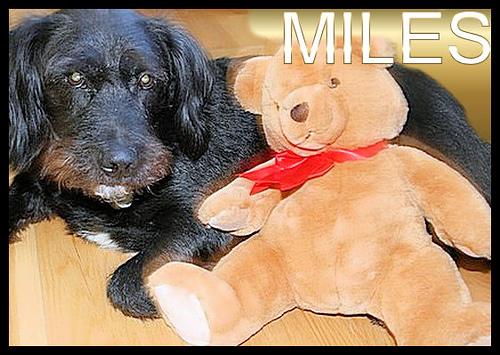} &
    \centering\includegraphics[width=0.9\linewidth,height=0.7\linewidth]{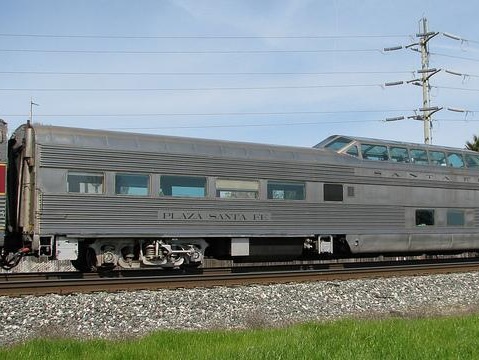} & \\
    \textit{\footnotesize Provide a brief description of the image and the relevant details of the person in the image.} &  \textit{\footnotesize Provide a brief description of the image and the relevant details of the person in the image.} &  \textit{\footnotesize What color is the dog in the image?} & \textit{\footnotesize What type of train is seen in the image?} \\
    \midrule
    \footnotesize\textbf{PreFLMR~\cite{lin2024preflmr}}: This is an image of Byrne at the Sydney film premiere of I Give It a Year in 2013. Rose Byrne went to Australian Theatre for [...] & 
    \footnotesize\textbf{PreFLMR~\cite{lin2024preflmr}}: This is an image of Tõnis Lukas on Estonian Science Communication Conference 2016. Tõnis Lukas went to [...] &
    \footnotesize\textbf{PreFLMR~\cite{lin2024preflmr}}: The dog in the image is white. & 
    \footnotesize\textbf{PreFLMR~\cite{lin2024preflmr}}: The train shown in the image is a passenger train. \\

    \midrule
    \footnotesize\textbf{\ours~\cite{caffagni2025recurrence}}: This is an image of Brie at the 2009 Los Angeles Film Festival. Alison Brie went to California Institute of the Arts, Royal [...] & 
    \footnotesize\textbf{\ours~\cite{caffagni2025recurrence}}: This is an image of Jean-Luc Warsmann (2016). Jean-Luc Warsmann went to Sciences Po, date of birth is 1965-10-22 [...] & 
    \footnotesize\textbf{\ours~\cite{caffagni2025recurrence}}: The dog in the image is brown, with some black markings as well. & 
    \footnotesize\textbf{\ours~\cite{caffagni2025recurrence}}: The train shown in the image is a passenger train. \\
    
    \midrule
    \footnotesize\textbf{\colorbox{ourcolor}{\oursnew (Ours):}} This is an image of Lucas at the 2011 WonderCon. Isabel Lucas date of birth is 1985-01-29, knows English, is a actor, film [...] & 
    \footnotesize\textbf{\colorbox{ourcolor}{\oursnew (Ours):}} This is an image of Rui Tavares (2013). Rui Tavares date of birth is 1972-07-29, is a member of LIVRE (political party), [...] & 
    \footnotesize\textbf{\colorbox{ourcolor}{\oursnew (Ours):}} The dog in the image is black. & 
    \footnotesize\textbf{\colorbox{ourcolor}{\oursnew (Ours):}} A Plaza Santa Fe passenger train is seen in the image. \\
    \bottomrule
    \end{tabular}
    }
    \vspace{-0.15cm}
    \caption{Qualitative results on the M2KR benchmark~\cite{lin2024preflmr}, for datasets that do not include document images.}
    \label{fig:qualitatives_m2kr_1}
\end{figure*}

\begin{figure*}[t]
    \centering
    \resizebox{\linewidth}{!}{
    \setlength{\tabcolsep}{.5em}
    \begin{tabular}{p{4.2cm} p{4.2cm} p{4.2cm} p{4.2cm} p{0.1mm}}
    \toprule
    \multicolumn{2}{c}{\textbf{OVEN}} & \multicolumn{2}{c}{\textbf{InfoSeek}} \\
    \midrule
    \centering\includegraphics[width=0.9\linewidth,height=0.7\linewidth]{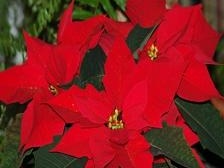} &
    \centering\includegraphics[width=0.9\linewidth,height=0.7\linewidth]{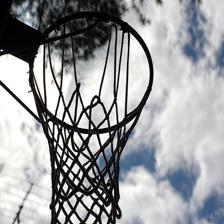} & 
    \centering\includegraphics[width=0.9\linewidth,height=0.7\linewidth]{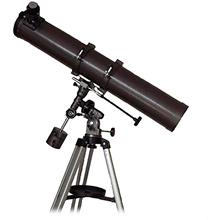} &
    \centering\includegraphics[width=0.9\linewidth,height=0.7\linewidth]{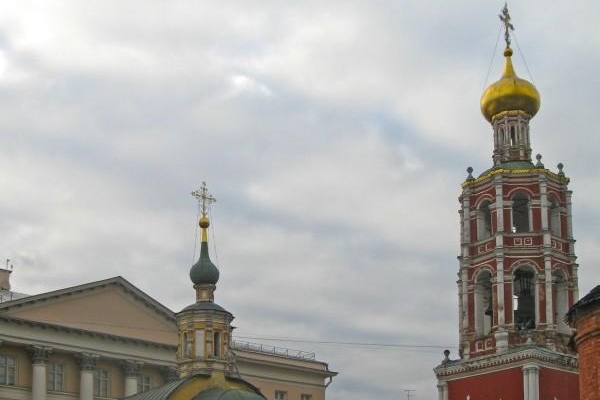} & \\
    \textit{\footnotesize what kind of plant is this?} &  \textit{\footnotesize which type of item is depicted in the image?} &  \textit{\footnotesize Who is the creator of this object?} & \textit{\footnotesize What is this building dedicated to?} \\
    \midrule
    \footnotesize\textbf{PreFLMR~\cite{lin2024preflmr}}: Hippeastrum is a genus of about 90 species and over 600 hybrids and cultivars of perennial herbaceous bulbous plants. & 
    \footnotesize\textbf{PreFLMR~\cite{lin2024preflmr}}: Handball (also known as team handball, European handball or Olympic handball) is a team sport in which two teams [...] & 
    \footnotesize\textbf{PreFLMR~\cite{lin2024preflmr}}: Optical axis of the Schmidt design creates a Schmidt-Newtonian telescope. The addition of a convex secondary mirror to [...] & 
    \footnotesize\textbf{PreFLMR~\cite{lin2024preflmr}}: [...] the reign of the Grand Prince Vsevolod the Big Nest of Vladimir-Suzdal to the honour of Saint Demetrius of Thessaloniki. \\
    \midrule
    \footnotesize\textbf{\ours~\cite{caffagni2025recurrence}}: The \textbf{poinsettia} (or ``Euphorbia pulcherrima'') is a commercially important plant species of the diverse spurge family [...] & 
    \footnotesize\textbf{\ours~\cite{caffagni2025recurrence}}: Netball is a ball sport played by two teams of seven players, usually on an indoor court, and is predominantly played by women. & 
    \footnotesize\textbf{\ours~\cite{caffagni2025recurrence}}: Manageable at large focal ratios -- most Schiefspieglers use f/15 or longer, which tends to restrict useful observation to the moon [...] & 
    \footnotesize\textbf{\ours~\cite{caffagni2025recurrence}}: [...] dedicated to \textbf{St Peter of Moscow}, was long regarded as a typical monument of the Naryshkin style and dated to 1692. \\
    
    \midrule
    \footnotesize\textbf{\colorbox{ourcolor}{\oursnew (Ours):}} The \textbf{poinsettia} (or ``Euphorbia pulcherrima'') is a commercially important plant species of the diverse spurge family [...] & 
    \footnotesize\textbf{\colorbox{ourcolor}{\oursnew (Ours):}} A \textbf{basket} is a container that is traditionally constructed from stiff fibers and can be made from a range of materials [...] & 
    \footnotesize\textbf{\colorbox{ourcolor}{\oursnew (Ours):}} [...] as the Gregorian telescope. \textbf{Isaac Newton} has been generally credited with building the first reflecting telescope in 1668. & 
    \footnotesize\textbf{\colorbox{ourcolor}{\oursnew (Ours):}} [...] dedicated to \textbf{St Peter of Moscow}, was long regarded as a typical monument of the Naryshkin style and dated to 1692. \\
    \addlinespace[1mm]
    \centering\includegraphics[width=0.75\linewidth,height=0.55\linewidth]{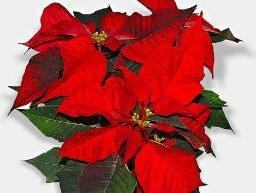} &
    \centering\includegraphics[width=0.75\linewidth,height=0.55\linewidth]{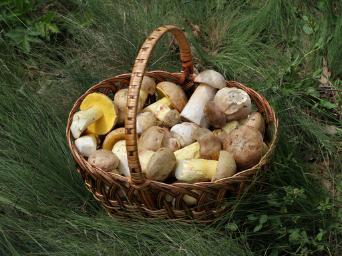}  & 
    \centering\includegraphics[width=0.75\linewidth,height=0.55\linewidth]{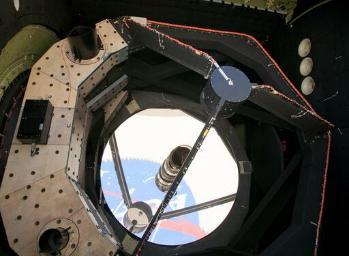} &
    \centering\includegraphics[width=0.75\linewidth,height=0.55\linewidth]{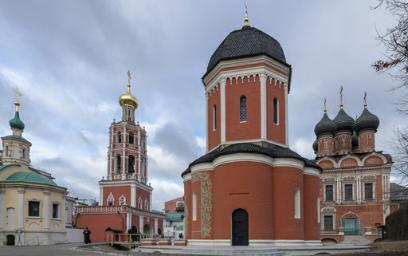} & \\
    \\
    \midrule
    \multicolumn{2}{c}{\textbf{E-VQA}} & \multicolumn{2}{c}{\textbf{OKVQA}} \\
    \midrule
    \centering\includegraphics[width=0.9\linewidth,height=0.7\linewidth]{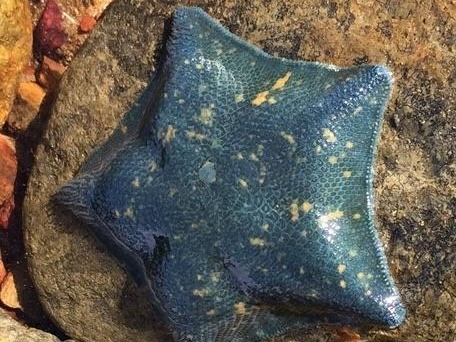} &
    \centering\includegraphics[width=0.9\linewidth,height=0.7\linewidth]{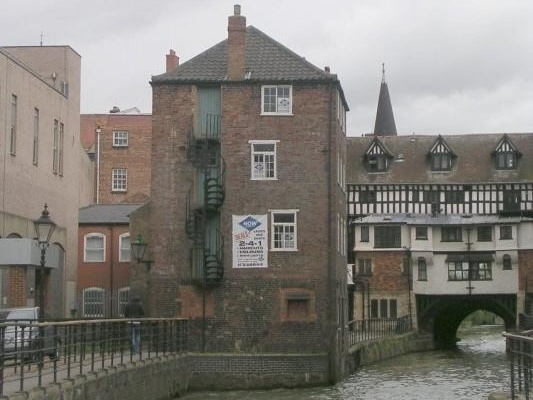} & 
    \centering\includegraphics[width=0.9\linewidth,height=0.7\linewidth]{images/qualitatives/m2kr/okvqa/15035.jpg} &
    \centering\includegraphics[width=0.9\linewidth,height=0.7\linewidth]{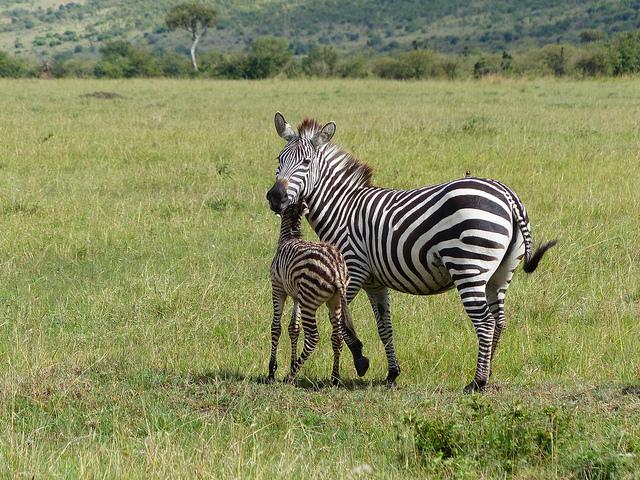} & \\
    \textit{\footnotesize In which part of the world does this animal live?} &  \textit{\footnotesize When was this bridge built?} &  \textit{\footnotesize What model of computer is this?} & \textit{\footnotesize Where would one find these animals?} \\
    \midrule
    \footnotesize\textbf{PreFLMR~\cite{lin2024preflmr}}: Henricia leviuscula, the ``Pacific blood star'', it is a species of sea star found along the Pacific coast of North America. & 
    \footnotesize\textbf{PreFLMR~\cite{lin2024preflmr}}: The bridge was built about \textbf{1160 AD} and a bridge chapel was built dedicated to Thomas Becket in 1235 on [...] &
    \footnotesize\textbf{PreFLMR~\cite{lin2024preflmr}}: A Compact \textbf{Macintosh} (or Compact \textbf{Mac}) is an all-in-one Apple \textbf{Mac} computer with a display integrated in the [...] & 
    \footnotesize\textbf{PreFLMR~\cite{lin2024preflmr}}: The Sudanian Savanna is a broad belt of tropical savanna that runs east and west across the \textbf{African continent}, from the [...] \\

    \midrule
    \footnotesize\textbf{\ours~\cite{caffagni2025recurrence}}: Evasterias troschelii is a species of starfish in the family Asteriidae. Its common names include the mottled star [...] & 
    \footnotesize\textbf{\ours~\cite{caffagni2025recurrence}}: The Ripon Canal is located in North Yorkshire, England. It was built by the canal engineer William Jessop to link the city of Ripon [...] & 
    \footnotesize\textbf{\ours~\cite{caffagni2025recurrence}}: The PC-D and PC-X were personal computers sold by Siemens between 1982 (PC-X)/1984 (PC-D) and 1986. The PC-D was [...] & 
    \footnotesize\textbf{\ours~\cite{caffagni2025recurrence}}: Assassination attempt by one of Loveless' henchwomen, West mistakes a female guest for a disguised Gordon resulting in [...] \\
    
    \midrule
    \footnotesize\textbf{\colorbox{ourcolor}{\oursnew (Ours):}} Patiriella regularis, or \textbf{New Zealand} common cushion star, is a sea star of the family Asterinidae, native to \textbf{New Zealand}. & 
    \footnotesize\textbf{\colorbox{ourcolor}{\oursnew (Ours):}} The bridge was built about \textbf{1160 AD} and a bridge chapel was built dedicated to Thomas Becket in 1235 on [...] & 
    \footnotesize\textbf{\colorbox{ourcolor}{\oursnew (Ours):}} Prior versions of the \textbf{Mac} mini were much more difficult to open. Some \textbf{Mac} mini owners used a putty knife or a pizza [...] & 
    \footnotesize\textbf{\colorbox{ourcolor}{\oursnew (Ours):}} Shaba National Reserve was the setting for the book and film ``Born Free'', for the film ``Out of \textbf{Africa}'' and for [...] \\
    \addlinespace[1mm]
    \centering\includegraphics[width=0.75\linewidth,height=0.55\linewidth]{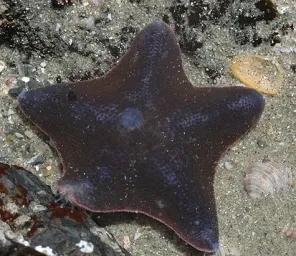} &
    \centering\includegraphics[width=0.75\linewidth,height=0.55\linewidth]{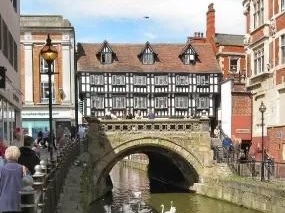}  & 
    \centering\includegraphics[width=0.75\linewidth,height=0.55\linewidth]{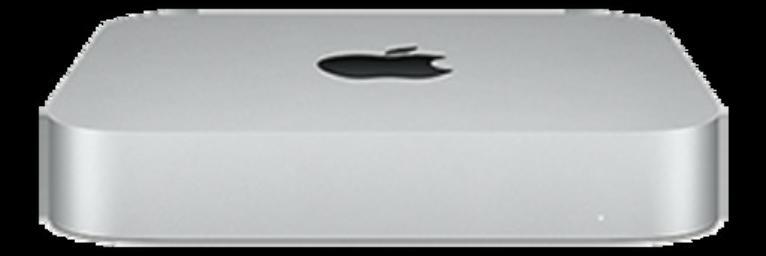} &
    \centering\includegraphics[width=0.75\linewidth,height=0.55\linewidth]{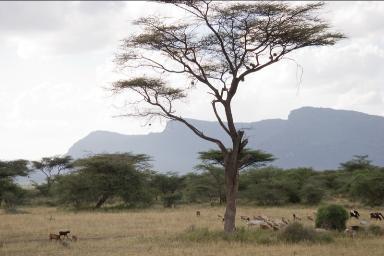} & \\
    \bottomrule
    \end{tabular}
    }
    \vspace{-0.15cm}
    \caption{Qualitative results on the M2KR benchmark~\cite{lin2024preflmr}, for datasets that include document images. We highlight the reference answer in bold font whenever it is found in the retrieved text. For \oursnew, we also add the document image attached to the text.}
    \label{fig:qualitatives_m2kr_2}
\end{figure*}

\tit{Qualitative Results on M2KR}
In Fig.~\ref{fig:qualitatives_m2kr_1} and Fig.~\ref{fig:qualitatives_m2kr_2}, we provide a qualitative comparison between PreFLMR, the original \ours, and our proposed \oursnew. In particular, to enable a direct comparison with PreFLMR, Fig.~\ref{fig:qualitatives_m2kr_1} focuses solely on M2KR datasets that do not include document images. As shown, \oursnew consistently retrieves information that is more contextually relevant and detailed for the given queries.
In contrast, Fig.~\ref{fig:qualitatives_m2kr_2} includes document images whenever available in the retrieved content\footnote{Note that both \ours and \oursnew employ datasets augmented with document images when available. For space constraints, in the reported qualitative results, we only include document images retrieved by \oursnew.}. These examples highlight that incorporating visual information from document images substantially enhances the ability of the model to answer queries accurately, effectively complementing textual content with visual context.

\begin{figure*}[t]
    \centering
    \resizebox{\linewidth}{!}{
    \setlength{\tabcolsep}{.75em}
    \begin{tabular}{p{2.5cm} p{4.8cm} p{4.8cm} p{4.8cm} p{0.1mm}}
    \toprule
    \multicolumn{5}{c}{\textbf{Task \#3:} $q^T\rightarrow(d^T,d^V)$} \\
    \midrule
    \multicolumn{5}{c}{\textit{\footnotesize Find a news image that matches the provided caption.}} \\
    \multicolumn{5}{c}{\textit{\footnotesize Having finished with the likes of Mercedes driver Lewis Hamilton, the world's press were treated to a rare display of candor from F1 team principals.}} \\
    \midrule
    \addlinespace[2mm]

    & \centering\includegraphics[height=0.55\linewidth]{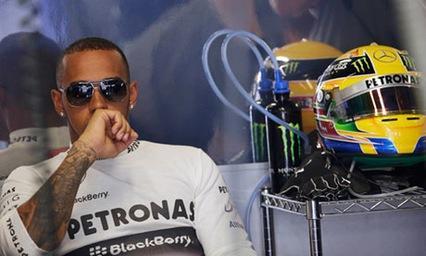} & \centering\includegraphics[height=0.55\linewidth]{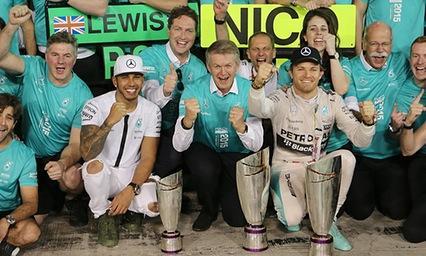} & \centering\includegraphics[height=0.55\linewidth]{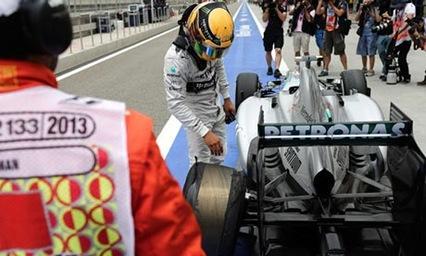} & \\    
    \addlinespace[1mm]
    \multirow{2}[0]{*}{\footnotesize\textbf{\centering UniIR~\cite{wei2024uniir}:}} & \footnotesize The Mercedes mechanics, as if impervious to the heat of the Lombardy sunshine, moved quickly on Friday, their urgent actions resembling those of. & \footnotesize Mercedes are to investigate why Lewis Hamilton, the fastest Formula One driver of the modern era, is now not even the quickest man in their team. & \footnotesize Lewis Hamilton has not only brought the best out of Mercedes this year but also given fresh impetus to his team mate, Nico Rosberg. & \\
    \midrule
    \addlinespace[2mm]
    
    & \centering\fcolorbox{red}{white}{\includegraphics[height=0.55\linewidth]{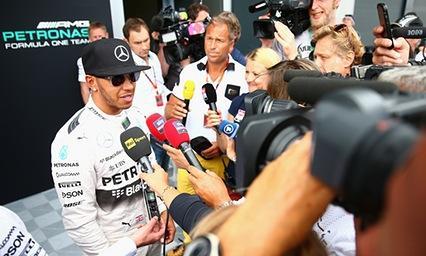}} & \centering\includegraphics[height=0.55\linewidth]{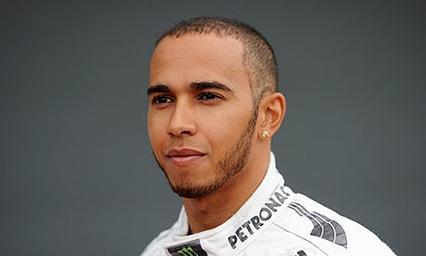} & \centering\includegraphics[height=0.55\linewidth]{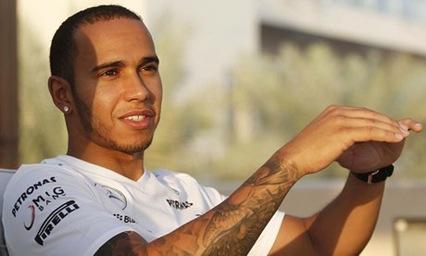} & \\    
    \addlinespace[1mm]
    \multirow{2}[0]{*}{\footnotesize\textbf{\centering \colorbox{ourcolor}{\oursnew (Ours):}}} & \footnotesize The second of the two press conferences held at a Formula One race weekend tends to be the drier. & \footnotesize Lewis Hamilton is banking on a return to one of his favorite tracks this weekend to kickstart his faltering season. & \footnotesize Lewis Hamilton says Mercedes can manage without Ross Brawn, the most successful team principal of his generation, because he believes no team depends. & \\
    \bottomrule
    
    \end{tabular}
    }
    \vspace{-0.15cm}
    \caption{Qualitative results on the Task \#3 of the M-BEIR benchmark~\cite{wei2024uniir}, using an example from the EDIS subset~\cite{liu2023edis}. Ground-truth image-text documents have a red frame around the image.}
    \label{fig:qualitatives_mbeir_edis_task_3}    
\end{figure*}

\begin{figure*}[t]
    \centering
    \resizebox{\linewidth}{!}{
    \setlength{\tabcolsep}{.75em}
    \begin{tabular}{p{2.5cm} p{5.4cm} p{5.4cm} p{5.4cm} p{0.1mm}}
    \toprule
    \addlinespace[2mm]
    \multicolumn{5}{c}{\textbf{Task \#8:} $(q^T,q^V)\rightarrow(d^T,d^V)$} \\
    \addlinespace[1mm]
    \midrule
    \addlinespace[1.5mm]
    & \centering\multirow{4}{*}{\includegraphics[height=0.49\linewidth]{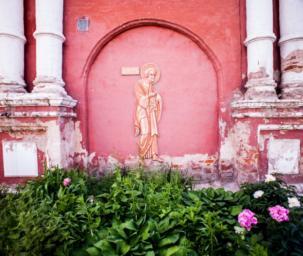}} & & \\
    & & & & \\
    & & \multicolumn{3}{l}{\textit{\footnotesize I want to address the query about this picture. Please pull up a relevant Wikipedia section and image.}} \\
    & & \multicolumn{3}{l}{\textit{\footnotesize What is this building dedicated to?}} \\
    & & & & \\
    & & & & \\
    \addlinespace[2mm]
    \midrule
    \addlinespace[2mm]

    & \centering\includegraphics[height=0.55\linewidth]{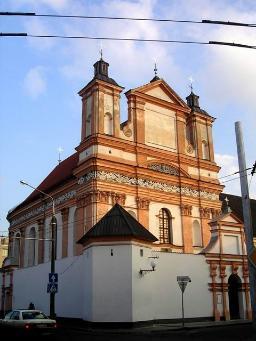} & \centering\includegraphics[height=0.55\linewidth]{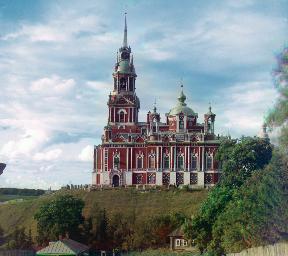} & \centering\includegraphics[height=0.55\linewidth]{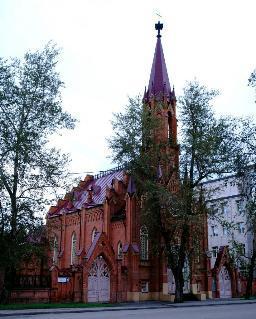} & \\    
    \addlinespace[1mm]
    \multirow{2}[0]{*}{\footnotesize\textbf{\centering UniIR~\cite{wei2024uniir}:}} & \footnotesize Church of Annunciation of Virgin Mary (Hrodna). but allowed nuns to stay and live there. Also, Benedictine nuns from Nyasvizh, Dominicans from [...] & \footnotesize Cathedral of St. Nicholas (Mozhaysk). The construction started in 1802 and finished only in 1814.The almost finished cathedral was badly damaged in 1812. & \footnotesize Church of Our Lady of the Assumption, Irkutsk. The Church of Our Lady of the Assumption also called the ``Polish Church'', it is a Catholic church in [...] & \\
    \midrule
    \addlinespace[2mm]

    & \centering\includegraphics[height=0.55\linewidth]{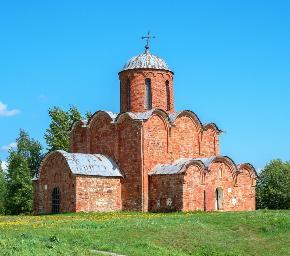} & \centering\fcolorbox{red}{white}{\includegraphics[height=0.55\linewidth]{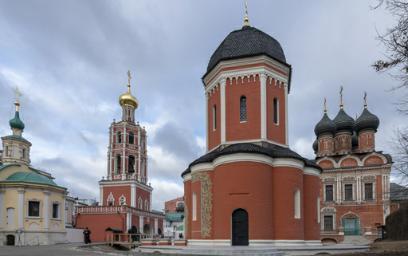}} & \centering\includegraphics[height=0.55\linewidth]{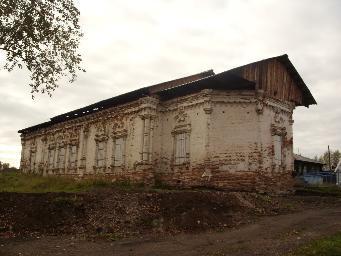} & \\    
    \addlinespace[1mm]
    \multirow{2}[0]{*}{\footnotesize\textbf{\centering \colorbox{ourcolor}{\oursnew (Ours):}}} & \footnotesize Transfiguration Church in Kovalyovo. Krasnorechyev. \#\# Architecture. The church is constructed in brick, and has one dome. It has a single apse and four square columns. This design is [...] & \footnotesize Vysokopetrovsky Monastery. in the Naryshkin Baroque style of architecture associated with their name. In the mid-18th century, several subsidiary structures were added, possibly based [...] & \footnotesize Church of Our Lady of the Sign, Verkhoturye. Church of Our Lady of the Sign - is an Orthodox church in Verkhoturye, Sverdlovsk oblast.The building was granted the status of regional [...] & \\
    \bottomrule
    
    \end{tabular}
    }
    \vspace{-0.15cm}
    \caption{Qualitative results on the Task \#8 of the M-BEIR benchmark~\cite{wei2024uniir}, using an example from the InfoSeek subset~\cite{chen2023can}. Ground-truth image-text documents have a red frame around the image.}
    \label{fig:qualitatives_mbeir_infoseek_task_8}    
\end{figure*}

\tit{Qualitative Results on M-BEIR} Fig.~\ref{fig:qualitatives_mbeir_edis_task_3} and Fig.~\ref{fig:qualitatives_mbeir_infoseek_task_8} present qualitative results comparing UniIR and \oursnew on two different tasks. In Task \#3, we employ an example from the EDIS dataset where only the document side is multimodal, whereas in Task \#8, where we use an example from the InfoSeek dataset, both the query and document sides are fully multimodal. The figures show the top-3 retrieved documents for each task, highlighting \oursnew’s ability to consistently retrieve the correct documents, while UniIR struggles to locate the relevant ones.

\begin{figure*}[t]
\begin{minipage}{0.325\linewidth}
\scriptsize{\textbf{Q:} What is the area in square kilometer of this lake?\vspace{0.1cm}}\\
\begin{minipage}{0.443\linewidth}
\includegraphics[width=1.\linewidth,height=0.88\linewidth]{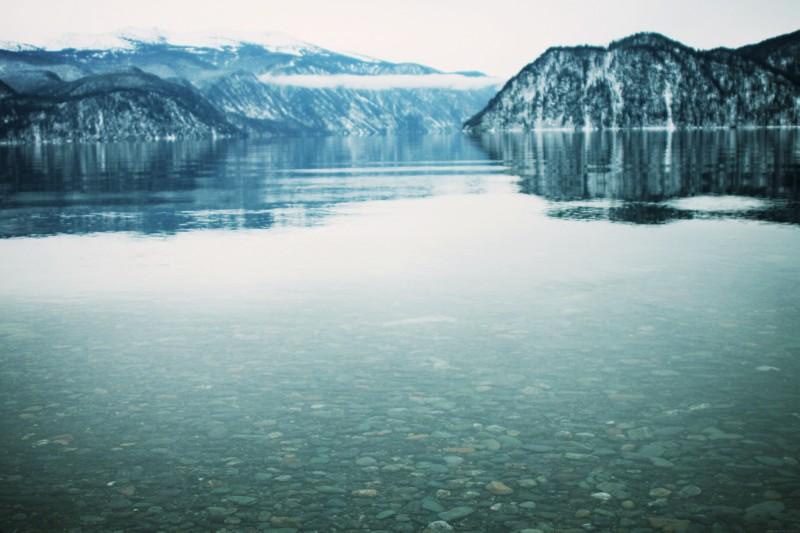}
\end{minipage}
\hfill
\begin{minipage}{0.53\linewidth}
\scriptsize{
\textbf{PreFLMR~\cite{lin2024preflmr}}:\\
25.2 km² \textcolor{red}{\xmark} \\
\textbf{\ours~\cite{caffagni2025recurrence}}:\\
Not enough information \textcolor{red}{\xmark} \\
\textbf{\oursnew (Ours):}\\
233 \textcolor[HTML]{00b050}{\cmark}
}
\end{minipage}
\vspace{0.2cm}
\end{minipage}
\hspace{0.02cm}
\begin{minipage}{0.325\linewidth}
\scriptsize{\textbf{Q:} In which year did this building come into service?\vspace{0.1cm}}\\
\begin{minipage}{0.443\linewidth}
\includegraphics[width=1.\linewidth,height=0.88\linewidth]{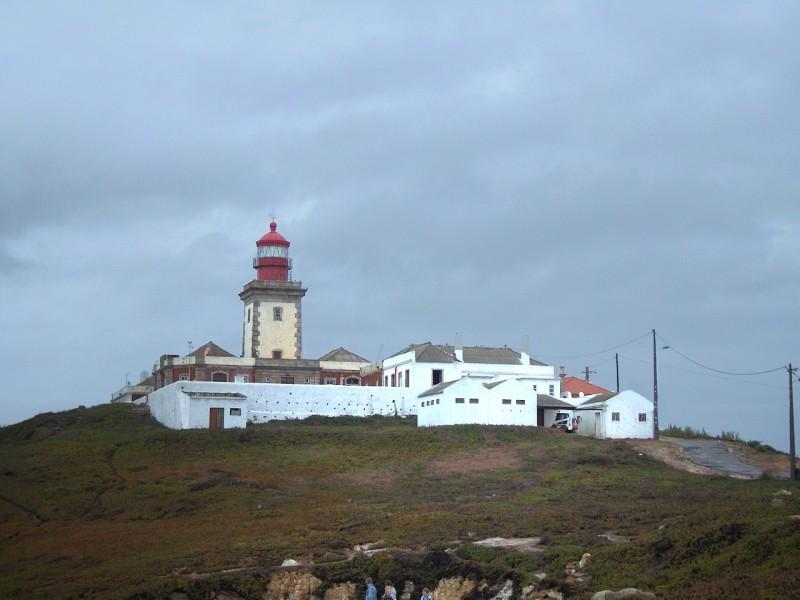}
\end{minipage}
\hfill
\begin{minipage}{0.53\linewidth}
\scriptsize{
\textbf{PreFLMR~\cite{lin2024preflmr}}:\\
1917 \textcolor{red}{\xmark} \\
\textbf{ReT~\cite{caffagni2025recurrence}}:\\
1914 \textcolor{red}{\xmark} \\
\textbf{\oursnew (Ours):} \\
1772 \textcolor[HTML]{00b050}{\cmark}
}
\end{minipage}
\vspace{0.2cm}
\end{minipage}
\hspace{0.02cm}
\vspace{0.1cm}
\begin{minipage}{0.325\linewidth}
\scriptsize{\textbf{Q:} What is the location of this garden?\vspace{0.1cm}}\\
\begin{minipage}{0.443\linewidth}
\includegraphics[width=1.\linewidth,height=0.88\linewidth]{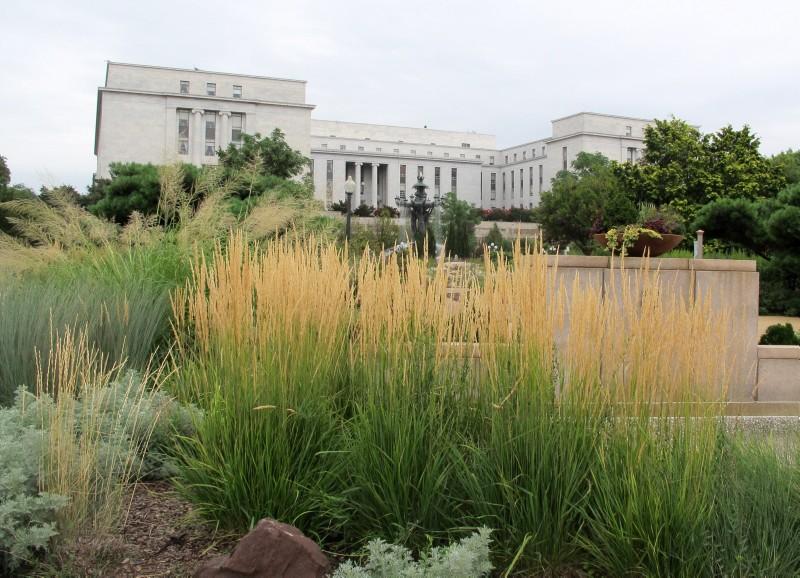}
\end{minipage}
\hfill
\begin{minipage}{0.53\linewidth}
\scriptsize{
\textbf{PreFLMR~\cite{lin2024preflmr}}:\\
Washington Park, Denver \textcolor{red}{\xmark} \\
\textbf{\ours~\cite{caffagni2025recurrence}}:\\
United States Botanic Garden \textcolor{red}{\xmark} \\
\textbf{\oursnew (Ours):}\\
National Mall \textcolor[HTML]{00b050}{\cmark}
}
\end{minipage}
\vspace{0.2cm}
\end{minipage}
\begin{minipage}{0.325\linewidth}
\scriptsize{\textbf{Q:} What is the height of this bridge in meter?\vspace{0.1cm}}\\
\begin{minipage}{0.443\linewidth}
\includegraphics[width=1.\linewidth,height=0.88\linewidth]{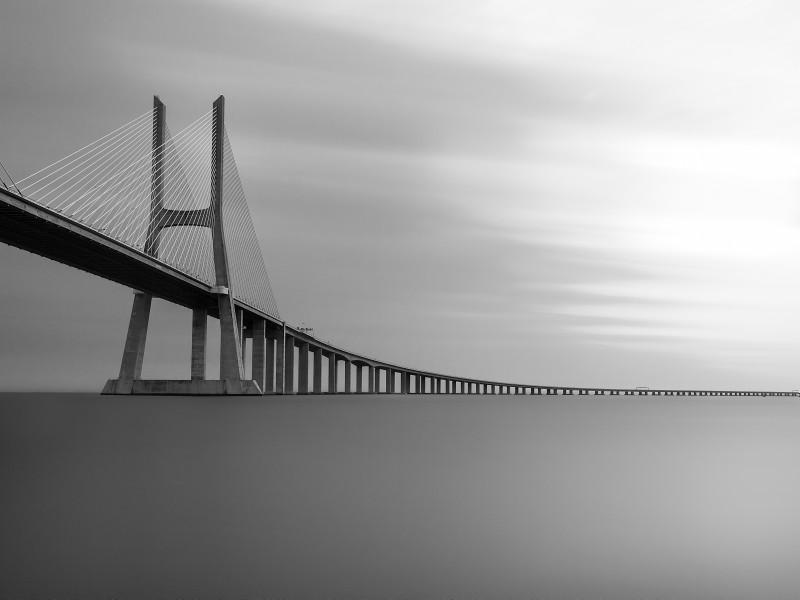}
\end{minipage}
\hfill
\begin{minipage}{0.53\linewidth}
\scriptsize{
\textbf{PreFLMR~\cite{lin2024preflmr}}:\\
70 \textcolor{red}{\xmark} \\
\textbf{\ours~\cite{caffagni2025recurrence}}:\\
95 \textcolor{red}{\xmark} \\
\textbf{\oursnew (Ours):}\\
150 \textcolor[HTML]{00b050}{\cmark}
}
\end{minipage}
\end{minipage}
\hspace{0.02cm}
\begin{minipage}{0.325\linewidth}
\scriptsize{\textbf{Q:} In which year did this building officially open?\vspace{0.1cm}}\\
\begin{minipage}{0.443\linewidth}
\includegraphics[width=1.\linewidth,height=0.88\linewidth]{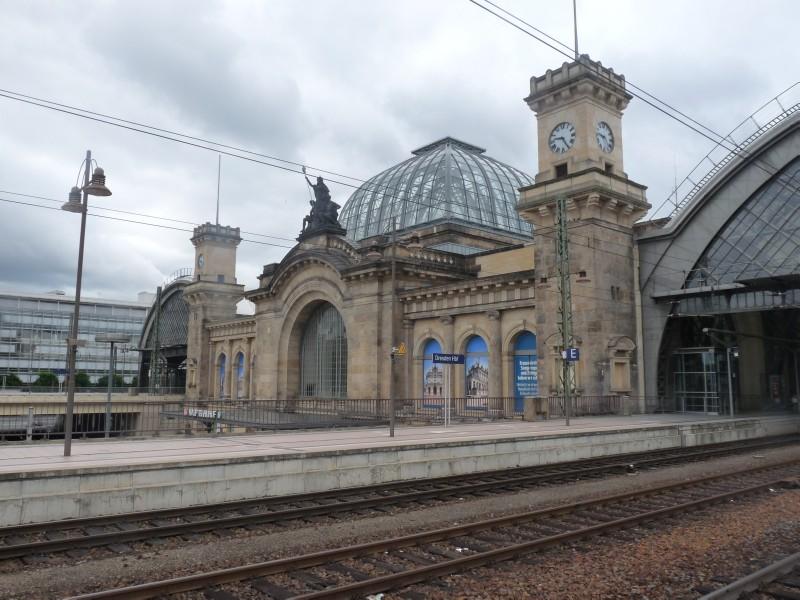}
\end{minipage}
\hfill
\begin{minipage}{0.53\linewidth}
\scriptsize{
\textbf{PreFLMR~\cite{lin2024preflmr}}:\\
1934 \textcolor{red}{\xmark} \\
\textbf{\ours~\cite{caffagni2025recurrence}}:\\
1841 \textcolor{red}{\xmark} \\
\textbf{\oursnew (Ours):}\\
1898 \textcolor[HTML]{00b050}{\cmark}
}
\end{minipage}
\end{minipage}
\hspace{0.02cm}
\vspace{0.1cm}
\begin{minipage}{0.325\linewidth}
\scriptsize{\textbf{Q:} Which city or region does this building locate in?\vspace{0.1cm}}\\
\begin{minipage}{0.443\linewidth}
\includegraphics[width=1.\linewidth,height=0.88\linewidth]{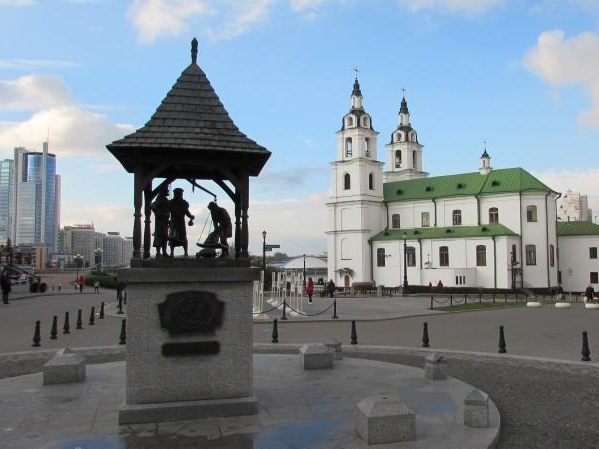}
\end{minipage}
\hfill
\begin{minipage}{0.53\linewidth}
\scriptsize{
\textbf{PreFLMR~\cite{lin2024preflmr}}:\\
Kamianets-Podilsky \textcolor{red}{\xmark} \\
\textbf{\ours~\cite{caffagni2025recurrence}}:\\
Kazan Kremlin \textcolor{red}{\xmark} \\
\textbf{\oursnew (Ours):}\\
Minsk \textcolor[HTML]{00b050}{\cmark}
}
\end{minipage}
\end{minipage}
\vspace{-0.25cm}
\caption{Sample results for the knowledge-intensive VQA task on the validation split of InfoSeek, augmenting Qwen2.5-VL with context retrieved by different multimodal retrieval models.}
\label{fig:qualitatives_rag_infoseek}
\vspace{-0.2cm}
\end{figure*}

\begin{figure*}[t]
\begin{minipage}{0.325\linewidth}
\scriptsize{\textbf{Q:} What gender was the deity worshiped at this temple in its earliest phase?\vspace{0.1cm}}\\
\begin{minipage}{0.443\linewidth}
\includegraphics[width=1.\linewidth,height=0.88\linewidth]{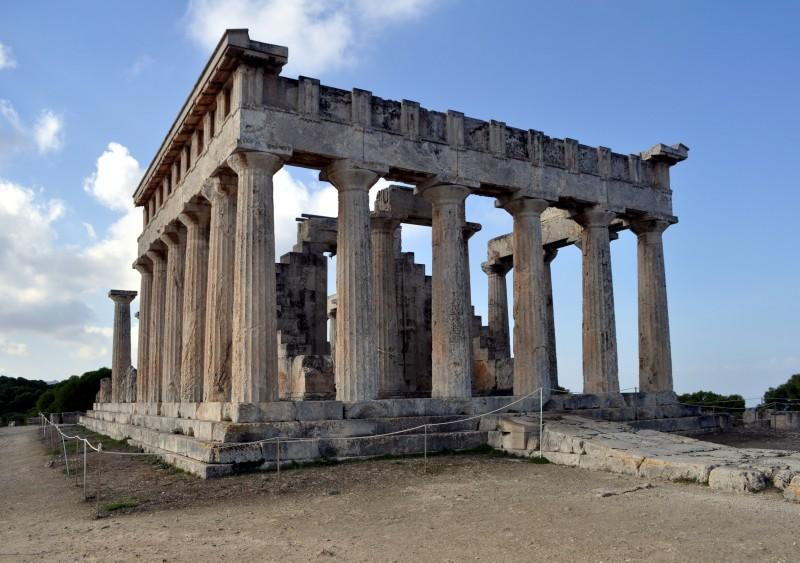}
\end{minipage}
\hfill
\begin{minipage}{0.53\linewidth}
\scriptsize{
\textbf{PreFLMR~\cite{lin2024preflmr}}:\\
Male \textcolor{red}{\xmark} \\
\textbf{\ours~\cite{caffagni2025recurrence}}:\\
Male \textcolor{red}{\xmark} \\
\textbf{\oursnew (Ours):}\\
Female \textcolor[HTML]{00b050}{\cmark}
}
\end{minipage}
\vspace{0.2cm}
\end{minipage}
\hspace{0.02cm}
\begin{minipage}{0.325\linewidth}
\scriptsize{\textbf{Q:} When did the San Diego savings bank leave this building?\vspace{0.1cm}}\\
\begin{minipage}{0.443\linewidth}
\includegraphics[width=1.\linewidth,height=0.88\linewidth]{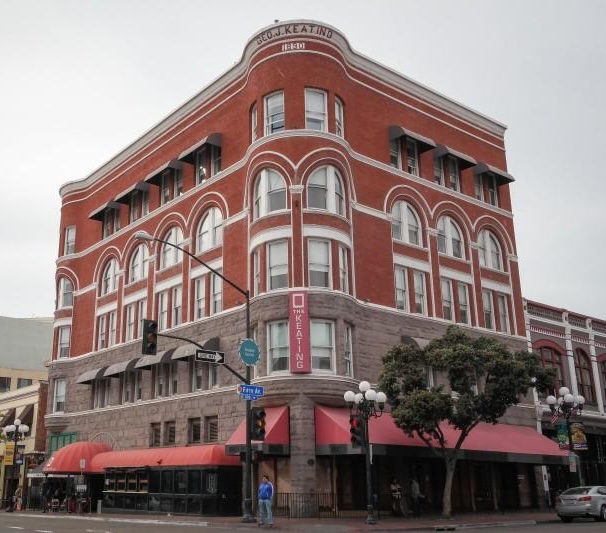}
\end{minipage}
\hfill
\begin{minipage}{0.53\linewidth}
\scriptsize{
\textbf{PreFLMR~\cite{lin2024preflmr}}:\\
March 18, 1994 \textcolor{red}{\xmark} \\
\textbf{ReT~\cite{caffagni2025recurrence}}:\\
The San Diego savings bank left the building in 1930 \textcolor{red}{\xmark} \\
\textbf{\oursnew (Ours):} \\
1912 \textcolor[HTML]{00b050}{\cmark}
}
\end{minipage}
\vspace{0.2cm}
\end{minipage}
\hspace{0.02cm}
\vspace{0.1cm}
\begin{minipage}{0.325\linewidth}
\scriptsize{\textbf{Q:} What was the building at this canal converted into in 2012?\vspace{0.1cm}}\\
\begin{minipage}{0.443\linewidth}
\includegraphics[width=1.\linewidth,height=0.88\linewidth]{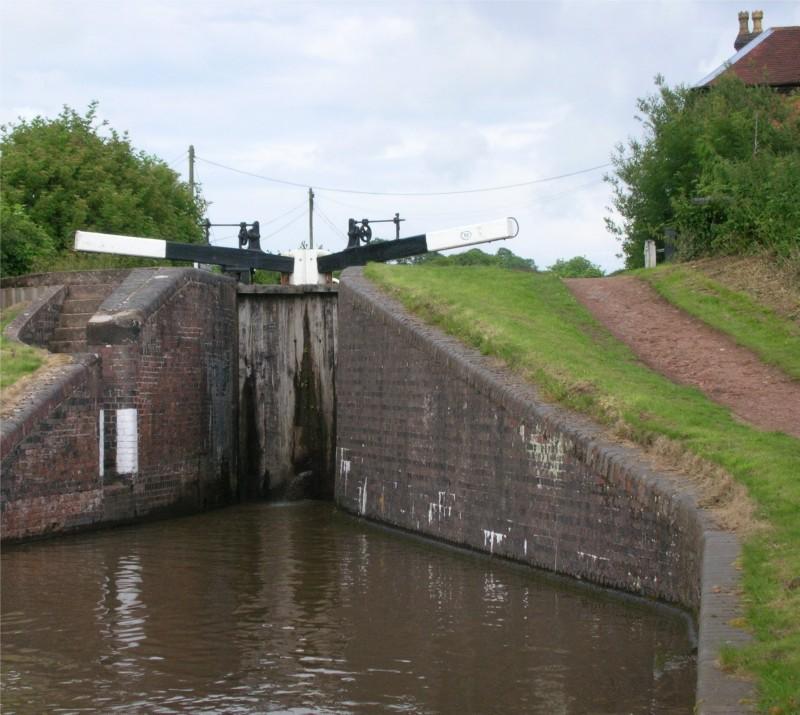}
\end{minipage}
\hfill
\begin{minipage}{0.53\linewidth}
\scriptsize{
\textbf{PreFLMR~\cite{lin2024preflmr}}:\\
A museum \textcolor{red}{\xmark} \\
\textbf{\ours~\cite{caffagni2025recurrence}}:\\
A museum and an heritage center \textcolor{red}{\xmark} \\
\textbf{\oursnew (Ours):}\\
Four residential apartments \textcolor[HTML]{00b050}{\cmark}
}
\end{minipage}
\vspace{0.2cm}
\end{minipage}
\begin{minipage}{0.325\linewidth}
\scriptsize{\textbf{Q:} When was this church established?\vspace{0.1cm}}\\
\begin{minipage}{0.443\linewidth}
\includegraphics[width=1.\linewidth,height=0.88\linewidth]{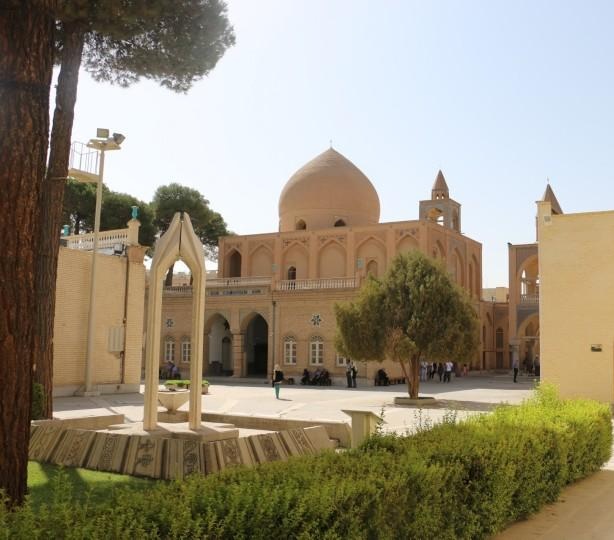}
\end{minipage}
\hfill
\begin{minipage}{0.53\linewidth}
\scriptsize{
\textbf{PreFLMR~\cite{lin2024preflmr}}:\\
1986 \textcolor{red}{\xmark} \\
\textbf{\ours~\cite{caffagni2025recurrence}}:\\
Not enough information \textcolor{red}{\xmark} \\
\textbf{\oursnew (Ours):}\\
1606 \textcolor[HTML]{00b050}{\cmark}
}
\end{minipage}
\end{minipage}
\hspace{0.02cm}
\begin{minipage}{0.325\linewidth}
\scriptsize{\textbf{Q:} Is the pale-billed woodpecker a large or small bird?\vspace{0.1cm}}\\
\begin{minipage}{0.443\linewidth}
\includegraphics[width=1.\linewidth,height=0.88\linewidth]{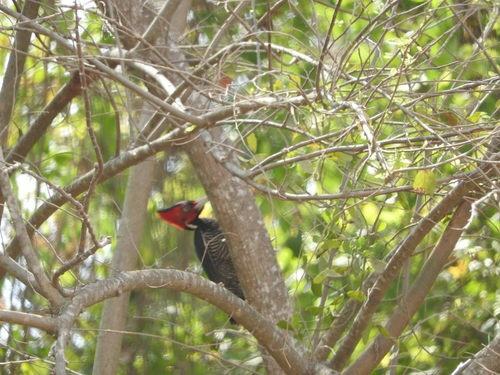}
\end{minipage}
\hfill
\begin{minipage}{0.53\linewidth}
\scriptsize{
\textbf{PreFLMR~\cite{lin2024preflmr}}:\\
The pale-billed woodpecker is a large bird \textcolor{red}{\cmark} \\
\textbf{\ours~\cite{caffagni2025recurrence}}:\\
The pale-billed woodpecker is a small bird \textcolor{red}{\xmark} \\
\textbf{\oursnew (Ours):}\\
Large \textcolor[HTML]{00b050}{\cmark}
}
\end{minipage}
\end{minipage}
\hspace{0.02cm}
\vspace{0.1cm}
\begin{minipage}{0.325\linewidth}
\scriptsize{\textbf{Q:} What color is this plant flowers range from yellow to?\vspace{0.1cm}}\\
\begin{minipage}{0.443\linewidth}
\includegraphics[width=1.\linewidth,height=0.88\linewidth]{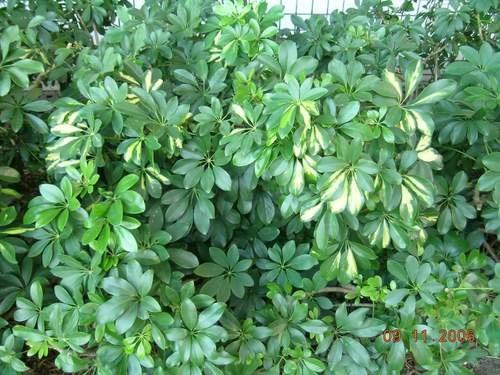}
\end{minipage}
\hfill
\begin{minipage}{0.53\linewidth}
\scriptsize{
\textbf{PreFLMR~\cite{lin2024preflmr}}:\\
Pink or lilac-red \textcolor{red}{\xmark} \\
\textbf{\ours~\cite{caffagni2025recurrence}}:\\
Lilac \textcolor{red}{\xmark} \\
\textbf{\oursnew (Ours):}\\
Green \textcolor[HTML]{00b050}{\cmark}
}
\end{minipage}
\end{minipage}
\vspace{-0.25cm}
\caption{Sample results for the knowledge-intensive VQA task on the test split of Encyclopedic-VQA, augmenting Qwen2.5-VL with context retrieved by different multimodal retrieval models.}
\label{fig:qualitatives_rag_evqa}
\vspace{-0.2cm}
\end{figure*}

\tit{Qualitative Results on Retrieval-Augmented VQA} 
Qualitative results on sample image-question pairs from InfoSeek and Encyclopedic-VQA are shown in Fig.~\ref{fig:qualitatives_rag_infoseek} and Fig.~\ref{fig:qualitatives_rag_evqa}, comparing answers generated by augmenting Qwen2.5-VL with context retrieved by different multimodal retrieval models, including PreFLMR, \ours, and \oursnew. The results demonstrate that \oursnew consistently retrieves more accurate and relevant documents, enabling better responses to specific multimodal questions and outperforming the other approaches.

\end{document}